\documentclass{article}
\usepackage[a4paper, portrait]{geometry}
\usepackage[english]{babel}
\usepackage[utf8]{inputenc}
\usepackage[T1]{fontenc}
\usepackage{helvet}
\usepackage{etoolbox}
\usepackage{graphicx}
\usepackage{titlesec}
\usepackage{caption}
\usepackage{booktabs}
\usepackage{xcolor} 
\usepackage[colorlinks=false,citebordercolor=cyan,linkbordercolor=blue,urlbordercolor=cyan]{hyperref}
\usepackage{caption}
\usepackage{scrextend}
\usepackage{fancyhdr}
\usepackage{graphicx}
\usepackage{mathrsfs}
\usepackage{amsmath}
\usepackage{mathtools}
\usepackage{tabto}
\usepackage{amsfonts}
\usepackage{amssymb}
\usepackage{float}
\usepackage{url}
\usepackage{bbm}
\usepackage{algorithm}
\usepackage{algpseudocode}
\usepackage{setspace}
\let\Algorithm\algorithm
\renewcommand\algorithm[1][]{\Algorithm[#1]\setstretch{1.4}}
\usepackage{subfigure}
\usepackage{pgfplots}
\RequirePackage[style = numeric, backend = biber,maxbibnames=99]{biblatex}
\usepackage{breakurl}

\definecolor{HeiRot}{RGB}{198,24,38}

\newcounter{dummy}
\numberwithin{dummy}{section}

\newtheorem{definition}[dummy]{Definition}
\newtheorem{remark}[dummy]{Remark}
\newtheorem{example}[dummy]{Example}

\newcommand{\id}{\text{id}}

\makeatletter
\newcommand{\algmargin}{\the\ALG@thistlm}
\makeatother
\newlength{\whilewidth}
\algdef{SE}[parWHILE]{parWhile}{EndparWhile}[1]
{\parbox[t]{\dimexpr\linewidth-\algmargin}{%
		\hangindent\whilewidth\strut\algorithmicwhile\ #1\ \algorithmicdo\strut}}{\algorithmicend\ \algorithmicwhile}%
\algnewcommand{\parState}[1]{\State%
	\parbox[t]{\dimexpr\linewidth-\algmargin}{\strut #1\strut}}

\fancypagestyle{plain}{
	\fancyhf{}

	}
\usepackage{tikz}
\usepackage{listofitems} 
\tikzstyle{mynode}=[thick,draw=blue,fill=gray!20,circle,minimum size=22]
\usetikzlibrary{arrows.meta} 
\usepackage[outline]{contour} 
\contourlength{1.4pt}
\tikzset{>=latex} 
\usepackage{xcolor}
\colorlet{myred}{red!80!black}
\colorlet{myblue}{blue!80!black}
\colorlet{mygreen}{green!60!black}
\colorlet{myorange}{orange!70!red!60!black}
\colorlet{mydarkred}{red!30!black}
\colorlet{mydarkblue}{blue!40!black}
\colorlet{mydarkgreen}{green!30!black}
\tikzstyle{node}=[thick,circle,draw=myblue,minimum size=22,inner sep=0.5,outer sep=0.6]
\tikzstyle{node in}=[node,green!20!black,draw=mygreen!30!black,fill=mygreen!25]
\tikzstyle{node hidden}=[node,blue!20!black,draw=myblue!30!black,fill=myblue!20]
\tikzstyle{node diff}=[node,orange!20!black,draw=myorange!30!black,fill=myorange!20]
\tikzstyle{node out}=[node,red!20!black,draw=myred!30!black,fill=myred!20]
\tikzstyle{connect}=[thick,mydarkblue] 
\tikzstyle{connect arrow}=[-{Latex[length=4,width=3.5]},thick,mydarkblue,shorten <=0.5,shorten >=1]
\tikzset{ 
	node 1/.style={node in},
	node 2/.style={node hidden},
	node 3/.style={node out},
}
\def\nstyle{int(\lay<\Nnodlen?min(2,\lay):3)} 

\makeatletter
\patchcmd{\@maketitle}{\LARGE \@title}{\fontsize{16}{19.2}\selectfont\@title}{}{}
\makeatother

\usepackage{authblk}

\newsavebox\affbox
\author[1]{\textbf{Evelyn Herberg}}

\affil[1]{ Interdisciplinary Center for Scientific Computing, Ruprecht-Karls-University of Heidelberg, 69120 Heidelberg, Germany
}

\titlespacing\section{0pt}{12pt plus 4pt minus 2pt}{0pt plus 2pt minus 2pt}
\titlespacing\subsection{12pt}{12pt plus 4pt minus 2pt}{0pt plus 2pt minus 2pt}
\titlespacing\subsubsection{12pt}{12pt plus 4pt minus 2pt}{0pt plus 2pt minus 2pt}

\titleformat{\section}{\normalfont\fontsize{10}{15}\bfseries}{\thesection.}{1em}{}
\titleformat{\subsection}{\normalfont\fontsize{10}{15}\bfseries}{\thesubsection.}{1em}{}
\titleformat{\subsubsection}{\normalfont\fontsize{10}{15}\bfseries}{\thesubsubsection.}{1em}{}

\titleformat{\author}{\normalfont\fontsize{10}{15}\bfseries}{\thesection}{1em}{}

\title{\textbf{\huge Lecture Notes: \\Neural Network Architectures}}
\date{April 2023}    

\begin{document}

\pagestyle{headings}	
\newpage
\setcounter{page}{1}
\renewcommand{\thepage}{\arabic{page}}

\captionsetup[figure]{labelfont={bf},labelformat={default},labelsep=period,name={Figure }}	\captionsetup[table]{labelfont={bf},labelformat={default},labelsep=period,name={Table }}
\setlength{\parskip}{0.5em}
	
\maketitle

\section*{Acknowledgement}
These lecture notes were written to serve as theoretical background for programming sessions given by Florian Wolf during the SPP1962 Young Researchers' workshop on Deep Learning in March 2023. 

I do not claim originality, but merely collected material from various sources and wrote it down in a cohesive way. Any mistakes that I may have introduced are of course at my fault.

Especially, parts of these lecture notes are based on a student seminar at the University of Heidelberg that Roland Herzog and I organized together. I thank the involved students: Deniz Aydin, Laurin Ernst, Yanxin Jia, Xinyu Liang, Hannah Rickmann, Viktor Stein von Kamienski, Xiao Wang and Zixiang Zhou for their contributions.  

The main literature for the seminar, and consequently also for these lecture notes, was \cite{goodfellow2016deep} and \cite{ng2022}. Additional sources are mentioned throughout the document.

Furthermore, I thank Harbir Antil and the CMAI work group at George Mason University for their guidance in understanding Machine Learning from an optimal control point of view. The notation in this document is highly influenced by the CMAI work group, cf. \cite{antildiazherberg,antil2022deep,antil2020fractional}.

\begin{center}
	\small 
	Please send comments and remarks to
	\href{mailto:evelyn.herberg@iwr.uni-heidelberg.de}{\tt evelyn.herberg@iwr.uni-heidelberg.de}.
\end{center}

\newpage 
\tableofcontents

\newpage
\section{Introduction}
\label{sec:Intro}
Machine Learning (ML) denotes the field of study in which algorithms infer from given data how to perform a specific task, without being explicitly programmed for the task (Arthur Samuel, 1959). Here, we consider a popular subset of ML algorithms: \textbf{Neural Networks}. The inspiration for a Neural Network (NN) originates from the human brain, where biological neurons (nerve cells) respond to the activation of other neurons they are connected to. At a very simple level, neurons in the brain take electrical inputs that are then channeled to outputs. The sensitivity of this relation also depends on the strength of the connection, i.e. a neuron may be more responsive to one neuron, then to another.

\begin{figure}[h]
	  \begin{minipage}[c]{0.6\textwidth}
		\includegraphics[width=0.9\textwidth]{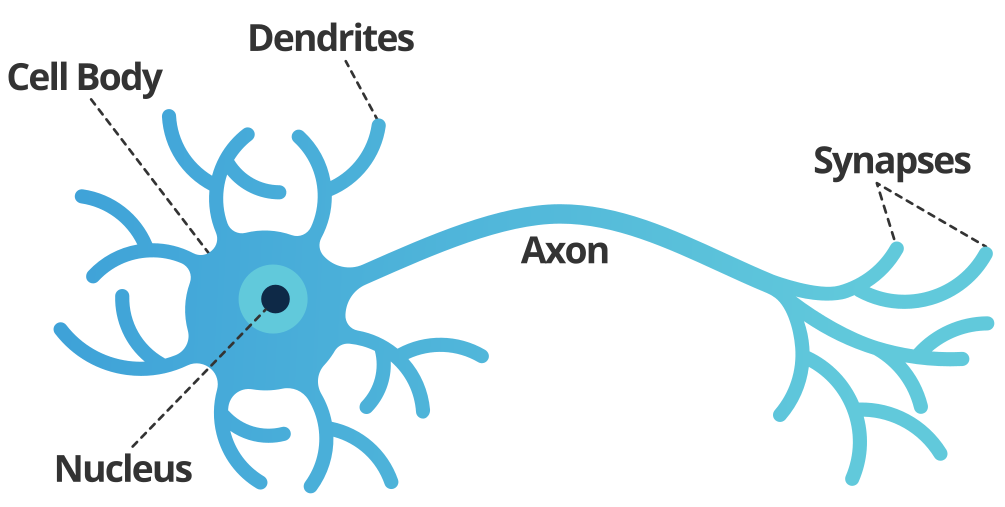}
	\end{minipage}\hfill
	\begin{minipage}[c]{0.4\textwidth}
		\caption{
			Brain Neuron Structure: electrical inputs are received through dendrites and transmitted via the axon to other cells. There are approximately 86 billion neurons in the human brain. Image modified from: \url{https://www.smartsheet.com/neural-network-applications}.
		} \label{fig:neuron}
	\end{minipage}
\end{figure}

For a single neuron/node with input $u \in \mathbb{R}^{n}$, 
a mathematical model, named the \textbf{perceptron} \cite{rosenblatt1958perceptron}, can be described as 
\begin{equation}\label{eq:perceptron}
	y = \sigma \left( \sum_{i=1}^{n} W_i u_i + b \right) = \sigma(W^{\top}u + b),
\end{equation}
where $y$ is the \textbf{activation} of the neuron/node, $W_i$ are the \textbf{weights} and $b$ is the \textbf{bias}.

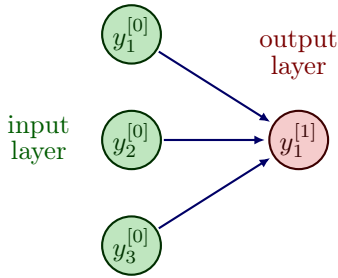
\begin{figure}[h]
	\begin{minipage}[c]{0.4\textwidth}
		\begin{center}
		\begin{tikzpicture}[x=2.2cm,y=1.4cm]
			\message{^^JNeural network with arrows}
			\readlist\Nnod{3,1} 
			\message{^^J  Layer}
			\foreachitem \N \in \Nnod{ 
				\edef\lay{\Ncnt} 
				\message{\lay,}
				\pgfmathsetmacro\prev{int(\Ncnt-1)} 
				\foreach \i [evaluate={\y=\N/2-\i; \x=\lay; \n=\nstyle;}] in {1,...,\N}{ 
					\node[node \n] (N\lay-\i) at (\x,\y) {$y_\i^{[\prev]}$};
					\ifnum\lay>1
					\foreach \j in {1,...,\Nnod[\prev]}{ 
						\draw[connect arrow] (N\prev-\j) -- (N\lay-\i); 
					}
					\fi 
				}
			}
			\node[left of =5,align=center,mygreen!60!black] at (0.9,-0.5) {input\\[-0.2em]layer};
			\node[above=8,align=center,myred!60!black] at (N\Nnodlen-1.90) {output\\[-0.2em]layer};	
		\end{tikzpicture}
	\end{center}
	\end{minipage} \hfill
	\begin{minipage}[c]{0.6\textwidth}
		\caption{Schematic representation of the perceptron with a three dimensional input. For generality we denote the input by $y^{[0]} = u$ and the output by $y^{[1]} = y$. The weights $W_i$ are applied on the arrows and the bias is added in the node $y^{[1]}_1$.}
		\label{fig:Perceptron}
	\end{minipage}
\end{figure}

The function $\sigma:\mathbb{R} \rightarrow \mathbb{R}$ is called \textbf{activation function}. Originally, in \cite{rosenblatt1958perceptron}, it was proposed to choose the Heaviside function as activation function to model whether a neuron fires or not, i.e.
\begin{equation*}
	\sigma(y) = \begin{cases}
		1 &\text{if } y\geq 0, \\
		0 &\text{if } y <0.
	\end{cases}
\end{equation*}
However, over time several other activation functions have been suggested and are being used. Typically, they are monotone increasing to remain in the spirit of the original idea, but continuous. 

Popular activation functions are, cf. \cite[p.90]{ng2022}
\begin{align*}
	\sigma(y) &= \frac{1}{1+ \exp(-y)} &&\text{sigmoid (logistic)}  ,\\
	\sigma(y) &= \tanh(y) = \frac{\exp(y)-\exp(-y)}{\exp(y)+\exp(-y)} &&\text{hyperbolic tangent} ,\\
	\sigma(y) &= \max\{y,0\} &&\text{rectified linear unit (ReLU)} ,\\
	\sigma(y) &= \max\{\alpha y,y\} &&\text{leaky ReLU} .
\end{align*}

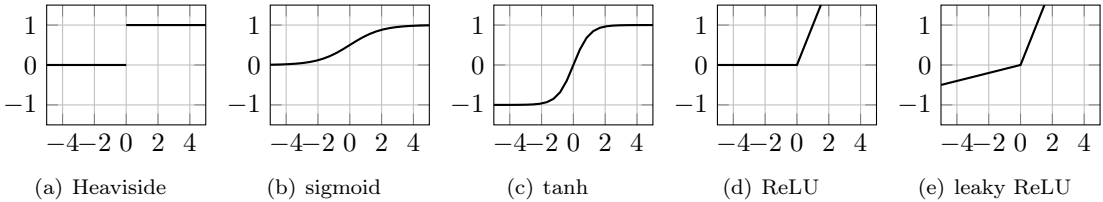
\begin{figure}[h]
\hspace{-0.5cm}
\subfigure[Heaviside]{\label{fig:Heaviside}
\begin{tikzpicture}
	\begin{axis}[xmin = -5, xmax = 5, ymin = -1.5, ymax = 1.5,grid=both,width=0.25\textwidth]
		\addplot[thick,samples at={-5,0}] {0};
		\addplot[thick,samples at={0,5}] {1};
	\end{axis}
\end{tikzpicture}
}\hspace{-0.2cm}
\subfigure[sigmoid]{\label{fig:sigmoid}	
\begin{tikzpicture}
	\begin{axis}[xmin = -5, xmax = 5, ymin = -1.5, ymax = 1.5,grid=both,width=0.25\textwidth]
		\addplot[thick] {1/(1+exp(-x))};
	\end{axis}
\end{tikzpicture}
}\hspace{-0.2cm}
\subfigure[tanh]{\label{fig:tanh}
\begin{tikzpicture}
	\begin{axis}[xmin = -5, xmax = 5, ymin = -1.5, ymax = 1.5,grid=both,width=0.25\textwidth]
		\addplot[thick] {(exp(x)-exp(-x))/(exp(x)+exp(-x))};
	\end{axis}
\end{tikzpicture}
}\hspace{-0.2cm}
\subfigure[ReLU]{\label{fig:relu}
\begin{tikzpicture}
		\begin{axis}[xmin = -5, xmax = 5, ymin = -1.5, ymax = 1.5,grid=both,width=0.25\textwidth]
		\addplot[thick] {max(0,x)};
	\end{axis}
\end{tikzpicture}
}\hspace{-0.2cm}
\subfigure[leaky ReLU]{\label{fig:leakyrelu}
\begin{tikzpicture}
	\begin{axis}[xmin = -5, xmax = 5, ymin = -1.5, ymax = 1.5,grid=both,width=0.25\textwidth]
		\addplot[thick] {max(0.1*x,x)};
	\end{axis}
\end{tikzpicture}
}
\caption{Popular activation functions. Leaky ReLU is displayed for $\alpha = 0.1$.}
\label{fig:activationfunctions}
\end{figure}

\begin{remark} \label{rem:nonlin}
	The nonlinearity of activation functions is an integral part of the Neural Networks success. Since concatenations of linear functions result again in a linear function, see e.g. \cite[p.90]{ng2022}, the complexity that can be achieved by using linear activation functions is limited. 
\end{remark}

While the sigmoid function approximates the Heaviside function continuously, and is differentiable, it contains an exponential operation, which is computationally expensive. Similar problems arise with the hyperbolic tangent function. However, the fact that $\tanh$ is closer to the identity function often helps speed up convergence, since it resembles a linear model, as long as the values are close to zero.
Another challenge that needs to be overcome is vanishing derivatives, which is visibly present for Heaviside, sigmoid and hyperbolic tangent. In contrast, ReLU is not bounded on positive values, while also being comparatively cheap to compute, because linear computations tend to be very well optimized in modern computing. Altogether, these advantages have resulted in ReLU (and variants thereof) becoming the most widely used activation function currently. As a remedy for the vanishing gradient on negative values, leaky ReLU was introduced. When taking derivatives of ReLU one needs to account for the non-differentiability at 0, but in numerical practice this is easily overcome.

With the help of Neural Networks we want to solve a task, cf. \cite[Section 5.1]{goodfellow2016deep}. Let the performance of the algorithm for the given task be measured by the \textbf{loss function} $L$, which needs to be adequately modeled. By $\mathcal{F}$ we denote the Neural Network. The variables that will be learned are the weights $W$ and biases $b$ of the Neural Network. Hence, we can formulate the following optimization problem, cf. \cite{antil2022deep,antildiazherberg,antil2020fractional}
\begin{equation}\label{eq:LP} \tag{$P$}
	\min_{W,b} \mathscr{L}(y,u,W,b)  \qquad \text{s.t.}\qquad  y = \mathcal{F}(u,W,b).
\end{equation}
One possible choice for $\mathcal{F}$ has already been given in \eqref{eq:perceptron}, the perceptron. In the subsequent sections we introduce and analyze various other Neural Network architectures. They all have in common that they contain weights and biases, so that the above problem formulation remains sensible.

Before we move on to different network architectures, we discuss the modeling of the loss function. Learning tasks can be divided into two subgroups: Supervised and Unsupervised learning. 

\subsection{Supervised Learning}
\label{subsec:Supervised}
In supervised learning we have given data $u$ with known supervision $S(u)$ (also called labels), so that the task is to match the output $y$ of the Neural Network to the supervision. These problems are further categorized depending on the known supervision, e.g. for $S(u) \in \mathbb{N}$ it is called a classification and for $S(u) \in \mathbb{R}$ a regression. Furthermore, the supervision $S(u)$ can also take more complex forms like a black and white picture of $256 \times 256$ pixels represented by $[0,1]^{256}$, a higher dimensional quantity, a sentence, etc. These cases are called structured output learning. 

Let us consider one very simple example, cf. \cite[Section 5.1.4]{goodfellow2016deep}.

\begin{example}{Linear Regression}\\ \label{ex:LR}
We have a given set of inputs $u^{(i)} \in \mathbb{R}^d$ with known supervisions $S(u^{(i)}) \in \mathbb{R}$ for $i=1,\ldots,N$. In this example we only consider weights $W \in \mathbb{R}^{d}$ and no bias. Additionally, let $\sigma = \id$. The perceptron network simplifies to 
\begin{equation*}
	y^{(i)} = W^{\top} u^{(i)},
\end{equation*} 
and the learning task is to find $W$, such that $y^{(i)} \approx S(u^{(i)})$. This can be modeled by the \textbf{mean squared error (MSE)} function
\begin{equation*}
	\mathscr{L}(\{y^{(i)}\}_i,\{u^{(i)}\}_i,W) := \frac{1}{2N} \sum_{i=1}^N \| y^{(i)} - S(u^{(i)}) \|^2.
\end{equation*}
By convention we will use $\| \cdot \| = \| \cdot \|_2$ throughout the lecture.
The chosen loss function is quadratic, convex and non-negative. We define 
\begin{equation*}
	U := \begin{pmatrix}
		(u^{(1)})^{\top} \\ \vdots \\ (u^{(N)})^{\top}
	\end{pmatrix} \in \mathbb{R}^{N\times d}, \qquad 
	S:= \begin{pmatrix}
		S(u^{(1)}) \\ \vdots \\ S(u^{(N)})
	\end{pmatrix} \in \mathbb{R}^{N},
\end{equation*}
so that we can write $\mathscr{L}(W) = \frac{1}{2} \| U W - S\|_2^2 $. Minimizing this function will deliver the same optimal weight $W$ as minimizing the MSE function defined above. 
We can now derive the gradient
\begin{equation*}
	\nabla_W \mathscr{L}(W) = U^{\top} U W - U^{\top} S 
\end{equation*}
and immediately find the stationary point $W = (U^{\top} U)^{-1} U^{\top} S$.
\end{example}

\subsection{Unsupervised Learning}
In unsupervised learning, only the input data $u$ is given and we have no knowledge of supervisions or labels. The algorithm is supposed to learn e.g. a structure or relation in the data. Some examples are k-clustering and principal component analysis (PCA). Modeling the loss function specifies the task and has a direct influence on the learning process. For illustration of this concept, we introduce the k-means algorithm, see eg. \cite[Chapter 10]{ng2022}, which is used for clustering.

\begin{figure}[h!]
	\centering
	\includegraphics[ width=0.85\textwidth]{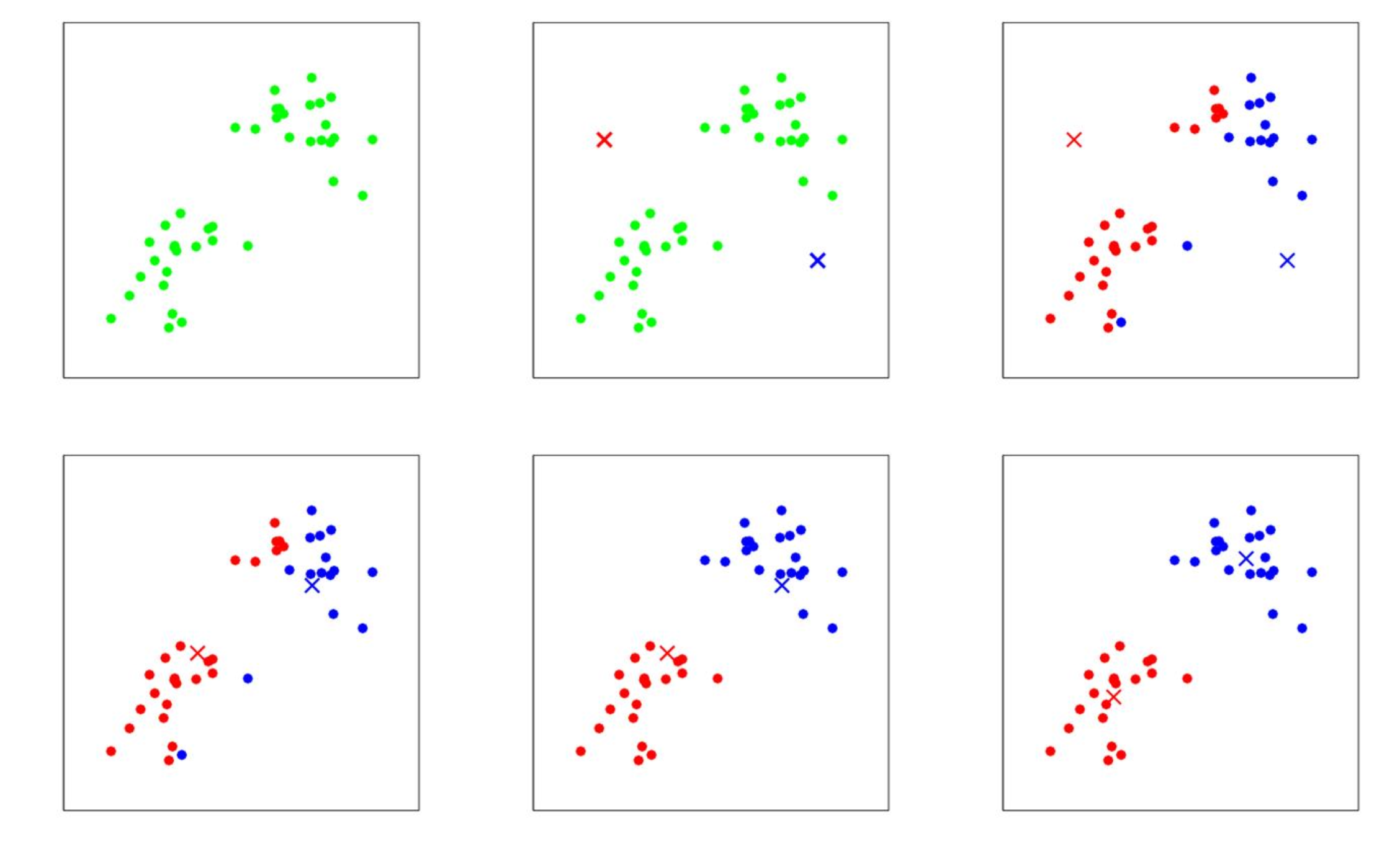}
	\caption{Visualization of k-means algorithm for $k=2$ clusters. Data points $u^{(i)}$ are indicated as dots, while cluster centroids $\mu_j$ are shown as crosses. Top, from left to right: Dataset, random initial cluster centroids $\mu_1$ (red) and $\mu_2$ (blue), every data point is assigned to either the red or blue cluster. Bottom, from left to right: cluster centroids are redefined, every data point is reassigned, cluster centroids are redefined again.
		Image source: \cite[Chapter 10]{ng2022}.}
	\label{fig:kmeans}
\end{figure}

\newpage
\begin{example}
	We have a set of given data points
	\begin{equation*}
		\left\{ u^{(i)}\right\}_{i=1}^N \in \mathbb{R}^{ d},
	\end{equation*}
	and a desired number of clusters $k \in \mathbb{N}$ with $k \leq N$ and typically $k \ll N$. Every data point is supposed to be assigned to a cluster.	
	Iteratively every data point is assigned to the cluster with the nearest centroid, and we redefine cluster centroids as the mean of the vectors in the cluster. The procedure is specified in Algorithm \ref{alg:kmeans} and illustrated for an example in Figure \ref{fig:kmeans}, which can be found e.g. in \cite[Chapter 10]{ng2022}. The loss function (also called distortion function in this setup) can be defined as \[\mathscr{L}(c,\mu):=\sum_{i=1}^{N}\|u^{(i)}-\mu_{c^{(i)}}\|^2,\]
	which is also a model of the quantity that we try to minimize in Algorithm \ref{alg:kmeans}. We have a non-convex set of points in $\mathbb{R}^d$, so the algorithm may converge to a local minimum. To prevent this, we run the algorithm many times, compare the resulting clusterings using the loss function, and choose the one with the minimal value attained in the loss function.
\end{example}

\begin{algorithm}[h!]	
	\caption{k-means clustering}
	\begin{algorithmic}
		\Require Initial cluster centroids $\mu_1,\ldots,\mu_k$
		\While{not converged}
		\For{$i=1:N$}
		\State $c^{(i)} := \arg\min_{j} \|u^{(i)}-\mu_j\|^2$
		\EndFor
		\For{$j=1:k$}
		\State $\mu_j \gets \frac{\sum_{i=1}^N 1_{\{c^{(i)} = j\}}u^{(i)}}{\sum_{i=1}^N 1_{\{c^{(i)}=j\}}}$
		\EndFor
		\EndWhile
	\end{algorithmic}
	\label{alg:kmeans}
\end{algorithm}

We will see various other loss functions $\mathscr{L}$ throughout the remainder of this lecture, all of them specifically tailored to the task at hand.

In the case of Linear Regression, we have a closed form derivative, so we are able to find the solution by direct calculus, while for k-means clustering the optimization was done by a tailored iteration. For general problems we will need a suitable optimization algorithm. We move on to introduce a few options.

\subsection{Optimization Algorithms} \label{subsec:alg}
Here, for simplicity we define $\theta$, which collects all variables, i.e. weights $W$ and bias $b$ and write the loss function as 
\[\mathscr{L}(\theta) = \frac{1}{N} \sum_{i=1}^N \mathscr{L}^{(i)}(\theta),\] 
which we want to minimize. Here, $\mathscr{L}^{(i)}$ indicates the loss function evaluated for data point $i$, for example with a MSE loss $\mathscr{L}^{(i)}(\theta) = \frac{1}{2} \| y^{(i)} - S(u^{(i)}) \|^2$.

First, let us recall the standard \textbf{gradient descent} algorithm, see e.g. \cite[Section 9.3]{boyd2004convex}, which is also known as steepest descent or batch gradient descent. 
\begin{algorithm}
	\caption{Gradient Descent}
	\begin{algorithmic}
		\Require Initial point $\theta^0$, step size $\tau>0$, counter $k=0$.
		\While{Stopping criterion not fulfilled}
		\State $\theta^{k+1} = \theta^k - \tau \cdot \nabla \mathscr{L}(\theta^k)$,
		\State $k \gets k+1$.
		\EndWhile
	\end{algorithmic}
	\label{alg:gd}
\end{algorithm}

Possible stopping criterion are e.g. setting a maximum number of iterations $k$, reaching a certain exactness $\|\mathscr{L}(\theta)\| < \epsilon $ with a small number $\epsilon>0$, or a decay in change $\| \theta^{k+1}-\theta^k\| < \epsilon$. Determining a suitable step size is integral to the success of the gradient descent method, especially since this algorithm uses the same step size $\tau$ for all components of $\theta$, which can be a large vector in applications. If may happen that in some components the computed descent direction is only providing descent in a small neighborhood, therefore requiring a small step size $\tau$.
It is also possible to employ a line search algorithm. However, this is not common in Machine Learning currently. Instead, typically a small step size is chosen, so that it will (hopefully) be not too large for any component of $\theta$, and then it may be adaptively increased. Furthermore, let us remark that the step size is often called \textbf{learning rate} in a Machine Learning context. 

Additionally, a grand challenge in Machine Learning tasks is that we have huge data sets, and the gradient descent algorithm has to iterate over all data points in every iteration, since $\mathscr{L}(\theta)$ contains all data points, which causes a tremendous computational cost. This motivates the use of the \textbf{stochastic gradient descent} algorithm, cf. \cite[Algorithm 1]{ng2022}, which only takes one data point into account per iteration. 

 \begin{algorithm}
 	\caption{Stochastic Gradient Descent (SGD)}
 	\begin{algorithmic}
 		\Require Initial point $\theta^0$, step size $\tau>0$, counter $k=0$, maximum number of iterations $K$.
 		\While{$k \leq K$}
 		\State Sample $j \in \{1,\ldots,N\}$ uniformly.
 		\State $\theta^{k+1} = \theta^k - \tau \cdot \nabla \mathscr{L}^{(j)}(\theta^k)$,
 		\State $k \gets k+1$.
 		\EndWhile
 	\end{algorithmic}
 	\label{alg:sgd}
 \end{algorithm}

Since the stochastic gradient descent method only calculates the gradient for one data point, it produces an irregular convergence behavior. Indeed, it does not necessarily converge at all, but for a large number of iterations $K$ it often produces a good approximation. In fact, actually converging in training the Neural Network is often not necessary/desired anyhow, since we want to have a solution that generalizes well to unseen data, rather than fit the given data points perfectly. Actually, the latter may lead to overfitting, cf. Section \ref{subsec:Overfit}.
Therefore, SGD is a computationally cheap, reasonable alternative to gradient descent. 
As a compromise, which generates a less irregular convergence behavior, there also exists \textbf{mini batch gradient descent}, cf. \cite[Algorithm 2]{ng2022}, where every iteration takes into account a subset (mini batch) of the data points. 

\begin{algorithm}
	\caption{Mini Batch Gradient Descent}
	\begin{algorithmic}
		\Require Initial point $\theta^0$, step size $\tau>0$, counter $k=0$, maximum number of iterations $K$, batch size $b\in \mathbb{N}$.
		\While{$k \leq K$}
		\State Sample $b$ examples $j_1,\ldots,j_b$ uniformly from $\{1,\ldots,N\}$
		\State $\theta^{k+1} = \theta^k - \tau \cdot \frac{1}{b} \sum_{i=1}^{b} \nabla \mathscr{L}^{(j_i)}(\theta^k)$,
		\State $k \gets k+1$.
		\EndWhile
	\end{algorithmic}
	\label{alg:mbgd}
\end{algorithm}

Finally, we introduce a sophisticated algorithm for stochastic optimization called \textbf{Adam}, \cite{adam}, see Algorithm \ref{alg:adam}. It is also a gradient-based method, and as an extension of the previous methods it employs adaptive estimates of so-called moments. 
\begin{algorithm}
	\caption{Adam. All operations on vectors are element-wise. $(g^k)^2$ indicates the element-wise square $g^k \odot g^k$, and $(\beta_1)^k, (\beta_2)^k$ denote the $k$-th power of $\beta_1$ and $\beta_2$, respectively.}
	\begin{algorithmic}
		\Require Initial point $\theta^0$, step size $\tau>0$, counter $k=0$, exponential decay rates for the moment estimates $\beta_1,\beta_2 \in [0,1)$, $\epsilon > 0$, stochastic approximation $\widetilde{\mathscr{L}}(\theta)$ of the loss function.
		\State $m_1^0 \gets 0$ (Initialize first moment vector)
		\State $m_2^0 \gets 0$ (Initialize second moment vector)
		\While{$\theta^k$ not converged}
			\State $g^{k+1} = \nabla_\theta \widetilde{\mathscr{L}}(\theta^{k})$
			\State $m_1^{k+1} = \beta_1 \cdot m_1^k + (1-\beta_1) \cdot g^{k+1}$
			\State $m_2^{k+1} = \beta_2 \cdot m_2^k + (1-\beta_2) \cdot (g^{k+1})^2$
			\State $m_1^{k+1} \gets \frac{m_1^{k+1}}{(1-(\beta_1)^k)}$
			\State $m_2^{k+1} \gets \frac{m_2^{k+1}}{(1-(\beta_2)^k)}$
			\State $\theta^{k+1} = \theta^k - \tau \cdot \frac{m_1^{k+1}}{ \left( \sqrt{m_2^{k+1}} + \epsilon \right)} $
			\State $k \gets k+1$
		\EndWhile 
	\end{algorithmic}
	\label{alg:adam}
\end{algorithm}
Good default settings in Adam for the tested machine learning problems are $\tau = 0.001$, $\beta_1 = 0.9, \beta_2 = 0.999$ and $\epsilon = 10^{-8}$, cf. \cite{adam}. Typically, the stochasticity of $\widetilde{\mathscr{L}}(\theta)$ will come from using mini batches of the data set, as in Mini Batch Gradient Descent, Algorithm \ref{alg:mbgd}. 

\begin{figure}[h!]	
\begin{minipage}{0.63\textwidth}
	\begin{remark} \label{rem:stat}
		In any case we need to be cautious when interpreting results, since independent of the chosen algorithm, we are dealing with a non-convex loss function, so that we can only expect convergence to stationary points. 
	\end{remark}
\caption{Simple example of a non-convex loss function with a local and a global minimum.}
\end{minipage}
\hspace{0.2cm}
\begin{minipage}{0.37\textwidth}
	\centering
	\includegraphics[width=\textwidth]{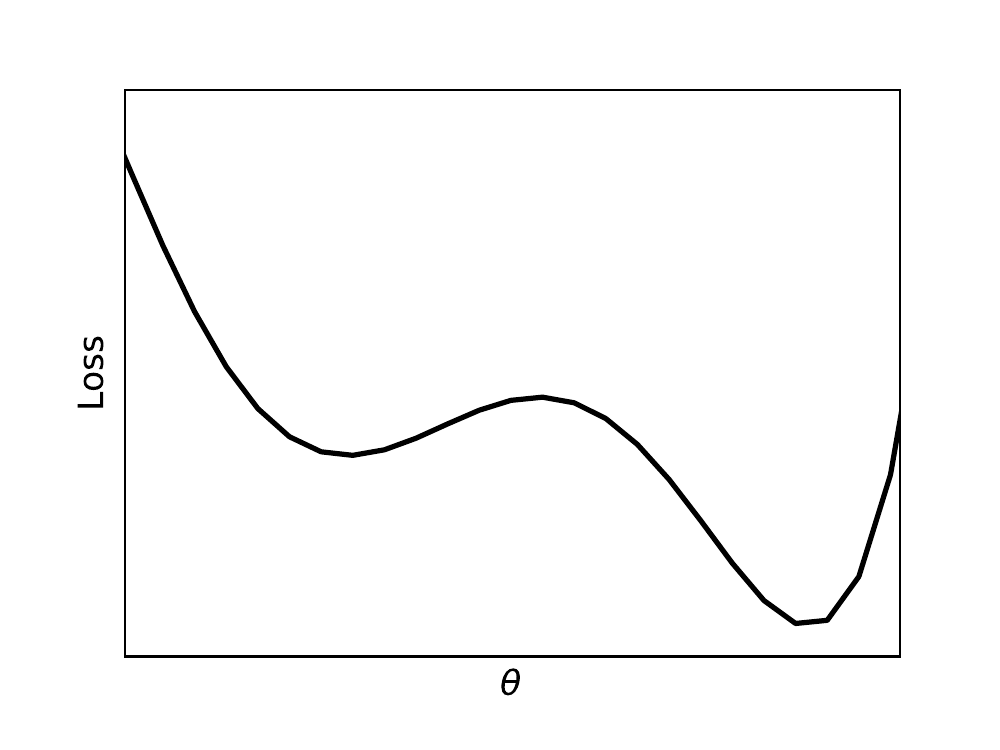}
	\label{fig:local}
\end{minipage}
\end{figure}

\newpage
In the following section we discuss how fitting the given data points and generalizing well to unseen data can be contradictory goals. 

\subsection{Overfitting and Underfitting}\label{subsec:Overfit}
As an example we discuss supervised learning with polynomials of degree $r$, cf. \cite[Section 1.3.3]{geiger2021DL}.
\begin{example}
Define
 \[p(u,W):=\sum_{j=0}^r W_j u^j= W^\top u, \] 
with $u=(u^0,...,u^r)^\top \in \mathbb{R}^{r+1}$ the potencies of data point $u$, and $W:=(W_0,...,W_r)^\top \in \mathbb{R}^{r+1}$.
The polynomial $p$ is linear in $W$, but not in $u$. As in Linear Regression (Example \ref{ex:LR}), we do not consider bias $b$ here. 
Our goal is to compute weights $W$, given data points $u^{(i)}$ with supervisions $S(u^{(i)})$, so that $p$ makes good predictions on data it hasn't seen before. We again employ the MSE loss function  
\[\mathscr{L}(W)=\frac{1}{2N}\sum_{i=1}^N \| p(u^{(i)},W)-S(u^{(i)}) \|^2\] 
As before, we write the loss in matrix-vector notation 
\[ \mathscr{L}(W)=\frac{1}{2N}\|U W - S\|^2\] 
where \[U:=\begin{pmatrix}
	u^{(1)}_0&u^{(1)}_1& \ldots &u^{(1)}_r\\
	\vdots& \vdots &  & \vdots\\
	u^{(m)}_0&u^{(m)}_1& \ldots &u^{(m)}_r
\end{pmatrix},\  
S:=\begin{pmatrix}
	S(u^{(1)})\\
	\vdots\\
	S(u^{(m)}).
\end{pmatrix}\]
The minimizer $W$ can be directly calculated, cf. Example \ref{ex:LR}.
\end{example}

\begin{figure}[h!]
	\centering
	\includegraphics[ width=0.99\textwidth]{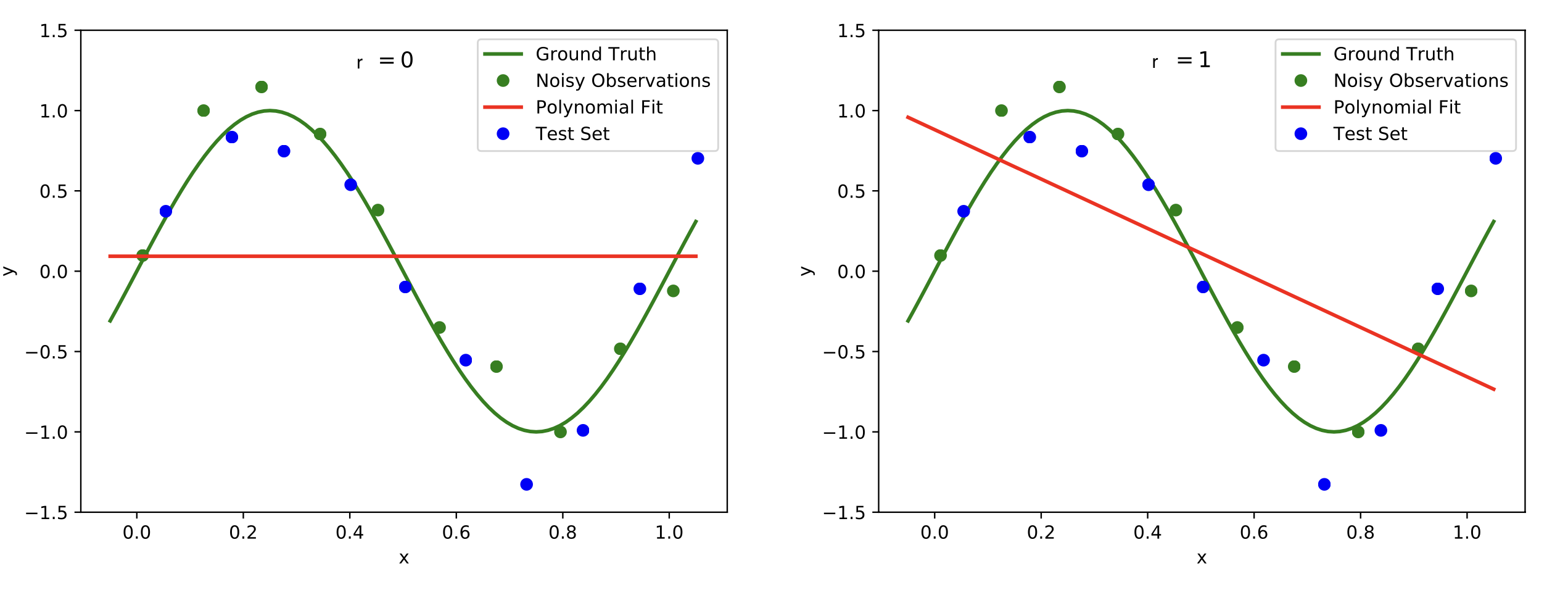}\\
	\hspace{-0.3cm}\includegraphics[ width=0.98\textwidth]{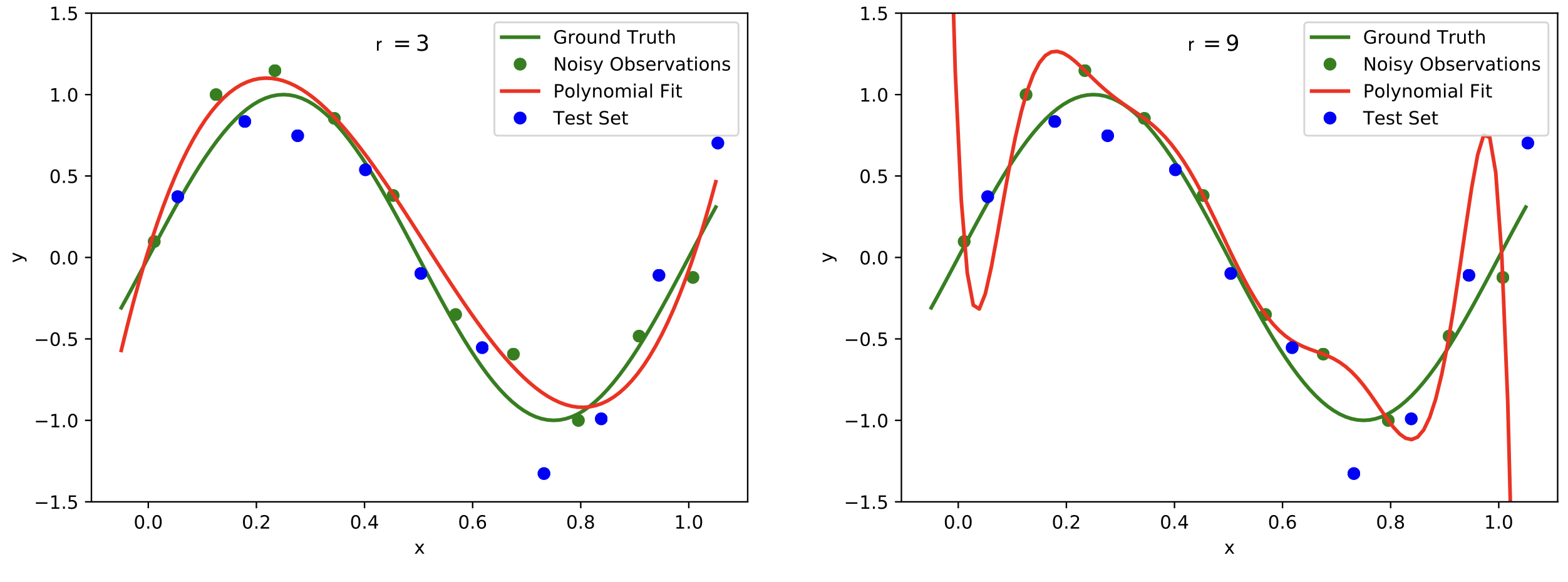}
	\caption{Plots of polynomials of various degrees r (red graph) fitted to the noisy data points (green dots) based on the ground truth (green graph). The model should extend well to the test set data (blue dots). We observe underfitting in the top row for $r=0$ (left) and $r=1$ (right). In the bottom left with $r=3$ reasonable results are achieved, while $r=9$ in the bottom right leads to overfitting. Image modified from: \cite[Fig. 1]{geiger2021DL}.}
	\label{fig:polynomials}
\end{figure}

To measure the performance of the polynomial curve fitting we compute the error on data points that were not used to determine the best polynomial fit, because we aim for a model that will generalize well. To this end, finding a suitable degree for the polynomial that we are fitting over the data points is crucial. If the degree is too low, we will encounter \textbf{underfitting}, see Figure \ref{fig:polynomials} top row. This means that the complexity of the polynomial is too low and the model does not even fit the data points. A remedy is to increase the degree of the polynomial, see Figure \ref{fig:polynomials} bottom left. However, increasing the degree too much may lead to \textbf{overfitting}, see Figure \ref{fig:polynomials} bottom right. The data points are fit perfectly, but the curve will not generalize well.

\begin{figure}[h!]	
	\begin{minipage}{0.7\textwidth}
\centering
\includegraphics[ width=\textwidth]{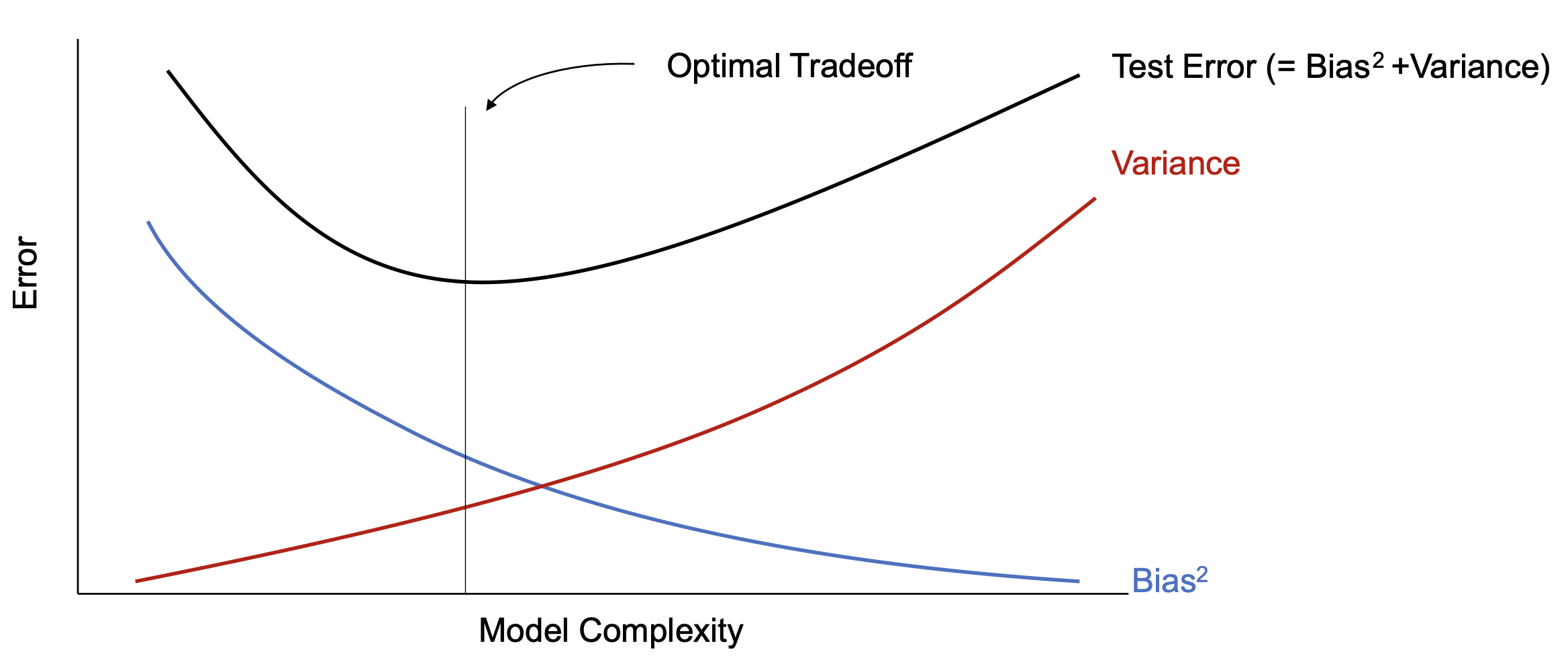}
	\end{minipage}
\hspace{0.2cm}
	\begin{minipage}{0.27\textwidth}
	\caption{Bias-variance trade-off. Image source: \cite[Fig. 8.8]{ng2022}.}
\label{fig:biasvariance}
	\end{minipage}
\end{figure}

We can characterize overfitting and underfitting by using some statistics, cf. \cite[Section 8.1]{ng2022}. A point estimator $g:\mathcal{U}^N \rightarrow \Theta$ (where $\mathcal{U}$ denotes the data space, and $\Theta$ denotes the parameter space) is a function which makes an estimation of the underlying parameters of the model. For example, the estimate for $\theta=W$ from Example \ref{ex:LR}: $\hat{\theta}=(U^\top U)^{-1}U^\top S$ (which we will denote with a hat in this subsection to emphasize that it is an estimation) is an example of a point estimator. We assume that the data from $\mathcal{U}^N$ is i.i.d, so that $\hat{\theta}$ is a random variable.
We can define the variance and the bias \[\text{Var}(\hat{\theta}):=\mathbb{E}(\hat{\theta}^2)-\mathbb{E}(\hat{\theta})^2,\ \text{Bias}(\hat{\theta}):=\mathbb{E}(\hat{\theta})-\theta,\]
with $\mathbb{E}$ denoting the expected value.
A good estimator has both, low variance and low bias. We can characterize overfitting with low bias and high variance, and underfitting with high bias and low variance. The bias-variance trade-off is illustrated in Figure \ref{fig:biasvariance}. Hence, we can make a decision based on mean squared error of the estimates
\[\text{MSE}(\hat{\theta}):=\mathbb{E}[(\hat{\theta}-{\theta})^2] =\text{Var}(\hat{\theta})+\text{Bias}(\hat{\theta})^2.\]

In general, it can be hard to guess a suitable degree for the polynomial beforehand. We could compute a fitting curve for different choices of $r$ and then compare the error on previously unseen data points of the validation data set, cf. Section \ref{subsec:Hyper}, to determine which one generalizes best. This will require solving the problem multiple times which is unfavorable, especially for large data sets. Also, the polynomial degree can only be set discretely. 
Another, continuous way is to introduce a penalization term in the loss function
\[\mathscr{L}_\lambda(\theta) := \mathscr{L}(\theta) + \lambda \|\theta \|^2 .\] 
This technique is also called \textbf{weight decay}, cf. \cite[Section 5.2.2]{goodfellow2016deep}. We can also use other norms, e.g. $\| \cdot \|_1$. Here, we can choose a large degree $r$ and for $\lambda$ big enough, we will still avoid overfitting, because many components of $\theta$ will be (close to) zero. Nonetheless, we need to be cautious with the choice of $\lambda$. If it is too big, we will face again the problem of underfitting. 

We see that choosing values for the degree $r$ and the penalization parameter $\lambda$ poses challenges, and will discuss this further in the next section.

\subsection{Hyperparameters and Data Set Splitting} \label{subsec:Hyper}
We call all quantities that need to be chosen before solving the optimization problem \textbf{hyperparameters}, cf. \cite[Section 5.3]{goodfellow2016deep}. Let us point out that hyperparameters are not learnt by the optimization algorithm itself, but nevertheless have an impact on the algorithms performance.
Examples of hyperparameters include the polynomial degree $r$, the scalar $\lambda$, all parameters in the optimization algorithms (Section \ref{subsec:alg}) like the step size $\tau$, and also the architecture of the Neural Network, and many more.

The impact of having a good set of hyperparameters can be tremendous, however finding such a set is not trivial. First of all, we split our given data into three sets. \textbf{training} data, \textbf{validation} data and \textbf{test} data (a 4:1:1 ratio is common). We have seen training and test data before. The data points that we are using as input to solve the optimization problem are called training data, and the unseen data points, which we use to evaluate whether the model generalizes well, are called test data. Since we don't want to mix different causes of error, we also introduce the validation data set. This will be used to compare different choices of hyperparameter configurations, i.e. we train the model on the training data for different hyperparameters, compute the error on the validation data set, choose the hyperparameter setup with the lowest error and finally evaluate the model on the test set. The reasoning behind this is that if we would use the test data set to determine the hyperparameter values, the test error may be not meaningful, because the hyperparameters have been optimized for this specific test set. Since we are using the validation set, we will have the test set with previously unseen data available to determine the generalization error without giving our network an advantage. 

Still, imagine you need to choose 5 hyperparameters and have 4 possible values that you want to try for each hyperparameter. This amounts to $4^5 = 1024$ 
combinations you have to run on the training data and evaluate on the validation set. In real applications the number of hyperparameters and possible values can be much larger, so that it is nearly infeasible to try every combination, but rather common to change one hyperparameter at a time. Luckily, some hyperparameters also have known good default values, like the hyperparameters for Adam Optimizer, Algorithm \ref{alg:adam}. Apart from that it is a tedious, manual work to try out, monitor and choose suitable hyperparameters.

Finally, we discuss the limitations of shallow Neural Networks, i.e. networks with only one layer. 

\subsection{Modeling logical functions}
Let us consider a shallow Neural Network with input layer $y^{[0]} \in \mathbb{N}^2$ and output layer $y^{[1]} \in \mathbb{N}$. 
\begin{figure}[h]
	\begin{minipage}[c]{0.4\textwidth}
		\begin{center}
			\begin{tikzpicture}[x=2.2cm,y=1.4cm]
				\message{^^JNeural network with arrows}
				\readlist\Nnod{2,1} 
				\message{^^J  Layer}
				\foreachitem \N \in \Nnod{ 
					\edef\lay{\Ncnt} 
					\message{\lay,}
					\pgfmathsetmacro\prev{int(\Ncnt-1)} 
					\foreach \i [evaluate={\y=\N/2-\i; \x=\lay; \n=\nstyle;}] in {1,...,\N}{ 
						\node[node \n] (N\lay-\i) at (\x,\y) {$y_\i^{[\prev]}$};
						\ifnum\lay>1
						\foreach \j in {1,...,\Nnod[\prev]}{ 
							\draw[connect arrow] (N\prev-\j) -- (N\lay-\i); 
						}
						\fi 
					}
				}
				\node[left of =5,align=center,mygreen!60!black] at (0.9,-0.5) {input\\[-0.2em]layer};
				\node[above=8,align=center,myred!60!black] at (N\Nnodlen-1.90) {output\\[-0.2em]layer};	
			\end{tikzpicture}
		\end{center}
	\end{minipage} \hfill
	\begin{minipage}[c]{0.6\textwidth}
		\caption{A simple perceptron (shallow Neural Network) with a two dimensional input.}
		\label{fig:Perceptron2}
	\end{minipage}
\end{figure}
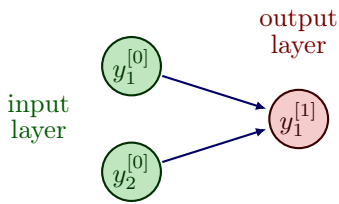

We model true by the value 1 and false by the value 0, which results in the following truth table for the logical "OR" function. 
\begin{table}[h]
	\centering
	\begin{tabular}{| c | c | c |}
		\hline
		input $y^{[0]}_1$ & input $y^{[0]}_2$ & $y^{[0]}_1$ OR $y^{[0]}_2$ (output $y^{[1]}_1$)\\
		\hline 
		0 & 0 & 0 \\
		\hline 
		1 & 0 & 1 \\
		\hline 
		0 & 1 & 1 \\
		\hline 
		1 & 1 & 1 \\
		\hline 
	\end{tabular}
\label{tab:OR}
\caption{Truth table for the logical "OR" function.}
\end{table}

With Heaviside activation function, we have 
\[y_1^{[1]} = \begin{cases} 
	1, &\text{if } W_1 y^{[0]}_1 + W_2 y^{[0]}_2 + b \geq 0,\\
	0, &\text{else }.
\end{cases}\]
The goal is now to choose $W_1,W_2,b$ so that we match the output from the truth table for given input. Obviously, $W_1 = W_2 = 1$ and $b=-1$ is a possible choice that fulfills the task. Similarly, one can find values for $W_1,W_2$ and $b$ to model the logical "AND" function.

Next, let us consider the logical "XOR" function with the following truth table.  
\begin{table}[h]
	\centering
	\begin{tabular}{| c | c | c |}
		\hline
		input $y^{[0]}_1$ & input $y^{[0]}_2$ & $y^{[0]}_1$ XOR $y^{[0]}_2$ (output $y^{[1]}_1$)\\
		\hline 
		0 & 0 & 0 \\
		\hline 
		1 & 0 & 1 \\
		\hline 
		0 & 1 & 1 \\
		\hline 
		1 & 1 & 0 \\
		\hline 
	\end{tabular}
	\label{tab:XOR}
	\caption{Truth table for the logical "XOR" function.}
\end{table}

In fact, the logical "XOR" function can not be represented by the given shallow Neural Network, since the data is not linearly separable, see e.g. \cite[Section 6.1]{goodfellow2016deep}. This motivates the introduction of additional layers in between the input and output layer, i.e. we choose a more complex function $\mathcal{F}$ in the learning problem \eqref{eq:LP}.

\begin{figure}[h]
	\centering
	\hspace{0.3cm}
	\includegraphics[width=0.6\textwidth]{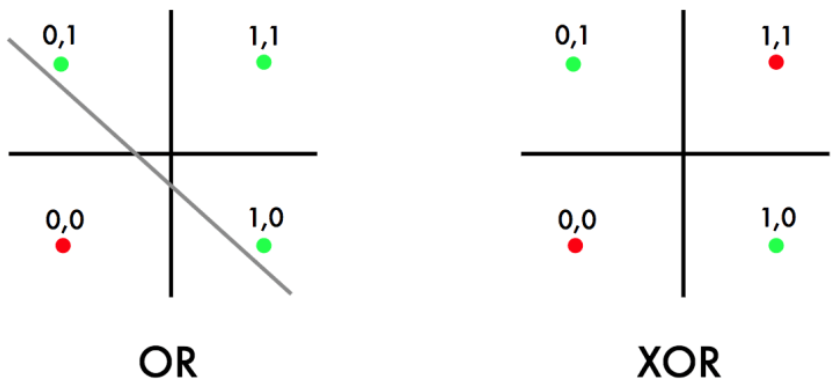}
	\label{fig:XORlinear}
	\caption{This illustration shows that the logical "OR" function is linearly separable, while the logical "XOR" function is not. Image modified from: \url{https://dev.to/jbahire/demystifying-the-xor-problem-1blk}.}
\end{figure}

\newpage
\section{Feedforward Neural Network}
\label{sec:FNN}
Introducing \textbf{hidden layers}, i.e. layers between the input and output layer, leads to \textbf{Feedforward Neural Networks (FNNs)}, also called \textbf{multilayer perceptrons (MLPs)}, cf. \cite[Section 6]{goodfellow2016deep}. Essentially, they are multiple perceptrons organized in layers, $\ell = 0, \ldots, L$, where every perceptron takes the output from the previous perceptron as input. The number of layers $L$ is called the \textbf{depth} of the network, while the number of neurons per layer $n_\ell$ is the \textbf{width} of the network. The input layer is denoted with $y^{[0]} = u \in \mathbb{R}^{n_0}$ and not counted in the depth of the network. A FNN is called \textbf{deep} if it has at least two hidden layers. We now indicate the weights from layer $\ell$ to $\ell+1$ by $W^{[\ell]} \in \mathbb{R}^{n_{\ell + 1} \times n_\ell}$ and the bias vector by $b^{[\ell]} \in \mathbb{R}^{n_{\ell+1}}$ for $\ell=0,\ldots,L-1$. To simplify notation, we extend the activation function $\sigma:\mathbb{R}\rightarrow \mathbb{R}$ to vector valued inputs, by applying it component-wise, so that $\sigma:\mathbb{R}^n \rightarrow \mathbb{R}^n$, with $(y_1,\ldots,y_n)^{\top} \mapsto (\sigma(y_1),\ldots,\sigma(y_n))^\top$. The FNN layers can be represented in the same way as perceptrons 
\begin{equation} \label{eq:FNNarchitecture}
	y^{[\ell]} = f_\ell (y^{[\ell-1]}) = \sigma^{[\ell]} ( W^{[\ell -1]} y^{[\ell-1]} + b^{[\ell-1]} ) \qquad \text{for } \ell=1,\ldots,L, 
\end{equation}
where the activation function $\sigma^{[\ell]}$ may differ from layer to layer. We call $y^{[\ell]}$ the \textbf{feature vector} of layer $\ell$.
Compactly, we can write a FNN as a composition of its layer functions, cf. \cite{antil2022deep,antildiazherberg}
\[ y^{[L]} = \mathcal{F}(u) = f_{L} \circ f_{L-2}\circ\ldots \circ f_1(u). \]
This formulation reinforces the choice of nonlinear activation function $\sigma$, cf. Remark \ref{rem:nonlin}. Otherwise, the output $y^{[L]}$ is linearly dependent on the input $u$ and hidden layers can be eliminated. Hence, with linear activation function, solving rather simple tasks like modeling the logical "XOR" function will not be possible. However, sometimes it can be favorable to have one linear layer.

\begin{remark}
	In practice, it is not uncommon that the last layer of a FNN is indeed linear, i.e. $y^{[L]} = W^{[L-1]} y^{[L-1]}$. As long as the previous layers are nonlinear this does not hinder the expressiveness of the FNN, and is typically used to attain a desired output dimension. In essence, $W^{[L-1]}$ can be seen as a reformatting, in this case. 
\end{remark}

However, in the remainder of this section we will consider the FNN architecture as introduced in \eqref{eq:FNNarchitecture}.
%
Let us now try again to represent the "XOR" logical function, this time by a FNN with one hidden layer.
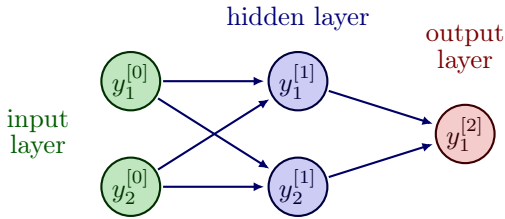
\begin{figure}[h]
	\begin{minipage}{0.5\textwidth}
		\begin{center}
			\begin{tikzpicture}[x=2.2cm,y=1.4cm]
				\message{^^JNeural network with arrows}
				\readlist\Nnod{2,2,1} 
				\message{^^J  Layer}
				\foreachitem \N \in \Nnod{ 
					\edef\lay{\Ncnt} 
					\message{\lay,}
					\pgfmathsetmacro\prev{int(\Ncnt-1)} 
					\foreach \i [evaluate={\y=\N/2-\i; \x=\lay; \n=\nstyle;}] in {1,...,\N}{ 
						\node[node \n] (N\lay-\i) at (\x,\y) {$y_\i^{[\prev]}$};
						\ifnum\lay>1
						\foreach \j in {1,...,\Nnod[\prev]}{ 
							\draw[connect arrow] (N\prev-\j) -- (N\lay-\i); 
						}
						\fi 
					}
				}
				\node[left of =5,align=center,mygreen!60!black] at (0.9,-0.5) {input\\[-0.2em]layer};
				\node[above=4,align=center,myblue!60!black] at (N2-1.90) {hidden layer};
				\node[above=8,align=center,myred!60!black] at (N\Nnodlen-1.90) {output\\[-0.2em]layer};	
			\end{tikzpicture}
		\end{center}
	\end{minipage} \hfill
	\begin{minipage}[c]{0.5\textwidth}
		\vspace{1cm}
		\caption{A feedforward network with a two dimensional input and one hidden layer with 2 nodes, i.e. $n_0 = n_1 = 2, n_2 = 1$ and $L=2$.}
		\label{fig:XOR}
	\end{minipage}
\end{figure}

The variable choices 
\[W^{[0]} = \begin{pmatrix}
	1 & 1 \\ -1 & -1
\end{pmatrix}, \quad 
b^{[0]} = \begin{pmatrix} -1 \\ 1 \end{pmatrix}, \quad
W^{[1]} = \begin{pmatrix}
	1 & 1
\end{pmatrix}, \quad 
b^{[1]} = -2, \]
solve the task and lead to the following truth table.
\begin{table}[h]
	\centering
	\begin{tabular}{| c | c | c | c | c |}
		\hline
		input $y^{[0]}_1$ & input $y^{[0]}_2$ & $y^{[1]}_1$ & $y^{[1]}_2$ & $y^{[0]}_1$ XOR $y^{[0]}_2$ (output $y^{[2]}_1$)\\
		\hline 
		0 & 0 & 0 & 1 & 0 \\
		\hline 
		1 & 0 & 1 & 1 & 1 \\
		\hline 
		0 & 1 & 1 & 1 & 1 \\
		\hline 
		1 & 1 & 1 & 0 & 0 \\
		\hline 
	\end{tabular}
	\label{tab:XOR2}
	\caption{Truth table for the logical "XOR" function modeled by the FNN from Figure \ref{fig:XOR} and given variable choices as above.}
\end{table}

Next, we formulate an optimization problem similar to \eqref{eq:LP} for multiple layers with the above introduced notation, cf. \cite{antil2022deep,antildiazherberg,antil2020fractional}
\begin{align}\label{eq:Pell} \tag{$P_\ell$}
	&\min_{\{W^{[\ell]}\}_\ell, \{b^{[\ell]}\}_\ell} \mathscr{L}\left(\{y^{[L](i)}\}_i,\{u^{(i)}\}_i, \{W^{[\ell]}\}_\ell, \{b^{[\ell]}\}_\ell \right) \\ 
	&\qquad\; \text{s.t.} \qquad y^{[L](i)} = \mathcal{F}\left( u^{(i)},\{W^{[\ell]}\}_\ell, \{b^{[\ell]}\}_\ell\right). \notag
\end{align}
If we collect again all variables in one vector $\theta$ this will have the following length:
\[ \underbrace{n_0\cdot n_1 + \ldots + n_{L-1} \cdot n_L}_{\text{weights}}+\underbrace{n_1 + \ldots + n_L.}_{\text{biases}}  \] 
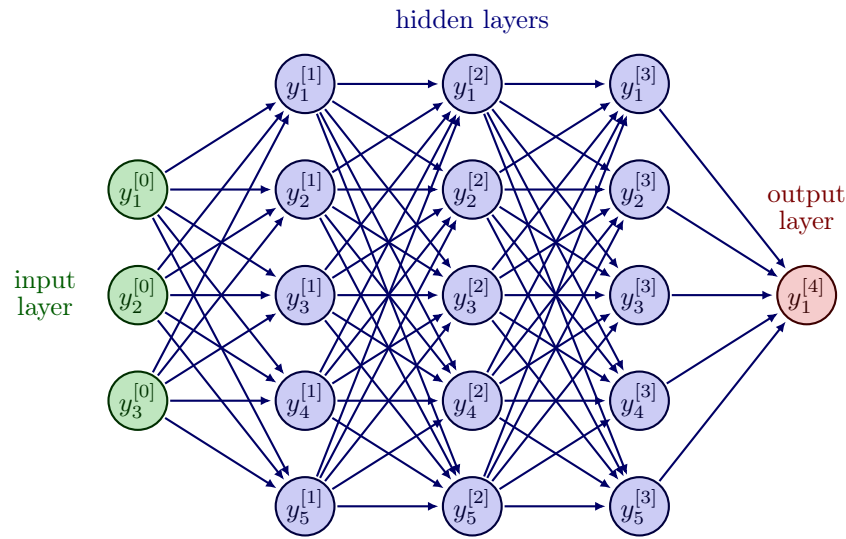
\begin{figure}[h]
		\begin{center}
			\begin{tikzpicture}[x=2.2cm,y=1.4cm]
				\message{^^JNeural network with arrows}
				\readlist\Nnod{3,5,5,5,1} 
				\message{^^J  Layer}
				\foreachitem \N \in \Nnod{ 
					\edef\lay{\Ncnt} 
					\message{\lay,}
					\pgfmathsetmacro\prev{int(\Ncnt-1)} 
					\foreach \i [evaluate={\y=\N/2-\i; \x=\lay; \n=\nstyle;}] in {1,...,\N}{ 
						\node[node \n] (N\lay-\i) at (\x,\y) {$y_\i^{[\prev]}$};
						\ifnum\lay>1
						\foreach \j in {1,...,\Nnod[\prev]}{ 
							\draw[connect arrow] (N\prev-\j) -- (N\lay-\i); 
						}
						\fi 
					}
				}
				\node[left of =5,align=center,mygreen!60!black] at (0.9,-0.5) {input\\[-0.2em]layer};
				\node[above=4,align=center,myblue!60!black] at (N3-1.90) {hidden layers};
				\node[above=8,align=center,myred!60!black] at (N\Nnodlen-1.90) {output\\[-0.2em]layer};	
			\end{tikzpicture}
		\end{center}
		\caption{A feedforward network with 3 hidden layers, layer widths $n_0 =3,  n_1 = n_2 = n_3 = 5, n_4 = 1$ and depth $L=4$. Collecting all variables of this network in a vector will give $\theta \in \mathbb{R}^{86}.$}
		\label{fig:FNN}
\end{figure}
%

The question that immediately arises is: How to choose the network architecture, i.e. depth and width of the FNN? The following discussion is based on \cite[Section 6.4]{goodfellow2016deep}.

\subsection{Depth and Width}
\label{subsec:depandwid}
The universal approximation theorem, see e.g. \cite{cybenko}, states that any vector-valued, multivariate, measurable (in particular, continuous) function $f:\mathbb{R}^{n_0} \rightarrow \mathbb{R}^{n_L}$ can be approximated with arbitrary small error by a Neural Network with one hidden layer. Hence, a first approach may be to choose a network with depth $L=2$ and increase the width until the desired accuracy is reached. 

However, this poses certain problems. First of all the universal approximation theorem does not imply that a training algorithm will actually reach the desired approximation, but rather that some set of parameters exists, that satisfies the requirement. The training algorithm might for example only find a local minimum or choose the wrong function as a result of overfitting. Another problem may be the sheer size of the layer required to achieve the wanted accuracy. In the worst case the network will need an exponential number of hidden units, with one hidden unit for each possible combination of inputs. Thus in practice, one is discouraged to use only one hidden layer, but instead to use deep networks.

Various families of functions are efficiently approximated by deep networks with a smaller width. If one desires to approximate the same function to the same degree of accuracy, the number of hidden units typically grows exponentially. This stems from the fact, that each new layer allows the network to make exponentially more connections, thus allowing a wider output of target functions. Another reason why one might choose deeper networks is due to the intuition, that our desired function may well be a composition of multiple functions. Each new layer adds a nonlinear layer function to our network, thus making it easier for the FNN to approximate composite functions. Also, heuristically we observe that deep networks typically outperform shallow networks.

Nonetheless, one main problem with deep FNNs is, that the gradient used for training is the product of the partial derivatives of each layer, as we will see in Section \ref{subsec:Backprop}. If these derivatives have small values, then the gradient for earlier layers can become very small. Thus training has a smaller if not even a negligible effect on the first layers when the network is too deep. This is especially a problem for sigmoid activation function, since its derivative is bounded by $\frac{1}{4}$.

As discussed in Section \ref{subsec:Hyper}, choosing hyperparameters like the depth and width is a non-trivial undertaking and currently, the method of choice is experimenting to find a suitable configuration for the given task. 

Additionally, in the optimization algorithms, cf. Section \ref{subsec:alg}, we need a starting point $\theta^0$ for the variables. Hence, we discuss how to initialize the weights and biases in the FNN.

\subsection{Initialization}
\label{subsec:Initial}
Recall that due to the non-convex loss function, we can only expect convergence to stationary points, cf. Remark \ref{rem:stat}. Consequently, the choice of the initial point $\theta^0$ can have great impact on the algorithm, since two different initial points can lead to two different results. An unsuitable initial point may even prevent convergence altogether. Similar to choosing hyperparameters, for the choice of initial points there exist several well-tested strategies, cf. \cite[Section 4.2.2]{geiger2021DL}, but it is still an active field of research. 

The naive approach would be to initialize $\theta = 0$ or with some other constant value. Unfortunately, this strategy has major disadvantages. With this initialization, all weights per layer in the Neural Network have the same influence on the loss function and will therefore have the same gradient. This leads to all those neurons evolving symmetrically throughout training, so that different neurons will not learn different things, which significantly reduces the expressiveness of the FNN. Let us remark that it is fine to initialize the biases $b^{[\ell]}$ with zero, as long as the weights $W^{[\ell]}$ are not initialized constant. Hence, it is sufficient to discuss initialization strategies for the weights.

We know now that the weights should be initialized in a way that they differ from each other to ensure \textbf{symmetry breaking}, i.e. preventing the neurons from evolving identically. One way to achieve this is \textbf{random initialization}. However, immediately the next question arises: How to generate those random values?

For example, weights can be drawn from a Gaussian distribution with mean zero and some fixed standard deviation. Choosing a small standard deviation, e.g. 0.01, may cause a problem known as vanishing gradients for deep networks, since the small neuron values will be multiplied with each other in the computation of gradients due to the chain rule, leading to exponentially decaying products. As a result learning can be very slow or even diverge. 
On the other hand, choosing a large standard deviation, e.g. 0.2, can result in exploding gradients, which is essentially the opposite problem, where the products grow exponentially and learning can become unstable, oscillate or even produce "NaN" values for the variables. Furthermore, in combination with saturating activation functions like sigmoid or $\tanh$ exploding variable values can lead to saturation of the activation function, which then leads to vanishing gradients and again hinder learning, cf. Figure \ref{fig:activationfunctions}. 

To find a good intermediate value for the standard deviation, \textbf{Xavier initialization} has been proposed in \cite{glorot2010understanding}, where the standard deviation value is chosen depending on the input size of the layer, i.e. 
\[\frac{1}{\sqrt{n_{\ell}}}\] 
for $W^{[\ell]}, \ell = 0,\ldots,L-1$. Note that since input sizes can vary, the Gaussian distribution that the initial values are drawn from, will also vary. This choice of weights in combination with $b^{[\ell]} = 0$ leads to $\text{Var}(y^{[\ell+1]}) = \text{Var}(y^{[\ell]})$. However, the Xavier initialization assumes zero centered activation functions, which is not fulfilled for sigmoid and all variants of ReLU. 
As a remedy, the \textbf{He initialization} has been proposed in \cite{he2015delving}, tailored especially to ReLU activation functions. Here, the standard deviation is also chosen depending on the input size of the layer, namely 
\[\sqrt{\frac{2}{n_\ell}}.\] 

Additionally, there also exists an approach to normalize feature vectors throughout the network.

\subsection{Batch Normalization}
\label{subsec:Batch}
Assume that we are employing an optimization algorithm, which passes through the data points in batches of size $b$ and that the nodes in hidden layers follow a normal distribution. Then \textbf{batch normalization} \cite{ioffe2015batch} aims at normalizing the feature vectors in a hidden layer over the given batch, to stabilize training, especially for unbounded activation functions, such as ReLU. It can be seen as insertion of an additional layer in a deep neural network, and this layer type is already pre-implemented in learning frameworks like tensorflow and pytorch: 
\begin{itemize}
	\item tf.keras.layers.BatchNormalization,
	\item torch.nn.BatchNorm1d (also 2d and 3d available). 
\end{itemize}

Say, we have given the current feature vectors $y^{[\ell]}_j \in \mathbb{R}^{n_\ell}$ of hidden layer $\ell$ for all elements in the batch, i.e. $j=1,\ldots,b$. The batch normalzation technique first determines the mean and variance
\[ \mu^{[\ell]} = \frac{1}{b} \sum_{j=1}^b y^{[\ell]}_j, \qquad (\sigma^2)^{[\ell]} = \frac{1}{b} \sum_{j=1}^b \left\| y^{[\ell]}_j - \mu^{[\ell]}\right\|^2. \]

Subsequently, the current feature vectors $y^{[\ell]}_j, j = 1,\ldots,b$ are normalized via 
\[ \hat{y}^{[\ell]}_j = \frac{ y_j^{[\ell]}-\mu^{[\ell]} }{ \sqrt{ (\sigma^2)^{[\ell]} + \epsilon }}, \] 
where $\epsilon \in \mathbb{R}$ is a constant that helps with numerical stability.
Finally, the output of the batch normalization layer is computed by 
\[ y^{[\ell+1]}_j = W^{[\ell]}  \hat{y}^{[\ell]}_j  + b^{[\ell]} \qquad \forall \, j=1,\ldots,b.\]
As usual, the weight $W^{[\ell]} \in \mathbb{R}^{n_{\ell+1} \times n_\ell}$ and bias $b^{[\ell]} \in \mathbb{R}^{n_{\ell +1}}$ are variables of the neural network and will be learned.

\begin{figure}[h!]
	\includegraphics[width=\textwidth]{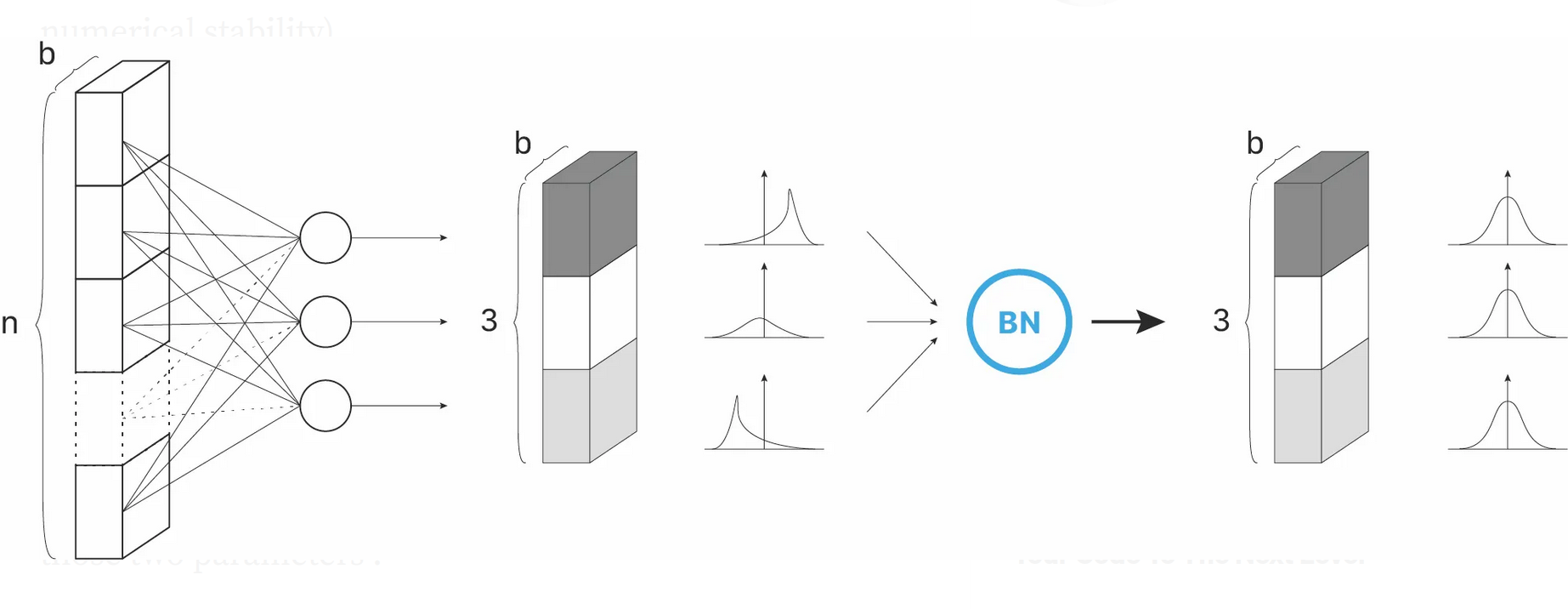}
	\label{fig:BN}
	\caption{Illustration of Batch Normalization applied to a hidden layer with 3 nodes and batch size $b$. We assume that every node can be modeled by a normal distribution. Image Source: \url{https://towardsdatascience.com/batch-normalization-in-3-levels-of-understanding-14c2da90a338}.}
\end{figure}

\begin{remark}
	The BN layer is a trainable layer, since it contains learnable variables.
\end{remark}

Let us now consider an example task that is commonly solved with FNNs.

\subsection{Classification Tasks}
\label{subsec:Class}
As mentioned in \ref{subsec:Supervised}, classification is a supervised learning task with labels $S(u) \in \mathbb{N}$. First, we consider \textbf{binary classification}, cf. e.g. \cite[Section 2.1]{geiger2021DL}, where we classify the data into two categories, i.e. a spam filter that determines whether an email is spam or not. We could construct our network so that it indicates the category which it concludes is the most likely. However, it can be useful to know the probability assigned to the output, so that we know how certain the decision is. Thus, we aim to have outputs between 0 and 1, so that they sum up to 1. In the case of binary classification, the second output is directly determined by the first. Consequently, it suffices to have a one dimensional output layer $y^{[L]} \in [0,1]$, which should predict 
\[P(y^{[L]} = 1 \, | \, u, \theta),\] 
i.e. the probability of the output being category 1 (e.g. spam) given the input $u$ and variables $\theta$. 

Assume that we have already set up a feedforward network up to the second to last layer $y^{[L-1]} \in \mathbb{R}$. It remains to choose the activation function $\sigma^{[L-1]}$ that enters the computation of $y^{[L]}$ and to model the loss function $\mathscr{L}$. 

\begin{figure}[h!]
\begin{minipage}{0.6\textwidth}
		Since we want $y^{[L]} \in [0,1]$, a common approach is to use the sigmoid activation function.
		\[\sigma(y) = \frac{1}{1+\exp(-y)} = \frac{\exp(y)}{\exp(y)+1} \in (0,1).\] 
\end{minipage}
\begin{minipage}{0.4\textwidth}
	\begin{center}
		\begin{tikzpicture}
			\begin{axis}[xmin = -5, xmax = 5, ymin = -0.25, ymax = 1.25,grid=both,width=\textwidth,height=0.6\textwidth]
				\addplot[thick] {1/(1+exp(-x))};
			\end{axis}
		\end{tikzpicture}
	\end{center}
\end{minipage}
\end{figure}

Let us remark that the cases $y^{[L]} \in \{0,1\}$ are not possible with this choice of activation function, but we are only computing approximations anyhow.

Next, we construct a loss function. To this end, we assume that the training data is a sample of the actual relationship we are trying to train, thus it obeys a probability function, which we want to recover. The main idea for the loss function is to maximize the likelihood of the input parameters, i.e. if the probability $P(y^{[L]} = S(u) \, | \, u,\theta) $ of generating the known supervision $S(u)$ is high for the input data $u$, the loss should be small, and vice versa. To model the probability we choose the Bernoulli distribution, which models binary classification:
\[ P\left(y^{[L]} = S(u) \, | \, u,\theta\right) = (y^{[L]})^{S(u)} (1-y^{[L]})^{(1-S(u))}. \]

To achieve small loss for large probabilities, we apply the logarithm and then maximize this function, so that the optimal network variables $\bar \theta$ can be determined as follows
\begin{align*}
	\bar{\theta} &= \underset{\theta}{\operatorname{argmax}} \sum_{i=1}^N \log\left(P\left(y^{[L](i)} = S(u^{(i)}) \, | \, u^{(i)},\theta\right)\right) \\
	&= \underset{\theta}{\operatorname{argmax}} \sum_{i=1}^N \log\left( (y^{[L](i)})^{S(u^{(i)})} (1-y^{[L](i)})^{(1-S(u^{(i)}))} \right)\\
	&= \underset{\theta}{\operatorname{argmax}} \sum_{i=1}^N S(u^{(i)}) \cdot \log(y^{[L](i)}) + (1-S(u^{(i)})) \cdot \log(1-y^{[L](i)}) \\
	&= \underset{\theta}{\operatorname{argmin}} \sum_{i=1}^N \underbrace{- S(u^{(i)}) \cdot \log(y^{[L](i)}) - (1-S(u^{(i)})) \cdot \log(1-y^{[L](i)})}_{=: \mathcal{L}\left(y^{[L](i)},S(u^{(i)})\right), \; \textbf{Binary Cross Entropy Loss}},
\end{align*}
where $y^{[L](i)}$ is a function of the network variables $\theta$.

In practice, we minimize the cross-entropy, since it is equivalent to maximizing the likelihood, but stays within our given frame of minimization problems. Let us assume that our data either has the label $S(u)= 1$ (spam) or $S(u) = 0$ (not spam), then the binary cross entropy loss is
\[  \mathscr{L}\left(y^{[L]},S(u)\right) = \begin{cases}
	-\log(y^{[L]}), &\text{if } S(u) = 1, \\
	-\log(1-y^{[L]}), &\text{if } S(u) = 0.
\end{cases} \]

\begin{figure}[h!]
	\centering
	\includegraphics[width=0.92\textwidth]{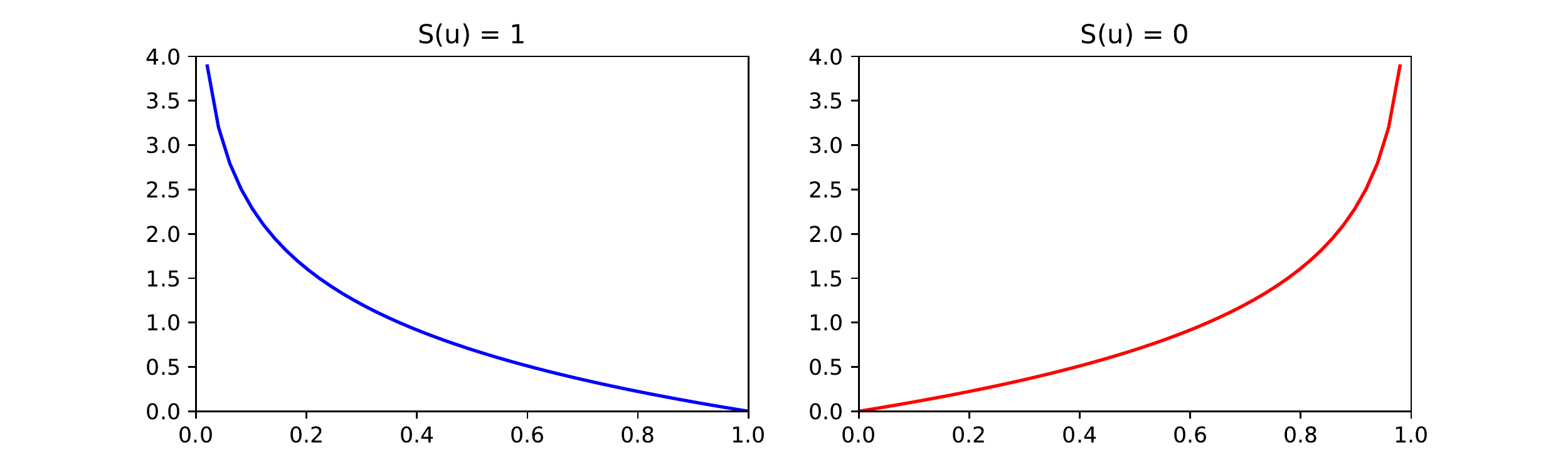}
	\caption{Binary Cross Entropy Loss for label $S(u) = 1$ (left) and label $S(u)=0$ (right).}
	\label{fig:crossentropy}
\end{figure}

We see in Figure \ref{fig:crossentropy} that the cross entropy loss for $S(u)=1$ goes to zero, for $y^{[L]} \nearrow 1$, and grows for $y^{[L]} \searrow 0$, as desired. On the other hand, the cross entropy loss for $S(u)=0$ goes to zero for $y^{[L]} \searrow 0$ and grows for $y^{[L]} \nearrow 1$.

Altogether, we have the following loss function for the binary classification task
\[ \mathscr{L}(\theta) = \frac{1}{N} \sum_{i=1}^N \mathcal{L}\left(y^{[L](i)}(\theta),S(u^{(i)})\right) .\]

\begin{figure}[h!]
	\centering
	\includegraphics[width=0.6\textwidth]{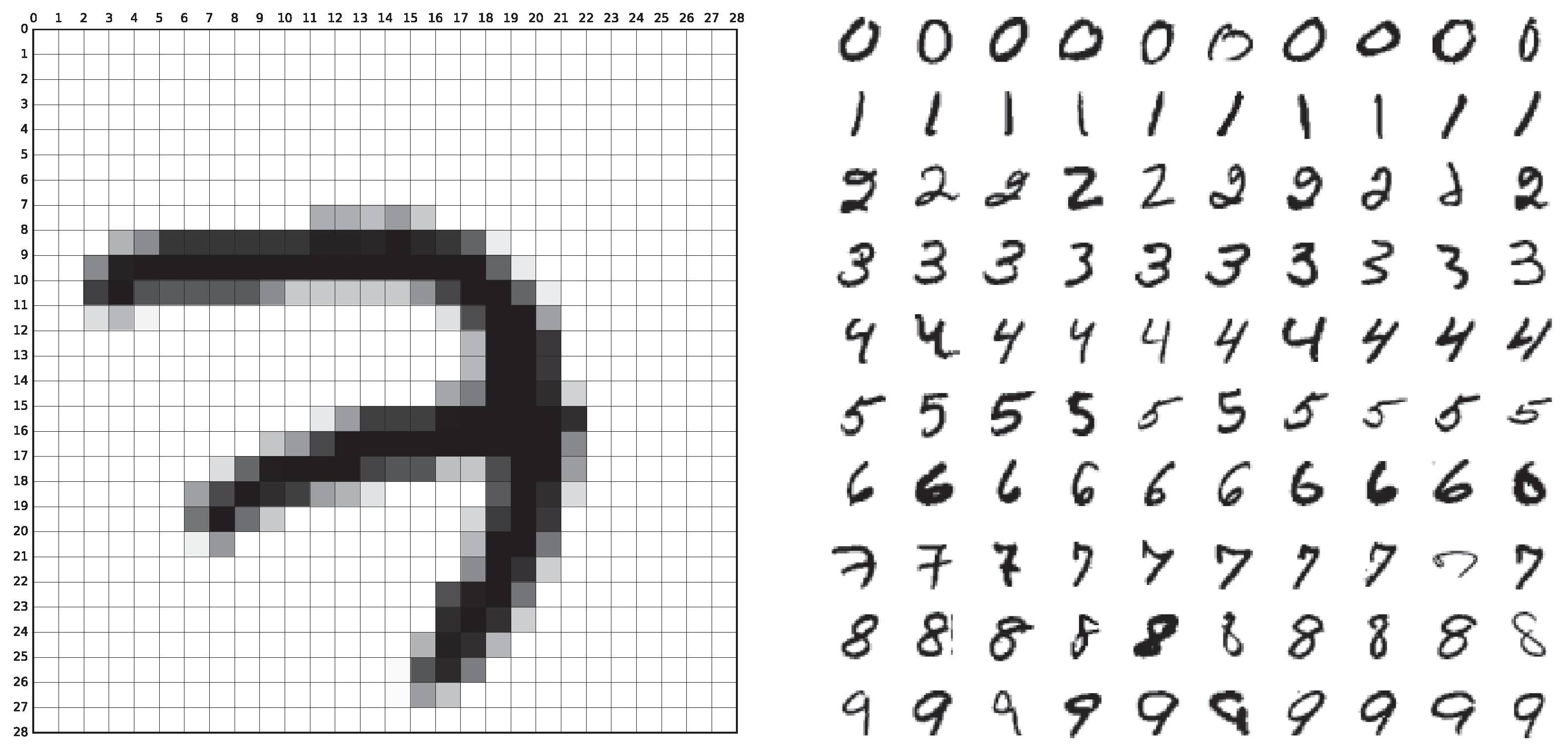}
	\caption{Example of the MNIST database. Sample belonging to the digit 7 (left) and 100 samples from all 10 classes (right). Image Source: \cite[Fig. 1]{mnist7}.}
	\label{fig:MNIST_7}
\end{figure}

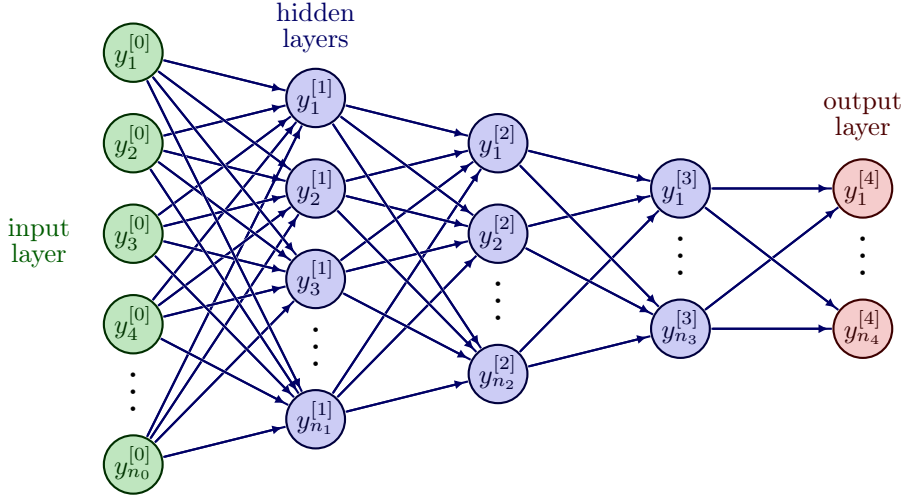
\begin{figure}[h]
	\begin{center}
		\begin{tikzpicture}[x=2.4cm,y=1.2cm]
			\readlist\Nnod{5,4,3,2,2} 
			\readlist\Nstr{n_0,n_{\prev},n_{\prev}} 
			\readlist\Cstr{y^{[0]},y^{[\prev]},y^{[\prev]}} 
			\def\yshift{0.55} 
			
			\foreachitem \N \in \Nnod{
				\def\lay{\Ncnt} 
				\pgfmathsetmacro\prev{int(\Ncnt-1)} 
				\foreach \i [evaluate={\c=int(\i==\N); \y=\N/2-\i-\c*\yshift;
					\x=\lay; \n=\nstyle;
					\index=(\i<\N?int(\i):"\Nstr[\n]");}] in {1,...,\N}{ 
					\node[node \n] (N\lay-\i) at (\x,\y) {$\strut\Cstr[\n]_{\index}$};
					
					\ifnumcomp{\lay}{>}{1}{ 
						\foreach \j in {1,...,\Nnod[\prev]}{ 
							\draw[white,line width=1.2,shorten >=1] (N\prev-\j) -- (N\lay-\i);
							\draw[connect] (N\prev-\j) -- (N\lay-\i);
						}
					}
				{}
				\ifnum\lay>1
				\foreach \j in {1,...,\Nnod[\prev]}{ 
					\draw[connect arrow] (N\prev-\j) -- (N\lay-\i); 
				}
				\fi 
					
				}
				\path (N\lay-\N) --++ (0,1+\yshift) node[midway,scale=1.6] {$\vdots$}; 
			}
			
			\node[left of=5,align=center,mygreen!60!black] at (0.9,-0.6) {input\\[-0.2em]layer};
			\node[above=2,align=center,mydarkblue] at (N2-1.90) {hidden\\[-0.2em]layers};
			\node[above=3,align=center,mydarkred] at (N\Nnodlen-1.90) {output\\[-0.2em]layer};
			
		\end{tikzpicture}
	\end{center}
	\caption{A feedforward network with 3 hidden layers, i.e. depth $L=4$. For the MNIST data set we have $n_0 = 784$ and $n_4 = 10$.}
	\label{fig:FNN_multi}
\end{figure}

\textbf{Multiclass classification} is a direct extension of binary classification. Here, the goal is to classify the data into multiple (at least 3) categories. A prominent example is the MNIST data set, where black and white pictures of handwritten digits (of size $28 \times 28$ pixels) are supposed to be classified as $\{0,1,2,3,4,5,6,7,8,9\}$, cf. Figure \ref{fig:MNIST_7}. A possible network architecture is illustrated in Figure \ref{fig:FNN_multi}.

For this task, we need a generalization of the sigmoid activation function, which will take $y^{[L-1]} \in \mathbb{R}^{n}$ and map it to $y^{[L]} \in [0,1]^n$, so that we have $\sum_{i=1}^n y^{[L]}_i = 1$, where $n_{L-1} = n_L = n$ is the number of classes. A suitable option is the softmax function, which is given component-wise by 
\[ \operatorname{softmax}(y)_i = \frac{\exp(y_i)}{\sum_{j=1}^n \exp(y_j)} \in (0,1), \qquad \text{for } i = 1, \ldots, n. \]
Keep in mind that e.g. in the MNIST case we have labels $S(u) \in \{0,1,2,3,4,5,6,7,8,9\}$, and in general we have multiple labels $S(u) \in \mathbb{N}$. We have seen before that maximizing the log-likelihood is a suitable choice for classification tasks. Since we want to formulate a minimization problem, we choose the negative log-likelihood as loss function 
\[  \mathscr{L} (\theta) = \frac{1}{N} \sum_{1=1}^N - \log \left(P\left(y^{[L](i)} = S(u^{(i)}) \, | \, u^{(i)},\theta\right)\right). \]

In this section we have seen yet another model for the loss function $\mathscr{L}$, and from Section \ref{subsec:alg} we know that in any case we will need the gradient $\nabla \mathscr{L}(\theta)$ to update our variables $\theta$. Let us discuss how frameworks like pytorch and tensorflow obtain this information. 

\subsection{Backpropagation}
\label{subsec:Backprop} 
The derivations are based on \cite[Section 6.5]{goodfellow2016deep} and \cite[Section 7.3]{ng2022}.
When a network, e.g. a FNN, takes an input $u$, passes it through its layers and finally computes an output $y^{[L]}$, the network \textbf{feeds forward} the information, which this is called \textbf{forward propagation}. Then a loss is assigned to the output and we aim at employing the gradient of the loss function $\nabla \mathscr{L}(\theta)$ to update the network variables $ \theta \in \mathbb{R}^K$. In a FNN we have $K = n_0\cdot n_1 + \ldots + n_{L-1} \cdot n_L+n_1 + \ldots + n_L. $ The gradient is then given by 
\begin{equation*}
	\nabla \mathscr{L}(\theta) = \begin{pmatrix}
		\frac{\partial \mathscr{L}(\theta)}{\partial \theta_1} \\
		\vdots \\
		\frac{\partial \mathscr{L}(\theta)}{\partial \theta_K}
	\end{pmatrix}.
\end{equation*}

To develop an intuition about the process, we discuss the following simple example.

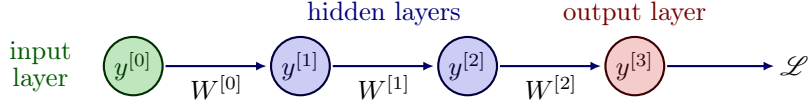
\begin{figure}[h]
	\begin{center}
		\begin{tikzpicture}[x=2.2cm,y=1.4cm]
			\message{^^JNeural network with arrows}
			\readlist\Nnod{1,1,1,1} 
			\message{^^J  Layer}
			\foreachitem \N \in \Nnod{ 
				\edef\lay{\Ncnt} 
				\message{\lay,}
				\pgfmathsetmacro\prev{int(\Ncnt-1)} 
				\foreach \i [evaluate={\y=\N/2-\i; \x=\lay; \n=\nstyle;}] in {1,...,\N}{ 
					\node[node \n] (N\lay-\i) at (\x,\y) {$y^{[\prev]}$};
					\ifnum\lay>1
					\foreach \j in {1,...,\Nnod[\prev]}{ 
						\draw[connect arrow] (N\prev-\j) -- (N\lay-\i); 
					}
					\fi 
				}
			}
			\draw[connect arrow] (N\Nnodlen-1) -- (4.85,-0.5) ;
			\node[left of =5,align=center,mygreen!60!black] at (0.9,-0.5) {input\\[-0.2em]layer};
			\node[above=4,align=center,myblue!60!black] at (2.5,-0.3) {hidden layers};
			\node[right of=8,align=center] at (4.5,-0.5) {$\mathscr{L}$};
			\node[above=8,align=center,myred!60!black] at (4,-0.4) {output layer};	
			\node at (1.5,-0.7) {$W^{[0]}$};
			\node at (2.5,-0.7) {$W^{[1]}$};
			\node at (3.5,-0.7) {$W^{[2]}$};
		\end{tikzpicture}
	\end{center}
	\caption{A feedforward network with 2 hidden layers, one node per layer and depth $L=3$.}
	\label{fig:FNN_simple}
\end{figure}

\begin{example}\label{ex:FNN_grad}
Consider a very simple FNN with one node per layer and assume that we only consider weights $W^{[\ell]} \in \mathbb{R}$ and no biases. 
For the network in Figure \ref{fig:FNN_simple} we have  $\theta = (W^{[0]},W^{[1]},W^{[2]})^{\top}$, 
\[ y^{[3]} = \sigma^{[3]}(W^{[2]}y^{[2]}) = \sigma^{[3]}\left(W^{[2]} \sigma^{[2]} (W^{[1]} y^{[1]}) \right) = \sigma^{[3]}\left(W^{[2]} \sigma^{[2]} \left(W^{[1]} \sigma^{[1]}(W^{[0]}y^{[0]})\right) \right), \]
and 
\[ \mathscr{L}(\theta) = \mathscr{L}(y^{[3]}(\theta))= \mathscr{L} \left(\sigma^{[3]}\left(W^{[2]} \sigma^{[2]} \left(W^{[1]} \sigma^{[1]}(W^{[0]}y^{[0]})\right) \right)\right) .\]
Computing the components of the gradient, we employ the chain rule to obtain e.g. 
\begin{align*}
	\frac{\partial \mathscr{L}}{\partial W^{[0]}} &= \frac{\partial \mathscr{L}}{\partial y^{[3]}} \cdot \frac{\partial y^{[3]}}{\partial W^{[0]}} \\
	&= \frac{\partial \mathscr{L}}{\partial y^{[3]}} \cdot \frac{\partial y^{[3]}}{\partial y^{[2]}} \cdot  \frac{\partial y^{[2]}}{\partial W^{[0]}}\\
	&= \frac{\partial \mathscr{L}}{\partial y^{[3]}} \cdot \frac{\partial y^{[3]}}{\partial y^{[2]}} \cdot
	\frac{\partial y^{[2]}}{\partial y^{[1]}} \cdot  \frac{\partial y^{[1]}}{\partial W^{[0]}},
\end{align*}
and in general for depth $L$ we get 
\[\frac{\partial \mathscr{L}}{\partial W^{[\ell]}} =  \frac{\partial \mathscr{L}}{\partial y^{[L]}} \cdot 
\prod_{j=L}^{\ell+2} \frac{\partial y^{[j]}}{\partial y^{[j-1]}} 
\cdot  \frac{\partial y^{[\ell+1]}}{\partial W^{[\ell]}}.\]
\end{example}

Essentially, to calculate the effect of a variable on the loss function we iterate backwards through the network, multiplying the derivatives of each layer. This is called \textbf{back propagation}, often abbreviated as \textbf{backprop}. 

To obtain a computationally efficient version of back propagation, we exploit the fact that parts of the derivatives can be recycled, broadly speaking. E.g. 
$ \frac{\partial \mathscr{L}} {\partial y^{[L]}}$
is a part of all derivatives. So, if we compute the derivative by $W^{[L-1]}$ first, we already have this component available and can reuse it in the computation of the derivative by $W^{[L-2]}$, etc.

In order to formalize the effective computation of derivatives in a backpropagation algorithm, we decompose the forward propagation into two parts, cf. e.g. \cite[Section 7.3.2]{ng2022}.
\begin{align*}
	\hspace{4cm} z^{[\ell]} &= W^{[\ell-1]}y^{[\ell-1]} + b^{[\ell-1]} &&\in \mathbb{R}^{n_\ell}, \hspace{4cm} \\
	y^{[\ell]} &= \sigma^{\ell]}(z^{[\ell]})   &&\in \mathbb{R}^{n_\ell}.
\end{align*}
This was not necessary in Example \ref{ex:FNN_grad}, because we only consider weights and no biases.  
Furthermore, we assume that the loss function $\mathscr{L}$ takes the final output $y^{[L]}$ as an input. Especially, no other feature vectors $y^{[\ell]}$ for $\ell\neq L$ enter the loss function directly. This is the case e.g. for mean squared error, cf. Example \ref{ex:LR}, and cross entropy, cf. Section \ref{subsec:Class}. 

In general, we now have by chain rule for all $\ell=0,\ldots,L-1$
\begin{align}
	\frac{\partial \mathscr{L}}{\partial W^{[\ell]}} &=  \frac{\partial \mathscr{L}}{\partial y^{[L]}} \cdot \prod_{j=L}^{\ell+2} \left( \frac{\partial y{[j]}}{\partial z^{[j]}} \cdot  \frac{\partial z^{[j]}}{\partial y^{[j-1]}} \right) \cdot \frac{\partial y^{[\ell+1]}}{\partial z^{[\ell + 1]}} \cdot \frac{\partial z^{[\ell+1]}}{\partial W^{[\ell]}}  ,  \label{eq:chainW}  \\
	\frac{\partial \mathscr{L}}{\partial b^{[\ell]}} &=  \frac{\partial \mathscr{L}}{\partial y^{[L]}} \cdot \prod_{j=L}^{\ell+2} \left( \frac{\partial y{[j]}}{\partial z^{[j]}} \cdot \frac{\partial z^{[j]}}{\partial y^{[j-1]}} \right) \cdot \frac{\partial y^{[\ell+1]}}{\partial z^{[\ell + 1]}} \cdot \frac{\partial z^{[\ell+1]}}{\partial b^{[\ell]}}  \label{eq:chainb}.
\end{align}
However, we have to understand these derivatives in detail. First of all, let us introduce the following definition from \cite{horn1990hadamard}. 

\begin{definition}
	Let $A,B \in \mathbb{R}^{m\times n}$ be given matrices, then $A \odot B \in \mathbb{R}^{m \times n}$, with entries 
		\[ (A\odot B)_{i,j} := (A)_{i,j} \cdot (B)_{i,j}, \qquad \text{for } i=1,\ldots,m, \, j=1,\ldots,n,\]
	is called the \textbf{Hadamard product} of $A$ and $B$.
\end{definition}

Furthermore, we define the derivative of the component-wise activation function $\sigma: \mathbb{R}^m \rightarrow \mathbb{R}^m$ as follows
\begin{equation*}
	\sigma' : \mathbb{R}^{m} \rightarrow \mathbb{R}^{m}, \qquad 
	z=
	\begin{pmatrix}
		z_1 \\ \vdots \\ z_m
	\end{pmatrix}
	\mapsto 
	\begin{pmatrix}
		\sigma'(z_1) \\ \vdots \\ \sigma'(z_m)
	\end{pmatrix} = \sigma'(z).
\end{equation*}

Let us introduce two special cases of multi-dimensional chain rule, cf. \cite[p.98]{ng2022}, which will prove helpful to calculate the derivatives. 
\begin{enumerate}
	\item Consider $a = \sigma(z) \in \mathbb{R}^m$, where $\sigma$ is a component-wise function, e.g. an activation function, $z \in \mathbb{R}^m$, and $f = f(a) \in \mathbb{R}$. Then, it holds
	\begin{equation}
		\; \frac{\partial f}{\partial z} = \frac{\partial f}{\partial a} \odot \sigma'(z) \;\; \in \mathbb{R}^{m}.   \label{eq:case1}
	\end{equation}
	\item Consider $z = W y + b \in \mathbb{R}^m$ and $f = f(z) \in \mathbb{R}$, with $W \in \mathbb{R}^{m\times n}$ and $y \in \mathbb{R}^n$. Then, it holds
			\begin{align}
				\hspace{4.5cm} \frac{\partial f}{\partial y} &= W^{\top} \cdot \frac{\partial f} {\partial z} &&\in \mathbb{R}^{n}, \hspace{4.5cm} \label{eq:case2a}\\
				\frac{\partial f}{\partial W} &= \frac{\partial f} {\partial z} \cdot y^{\top} &&\in \mathbb{R}^{m\times n}, \label{eq:case2b}\\
				\frac{\partial f}{\partial b} &= \frac{\partial f} {\partial z} &&\in \mathbb{R}^m. \label{eq:case2c}
			\end{align}
\end{enumerate}

We can now start working our way backwards through the network to get all derivatives. Assume that we know $\frac{\partial \mathscr{L}} {\partial y^{[L]}} \in \mathbb{R}^{n_L}$, which will depend in detail on the choice of loss function. We can employ \eqref{eq:case1} with $f = \mathscr{L}, z = z^{[L]}, a = y^{[L]} $ to compute 
\begin{equation*}
	\bar{z}^{[L]} := \frac{\partial \mathscr{L}}{\partial z^{[L]}} = \frac{\partial \mathscr{L}}{\partial y^{[L]}} \odot (\sigma^{[L]})'(z^{[L]}) \qquad \in \mathbb{R}^{n_L}.
\end{equation*}
Here, we employ a typical notation from automatic differentiation (AD), i.e. the gradient of the loss with respect to a certain variable is denoted by the name of that variable with an overbar. Now, we know 
\begin{equation*}
	\bar{y}^{[L-1]} := \frac{\partial \mathscr{L}}{\partial y^{[L-1]}} = \bar{z}^{[L]} \cdot \frac{\partial z^{[L]}}{\partial y^{[L-1]}} \qquad \in \mathbb{R}^{n_{L-1}},
\end{equation*}
i.e. we can reuse the previously derived gradient. Furthermore, from \eqref{eq:case2a} with $f = \mathscr{L}$, $y = y^{[L-1]}, W = W^{[L-1]}, z = z^{[L]}$ we deduce 
\begin{equation*}
	\bar{y}^{[L-1]} = (W^{[L-1]})^{\top}\bar{z}^{[L]}. 
\end{equation*} 
Subsequently, we use $\bar{y}^{[L-1]}$ to compute $\bar{z}^{[L-1]}$, and so forth. In this way we can keep iterating to build up the products in \eqref{eq:chainW} and \eqref{eq:chainb}. 

In every layer,$\ell=0,\ldots,L-1$, we also want to determine $\bar{W}^{[\ell]}:= \frac{\partial \mathscr{L}}{\partial W^{[\ell]}} $ and $\bar{b}^{[\ell]}:= \frac{\partial \mathscr{L}}{\partial b^{[\ell]}}$. We show this exemplary for $\ell = L-1$. It holds
\begin{align*}
	\bar{W}^{[L-1]} &= \bar{z}^{[L]} \cdot \frac{\partial z^{[L]}}{\partial W^{[L-1]}}  \qquad \in  \mathbb{R}^{n_L \times n_{L-1}}, \\
	\bar{b}^{[L-1]} &= \bar{z}^{[L]} \cdot \frac{\partial z^{[L]}}{\partial b^{[L-1]}} \qquad\;\,\, \in \mathbb{R}^{n_{L}}.
\end{align*}
Making use of \eqref{eq:case2b} and \eqref{eq:case2c} with the same choices as in the computation of $\bar{y}^{[L-1]}$ and $b = b^{[L-1]}$, we get
\begin{align*}
	\bar{W}^{[L-1]} &= \bar{z}^{[L]}  (y^{[L-1]})^{\top} , \\
	\bar{b}^{[L-1]} &= \bar{z}^{[L]}.
\end{align*}
With this technique, we have an iterative way to efficiently calculate all gradients needed for the variable update in the optimization method, cf. Section \ref{subsec:alg}.

\begin{remark}$\,$
	\begin{enumerate}
		\item[(i)] It is even more elegant and efficient to update $W^{[\ell]}$ and $b^{[\ell]}$ during backpropagation, i.e. on the fly. This way we do not need to store the gradient and can overwrite the weight and bias variables. It is only necessary to save the current gradients for the next loop, so we could rewrite the backpropagation algorithm with temporary gradient values. This only works if the stepsize / learning rate $\tau$ is previously known and fixed for all variables, since for a line search we would need to know the full gradient and could only update the variables afterwards.
		\item[(ii)]Considering we have $N$ training data points, the backpropagation algorithm has to take all of them into account. When the loss is a sum of loss functions for each data point, this can be easily incorporated into the algorithm by looping over $i=1,\ldots,N$ and introducing a sum where necessary.
	\end{enumerate}
\end{remark}

Altogether, we formulate Algorithm \ref{alg:backpropagation}, which collects the gradients with respect to the weights and biases in one final gradient vector $\nabla \mathscr{L} (\theta)$. 

\begin{algorithm}[h!]
	\caption{Backpropagation.}
	\begin{algorithmic}
		\Require Training data set $\{u^{(i)}, S(u^{(i)})\}_{i=1}^N$.
		\Require Current weights $W^{[\ell]}$ and biases $b^{[\ell]}$ for $\ell=0,\ldots,L-1$.
		\Require Activation functions $\sigma^{[\ell]}$ for $\ell=1,\ldots,L$.
		\Require Loss function $\mathscr{L}(y^{[L]})$ and its gradient $\nabla \mathscr{L}(y^{[L]}) = \frac{\partial \mathscr{L}} {\partial y^{[L]}} \in \mathbb{R}^{n_L}$.
		\State $y^{[0](i)} = u^{(i)} \in \mathbb{R}^{n_0} \quad$ for $i = 1, \ldots,N$.
		\For{$\ell=1,\ldots,L$}
			\State $z^{[\ell](i)} = W^{[\ell-1]}y^{[\ell-1]} + b^{[\ell-1]} \in \mathbb{R}^{n_\ell} \quad$ for $i = 1, \ldots,N$,
			\State $y^{[\ell](i)} = \sigma^{[\ell]}(z^{[\ell](i)}) \hspace{1.55cm} \in \mathbb{R}^{n_\ell} \quad$ for $i = 1, \ldots,N$. 
		\EndFor
		\State Compute loss $\mathscr{L} = \frac{1}{N} \sum_{1=1}^N \mathscr{L}(y^{[L](i)}) \in \mathbb{R}$.
		\State $\bar{y}^{[L](i)} = \frac{1}{N} \cdot \nabla \mathscr{L}(y^{[L](i)}  ) \in \mathbb{R}^{ n_L} \quad$ for $i = 1, \ldots,N$.  
		\For{$\ell = L,L-1,\ldots,1$}
		\State $\bar{z}^{[\ell](i)} \hspace{0.36cm} = \bar{y}^{[\ell](i)} \odot (\sigma^{[\ell]})' (z^{[\ell](i)}) \hspace{0.24cm} \in \mathbb{R}^{n_\ell} \quad$ \hspace{0.19cm} for $i = 1, \ldots,N$,
		\State $\bar{y}^{[\ell-1](i)} = (W^{[\ell-1]})^{\top}\bar{z}^{[\ell](i)}                                   \hspace{1.07cm} \in \mathbb{R}^{n_{\ell-1}} \quad$ for $i = 1, \ldots,N$,
		\State $\bar{W}^{[\ell-1]} \hspace{0.13cm} = \sum_{1=1}^N  \bar{z}^{[\ell](i)} (y^{[\ell-1](i)})^{\top}   \in \mathbb{R}^{ n_\ell \times n_{\ell-1}}$,
		\State $\bar{b}^{[\ell-1]} \hspace{0.37cm} = \sum_{1=1}^N \bar{z}^{[\ell](i)}                               \hspace{1.66cm} \in \mathbb{R}^{n_\ell}$.
		\EndFor
	\end{algorithmic}
	\label{alg:backpropagation}
\end{algorithm}

In frameworks like pytorch and tensorflow, backpropagation is already implemented, e.g. in pytorch the function "autograd" handles the backward pass. Broadly speaking, autograd collects the data and all executed operations in a directed acyclic graph. In this graph the inputs are the leaves, while the outputs are the roots. Now to automatically compute the gradients, the graph can be traced from roots to leaves, employing the chain rule. This coincides with the computations that we just derived by hand. For more details we refer to \url{https://pytorch.org/tutorials/beginner/blitz/autograd_tutorial.html}.

\newpage
\section{Convolutional Neural Network}
\label{sec:CNN}
\begin{figure}[ht!]
	\centering
	\includegraphics[width=0.41\textwidth]{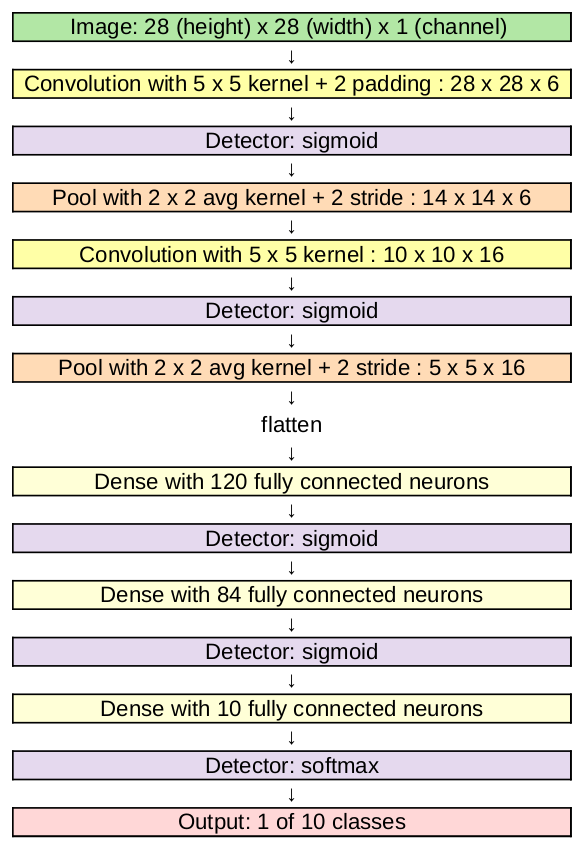}
		\caption{Architecture of LeNet-5.}
	\label{fig:LeNet}
\end{figure}

%
In this section, based on \cite{cs231n},\cite[Section 9]{goodfellow2016deep}, we consider Neural Networks with a different architecture: \textbf{convolutional neural networks} (CNNs or ConvNets). They were first introduced by Kunihiko Fukushima in 1980 under the name "neocognitron" \cite{neo}. Famous examples of convolutional neural networks today are "LeNet" \cite{lenet}, see Figure \ref{fig:LeNet} and "AlexNet" \cite{alexnet}. 

As a motivation, consider a classification task where the input is an image of size $n_{0,1} \times n_{0,2}$ pixels. We want to train a Neural Network so that it can decide, e.g. which digit is written in the image (MNIST data set). We have seen in Figure \ref{fig:FNN_multi} that the image with $n_{0,1}=n_{0,2}=28$ has been reshaped (vectorized, flattened) into a vector in $\mathbb{R}^{n_{0,1}\cdot n_{0,2}} = \mathbb{R}^{784}$, so that we can use it as an input for a regular FNN.
However, this approach has several disadvantages:
\begin{enumerate}
	\item Vectorization causes the input image to loose all of its spatial structure, which could have been helpful during training. 
	\item Let e.g. $n_{0,1}=n_{0,2}=1000$, then $n_0 = 10^6$ and the weight matrix $W^{[0]} \in \mathbb{R}^{n_1 \times 10^6}$ contains an enormous number of optimization variables. This can make training very slow or even infeasible. 
\end{enumerate}

On the contrary, convolutional neural networks are designed to exploit the relationships between neighboring pixels. In fact, the input of a CNN is typically a matrix or even a three-dimensional tensor, which is then passed through the layers while maintaining this structure. 
CNNs take small patches, e.g. squares or cubes, from the input images and learn features from them. Consequently, they can subsequently recognize these features in other images, even when they appear in other parts of the image.

In Figure \ref{fig:LeNet} we see the architecture of "LeNet-5". The inputs are images, where we have 1 channel, because we consider grayscale images. At first we have two sequences of convolution layer (yellow), Section \ref{subsec:convlayers}, detector layer (violet), Section \ref{subsec:detlayers}, and pooling layer (orange), Section \ref{subsec:poollayers}. These layers retain the multidimensional structure of the input. Since this network is built for a classification tasks, the output should be a vector of 10. Consequently, the multi-dimensional output of a hidden layer is flattened, i.e. vectorized, and the remaining layers are fully connected layers (bright yellow) as we have seen in FNNs. 

In other, larger architectures, like AlexNet, cf. Figure \ref{fig:alex}, to avoid overfitting with large fully connected layers, a technique called \textbf{dropout} is applied. The key idea is to randomly drop units with a given probability and their connections from the neural network during training, for more details we refer to \cite{dropout}.

\begin{figure}[h!]
	\begin{minipage}{0.6\textwidth}
	\centering
	\includegraphics[width=0.9\textwidth]{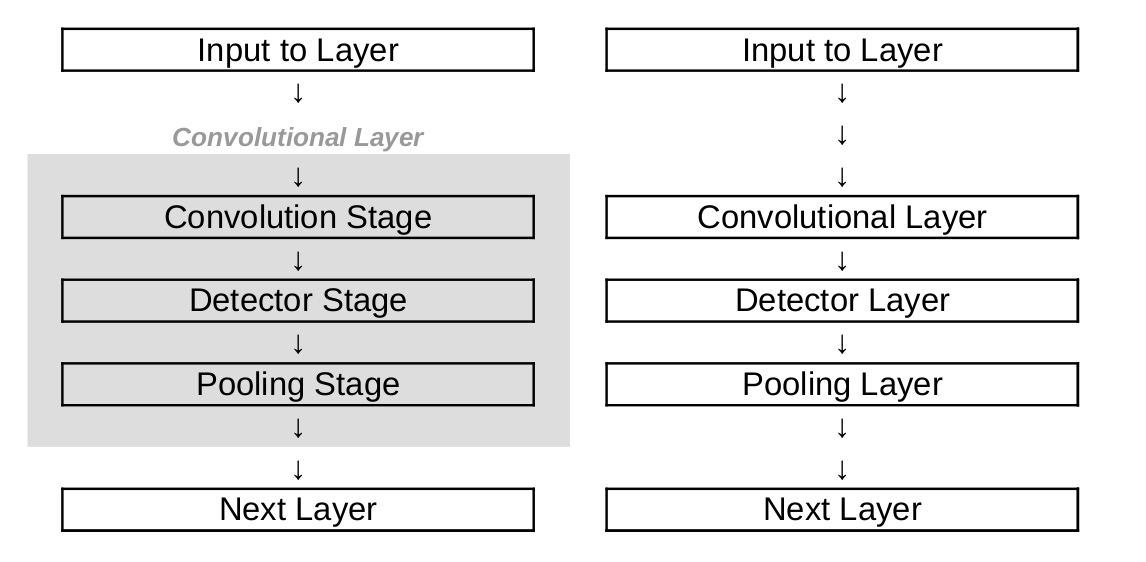}
	\end{minipage}
	\begin{minipage}{0.39\textwidth}
		\caption{Convolutional layer (gray) consisting of stages (left) compared to viewing the operations as separate layers (right). We use the terminology as depicted on the right hand side, and refer to convolutional, detector and pooling layers as separate layers. }
		\label{fig:CNN_twoways}
	\end{minipage}
\end{figure}

\begin{remark}$\,$
	\begin{enumerate}
		\item[(i)] We will view convolution, detector and pooling layers as separate layers. However, it is also possible to define a convolutional layer to consist of a convolution, detector and pooling stage, cf. Figure \ref{fig:CNN_twoways}. This can be a source of confusion when referring to convolutional layers, which we should be aware of. 
		\item[(ii)] Throughout the remainder of this section we omit layer indices $\ell$ to simplify notation, and we indicate the data with capital letter $Y$ to clarify that they are matrices or tensors. 
	\end{enumerate}
\end{remark}

Before we move on to a detailed introduction of the different layer types in CNNs, let us recall the mathematical concept of a convolution.

\subsection{Convolution}
As explained in \cite[Section 9.1]{goodfellow2016deep}, in general, convolution describes how one function influences the shape of another function.  But it can also be used to apply a weight function to another function, which is how convolution is used in convolutional neural networks. 

\begin{definition}
	Let $f, g: \mathbb{R}^n \to \mathbb{R}$ be two functions. If both $f$ and $g$ are integrable with respect to Lebesgue measure, we can define the \textbf{convolution} as:
	\begin{align*}
		c(t) = (f \ast g) (t) = \int f(x) g(t-x) \, dx,
	\end{align*}
	for some $t \in \mathbb{R}^n$. Here, $f$ is called the input and $g$ is called the kernel. The new function $c: \mathbb{R}^n \rightarrow \mathbb{R}$ is called the feature map.
\end{definition}
However, for convolutional neural networks we need the discrete version. 
\begin{definition}
	Let $f, g: \mathbb{Z}^n \to \mathbb{R}$ be two discrete functions. The \textbf{discrete convolution} is then defined as:
	\begin{align*}
		c(t) = (f \ast g) (t) = \sum_{x \in \mathbb{Z}^n} f(x) g(t - x),
	\end{align*}
	for some $t \in \mathbb{Z}^n$. 
\end{definition}
A special case of the discrete convolution is setting $f$ and $g$ to $n$-dimensional vectors and using the indices as arguments. We illustrate this approach in the following example.

\begin{example}
	Let $X$ and $Y$ be two random variable each describing the outcome of rolling a dice. The probability mass functions are defined as:
	\begin{align*}
		f_X(t) = f_Y(t) = 
		\begin{cases}
			\frac{1}{6}, &\text{if } t  \in \{ 1,2,3,4,5,6\}, \\
			0, &\text{if } t \in \mathbb{Z}\setminus\{ 1,2,3,4,5,6\}.
		\end{cases}
	\end{align*}
	We aim at calculating the probability that the sum of both dice rolls equals nine. To this end, we take the vectors of all possible outcomes and arrange them into two rows. Here, we flip the second vector and slide it to the right, such that the numbers which add to nine align.
	\begin{center}
		\begin{tikzpicture}
			\foreach \x in {1, 2, 3, 4, 5, 6}
			\draw (8 - \x , 0) rectangle (9 - \x, -1) node[midway]{\x};
			\foreach \x in {1, 2, 3, 4, 5, 6}
			\draw (\x - 1, 0) rectangle (\x, 1) node[midway]{\x};   
		\end{tikzpicture}
	\end{center}
	
	Now, we replace the outcomes with their respective probabilities, multiply the adjacent components and add up the results.
	
	\begin{center}
		\begin{tikzpicture}
			\foreach \x in {1, 2, 3, 4, 5, 6}
			\draw (\x - 1, 0) rectangle (\x, 1) node[midway] {$\frac{1}{6}$};
			\foreach \x in {1, 2, 3, 4, 5, 6}
			\draw (8 - \x , 0) rectangle (9 - \x, -1) node[midway] {$\frac{1}{6}$};		
		\end{tikzpicture}
	\end{center}
	
	This gives
	\[ f_{X+Y}(9) = \frac{1}{36} + \frac{1}{36} +\frac{1}{36} +\frac{1}{36}  = \frac{1}{9}, \]
	i.e. the probability that the sum of the dice equals nine is $\frac{1}{9}.$
	
	In fact, all the steps we have just done are equivalent to calculating a discrete convolution:
	\begin{align*}
		f_{X+Y}(9) = \sum_{x = 1}^6 f_X(x)f_Y(9- x)  = (f_X \ast f_Y) (9)
	\end{align*}
\end{example}

\subsection{Convolutional Layer}
\label{subsec:convlayers}
For the convolutional layers in CNNs we define convolutions for matrices, cf. e.g. \cite[(9.4)]{goodfellow2016deep}. This can be extended to tensors straight forward.  

\begin{definition}
	Let $Y \in \mathbb{R}^{n_1 \times n_2}$ and $K \in \mathbb{R}^{m_1 \times m_2}$ be given matrices, such that $m_1 \leq n_1$ and $m_2 \leq n_2$. The \textbf{convolution} of $Y$ and $K$ is denoted by $Y \ast K$ with entries
	\[ \left[ Y \ast K \right]_{i,j} := \sum_{k=1}^{m_1} \sum_{l=1}^{m_2} K_{k,l} Y_{i+m_1-k, j+m_2-l}, \]
	for $1 \leq i \leq n_1-m_1 + 1$ and $1 \leq j \leq n_2 - m_2 +1$. Here, $Y$ is called the input and $K$ is called the kernel. 
\end{definition}

In Machine Learning often the closely related concept of \textbf{(cross) correlation}, cf. e.g. \cite[(9.6)]{goodfellow2016deep}, is used, and incorrectly referred to as convolution, where 
\[ \left[ Y \circledast K \right]_{i,j} := \sum_{k=1}^{m_1} \sum_{l=1}^{m_2} K_{k,l} Y_{i \textcolor{red}{-1+k}, j \textcolor{red}{-1+l}}, \]
for $1 \leq i \leq n_1-m_1 + 1$ and $1 \leq j \leq n_2 - m_2 +1$. The (cross) correlation has the same effect as convolution, if you flip both, rows and columns of the kernel $K$, see the changed indices indicated in red. Since we learn the kernel anyway, it is irrelevant whether the kernel is flipped, thus either concept can be used. 

We illustrate the matrix computations with an example. 

\begin{example}\label{ex:kernel}
	For this example we have the data matrix
	\begin{equation*}
		Y = 
		\begin{pmatrix}
			1  & 5 & -2 & 0 & 2 \\
			3  & 8 & 7 & 1 & 0 \\
			-1 & 0 & 1 & 2 & 3 \\
			4  & 2 & 1 & -1 & 2  
		\end{pmatrix}, 
	\end{equation*}
and the kernel 
	\begin{equation*}
		K = 
		\begin{pmatrix}
			1 & 2 & 3 \\ 
			4 & 5 & 6 \\
			7 & 8 & 9
		\end{pmatrix}.
	\end{equation*}
	The computation of $[Y \ast K]_{1,1}$ can be illustrated as follows
	\begin{equation*}
		[Y \ast K]_{1,1} = 
		\begin{pmatrix}
			+1\cdot 9  & +5 \cdot 8 & -2\cdot 7 & \textcolor{gray}{0} & \textcolor{gray}{2} \\
			+3 \cdot 6  & +8\cdot 5 & +7\cdot 4 & \textcolor{gray}{1} & \textcolor{gray}{0} \\
			-1 \cdot 3 & +0\cdot 2 & +1\cdot 1 & \textcolor{gray}{2} & \textcolor{gray}{3} \\
			\textcolor{gray}{4}  & \textcolor{gray}{2} & \textcolor{gray}{1} & \textcolor{gray}{-1} & \textcolor{gray}{2}  
		\end{pmatrix}
		= 9 +40 -14 + 18 +40 +28 - 3 + 0 +1 = 119.
	\end{equation*}
	The gray values of $Y$ are not used in the computation.
	Here, we see that $K$ is flipped when used in the convolution. This also clarifies, how the (cross) correlation can be more intuitive, where  
	\begin{equation*}
		[Y \circledast K]_{1,1} = 
		\begin{pmatrix}
			+1\cdot 1  & +5 \cdot 2 & -2\cdot 3 & \textcolor{gray}{0} & \textcolor{gray}{2} \\
			+3 \cdot 4  & +8\cdot 5 & +7\cdot 6 & \textcolor{gray}{1} & \textcolor{gray}{0} \\
			-1 \cdot 7& +0\cdot 8 & +1\cdot 9 & \textcolor{gray}{2} & \textcolor{gray}{3} \\
			\textcolor{gray}{4}  & \textcolor{gray}{2} & \textcolor{gray}{1} & \textcolor{gray}{-1} & \textcolor{gray}{2}  
		\end{pmatrix}
		= 1 + 10 - 6 + 12 +40 + 42 -7 + 0 +9 = 101.
	\end{equation*}
	In a similar way we can proceed to calculate the remaining values by shifting the kernel over the matrix
	\begin{equation*}
	[Y \circledast K]_{1,2} = 
	\begin{pmatrix}
		\textcolor{gray}{1}  & +5 \cdot 1 & -2\cdot 2 & +0\cdot 3 & \textcolor{gray}{2} \\
		\textcolor{gray}{3}  & +8\cdot 4 & +7\cdot 5 & +1 \cdot 6 & \textcolor{gray}{0} \\
		\textcolor{gray}{-1} & +0\cdot 7 & +1\cdot 8 & +2 \cdot 9 & \textcolor{gray}{3} \\
		\textcolor{gray}{4}  & \textcolor{gray}{2} & \textcolor{gray}{1} & \textcolor{gray}{-1} & \textcolor{gray}{2}  
	\end{pmatrix}
	= 5 -4 + 0 +32 +35 +6 + 0 + 8 +18= 100 .
	\end{equation*}	
	Altogether, we get 
	\begin{equation*}
		Y \ast K = 
		\begin{pmatrix}
			119 & 120 & 53  \\
			155 & 155 & 102
		\end{pmatrix} \qquad \text{and} \qquad 
		Y \circledast K = 
		\begin{pmatrix}
			101 & 100 & 87 \\
			95 & 55 & 58
		\end{pmatrix}.
	\end{equation*}
	Especially, $Y \ast K \neq Y \circledast K$.
\end{example}

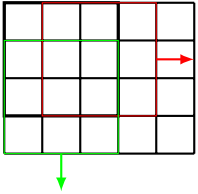
\begin{figure}[h!]
	\begin{minipage}{0.3\textwidth}
		\centering 
		\begin{tikzpicture}
			\draw[step=5mm,black,thick] (0,0) grid (2.5, 2); 
			\draw[step=5mm,black, very thick] (0,0.5) rectangle (1.5, 2); 
			\draw[step=5mm,red] (0.5,0.5) rectangle (2, 2); 
			\draw[step=5mm,green] (0,0) rectangle (1.5, 1.5); 
			\draw[thick,->,red] (2, 1.25) -- (2.5, 1.25);
			\draw[thick,->,green] (0.75, 0) -- (0.75, -0.5);
		\end{tikzpicture}
	\end{minipage}
	\begin{minipage}{0.7\textwidth}
		\caption{An image of size $4 \times 5$ is divided in blocks of size $3 \times 3$ by moving one pixel at a time either horizontally or vertically, as shown exemplary in red and green. Here, the black square is denoted by the index $(1,1)$, the red one by $(1,2)$ and the green one by $(2,1)$.}
		\label{fig:3times3}
	\end{minipage}
\end{figure}

The kernel size, which is typically square, e.g. $m \times m$, is a hyperparameter of the CNN. Furthermore, the convolutional layer has additional hyperparameters that need to be chosen. We have seen that a convolution with a $m \times m$ kernel reduces the dimension from $n_1\times n_2$ to $n_1 - m + 1 \times n_2 - m + 1$. To retain the image dimension we can use \textbf{(zero) padding}, cf. \cite{cs231n}, applied to the input $Y$ with $p \in \mathbb{N}_0$. Choosing $p=0$ yields $Y$ again, whereas $p=1$ results in
	\begin{equation*}
	\hat Y = 
	\begin{pmatrix}
		0 & 0 & 0 & 0 & 0 & 0 & 0\\
		0 & 1 & 5 &-2 & 0 & 2 & 0\\
		0 & 3 & 8 & 7 & 1 & 0 & 0\\
		0 &-1 & 0 & 1 & 2 & 3 & 0 \\
		0 & 4 & 2 & 1 &-1 & 2 & 0 \\
		0 & 0 & 0 & 0 & 0 & 0 & 0 
	\end{pmatrix},
\end{equation*}
for $Y$ from Example \ref{ex:kernel}. Consequently, the padded matrix $\hat Y$ is of dimension $(n_1 +2p) \times (n_2+2p) $. To retain the image dimension, we need to choose $p$ so that 
\begin{align*}
(n_1 + 2p) - m + 1 &= n_1, \\
(n_2 + 2p) - m + 1 &= n_2 ,
\end{align*}
i.e. $p = \frac{m-1}{2}$, which is possible for any odd $m$. 

\begin{figure}[h!]
	\begin{minipage}{0.3\textwidth}
		\centering 
		\begin{tikzpicture}
			\draw[step=5mm,black,thick] (0,0) grid (3.5, 3.5); 
			\draw[step=5mm,black, very thick] (0,2) rectangle (1.5, 3.5); 
			\draw[step=5mm,red] (1,2) rectangle (2.5, 3.5); 
			\draw[step=5mm,green] (0,1) rectangle (1.5, 2.5); 
			\draw[thick,->,red] (2.5, 2.75) -- (3, 2.75);
			\draw[thick,->,green] (0.75, 1) -- (0.75, 0.5);
		\end{tikzpicture}
	\end{minipage}
	\begin{minipage}{0.7\textwidth}
		\caption{A visualization of the convolution of a $7 \times 7$ images with a $3 \times 3$ kernel and stride $s=2$.}
		\label{fig:3times3_2}
	\end{minipage}
\end{figure}
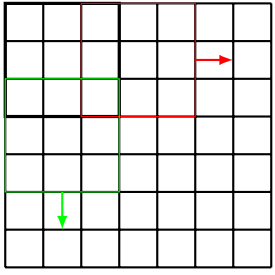

Furthermore, we can choose the \textbf{stride} $s\in\mathbb{N}$, which indicates how far to move the kernel. For example, in Figure \ref{fig:3times3} the stride is chosen as $s = 1$, while the stride in Figure \ref{fig:3times3_2} is $s=2$.

Let us remark that a stride $s>1$ reduces the output dimension of the convolution to 
\[\left(\frac{n_1-m}{s}+1\right) \times \left(\frac{n_2-m}{s}+1\right). \] 

Altogether, we can describe the convolutional layer. It consists of $M \in \mathbb{N}$ filters with identical hyperparameters: kernel size $m \times m$, padding $p$ and stride $s$, but each of them has its own learnable kernel $K$. Consequently, the filters in this layer have $M\cdot m^2$ variables in total.  Applying all $M$ filters to an input matrix $Y \in \mathbb{R}^{n_1 \times n_2}$ leads to an output of size
\[ \left( \frac{n_1 + 2p - m}{s} +1 \right) \times \left( \frac{n_2 + 2p - m}{s} +1 \right) \times M, \]
where the results for all $M$ filters are stacked, cf. \cite{cs231n}. 
Typically, the \textbf{depth} $M$ is chosen as a power of 2, and growing for deeper layers, while height and width are shrinking, cf. Figure \ref{fig:alex}.

Obviously, the output is a tensor with three dimensions, hence the subsequent layers need to process 3-tensor-valued data. In fact, for colored images already the original input of the network is a tensor. The (cross) correlation operation (and also the convolution operation) can be generalized to this case in the following way. 

\begin{figure}[h!]
	\centering
	\includegraphics[width=0.8\textwidth]{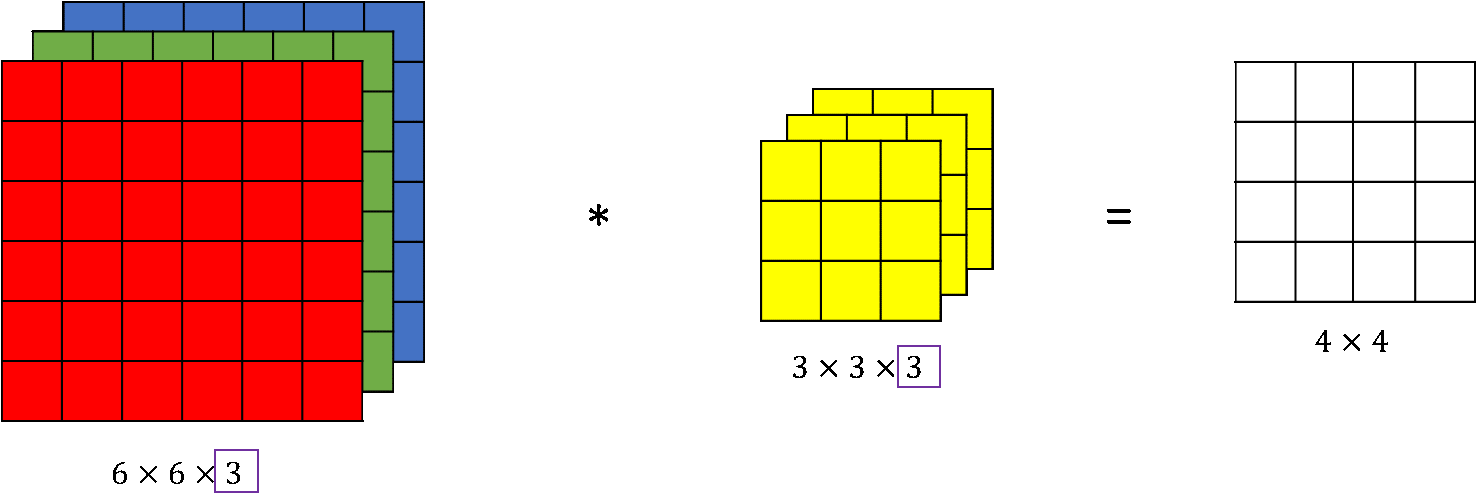}
	\caption{Illustration of a convolution on a tensor, specifically a colored image (with red, green, blue color channels) with a three-dimensional kernel and its result, which is a matrix. Here, no padding $p=0$ and a single stride $s=1$ is employed. Image Source: \url{https://datahacker.rs/convolution-rgb-image/}.}
	\label{fig:rgb_conv}
\end{figure}

Assume we have an input tensor of size $ n_1 \times n_2 \times n_3,$ then we choose a three dimensional kernel of size 
\[ m \times m \times n_3 ,\]
i.e. the depth coincides. No striding or padding is applied in the third dimension. Hence, the output is of dimension
\[ \left( \frac{n_1 + 2p - m}{s} +1 \right) \times \left( \frac{n_2 + 2p - m}{s} +1 \right) \times 1, \]
which can be understood as a matrix by discarding the redundant third dimension, cf. Figure \ref{fig:rgb_conv}.
Doing this for $M$ filters, again leads to the output being a 3-tensor. 
	
\begin{figure}[ht!]
	\begin{minipage}{0.4\textwidth}
		\centering
		\begin{tikzpicture}
			\node[thick,circle,draw=black,fill=myorange!30,minimum size=22,inner sep=0.5,outer sep=0.6] (1) at (0,4) {$y^{[0]}_1$};
			\node[thick,circle,draw=black,fill=myorange!30,minimum size=22,inner sep=0.5,outer sep=0.6] (2) at (0,3) {$y^{[0]}_2$};
			\node[thick,circle,draw=black,fill=myorange!30,minimum size=22,inner sep=0.5,outer sep=0.6] (3) at (0,2) {$y^{[0]}_3$};
			\node[thick,circle,draw=black,fill=myorange!30,minimum size=22,inner sep=0.5,outer sep=0.6] (4) at (0,1) {$y^{[0]}_4$};
			\node[thick,circle,draw=black,fill=myorange!30,minimum size=22,inner sep=0.5,outer sep=0.6] (5) at (0,0) {$y^{[0]}_5$};
			\node[thick,circle,draw=black,minimum size=22,inner sep=0.5,outer sep=0.6] (6) at (2,4) {$y^{[1]}_1$};
			\node[thick,circle,draw=black,fill=myorange!30,minimum size=22,inner sep=0.5,outer sep=0.6] (7) at (2,3) {$y^{[1]}_2$};
			\node[thick,circle,draw=black,fill=myorange!30,minimum size=22,inner sep=0.5,outer sep=0.6] (8) at (2,2) {$y^{[1]}_3$};
			\node[thick,circle,draw=black,fill=myorange!30,minimum size=22,inner sep=0.5,outer sep=0.6] (9) at (2,1) {$y^{[1]}_4$};
			\node[thick,circle,draw=black,minimum size=22,inner sep=0.5,outer sep=0.6] (10) at (2,0) {$y^{[1]}_5$};
			\node[thick,circle,draw=black,minimum size=22,inner sep=0.5,outer sep=0.6] (11) at (4,4) {$y^{[2]}_1$};
			\node[thick,circle,draw=black,minimum size=22,inner sep=0.5,outer sep=0.6] (12) at (4,3) {$y^{[2]}_2$};
			\node[thick,circle,draw=black,fill=myorange!30,minimum size=22,inner sep=0.5,outer sep=0.6] (13) at (4,2) {$y^{[2]}_3$};
			\node[thick,circle,draw=black,minimum size=22,inner sep=0.5,outer sep=0.6] (14) at (4,1) {$y^{[2]}_4$};
			\node[thick,circle,draw=black,minimum size=22,inner sep=0.5,outer sep=0.6] (15) at (4,0) {$y^{[2]}_5$};
			\draw[-{Latex[length=4,width=3.5]},thick,shorten <=0.5,shorten >=1] (1.east) to (6);
			\draw[-{Latex[length=4,width=3.5]},thick,shorten <=0.5,shorten >=1] (1.east) to (7);
			\draw[-{Latex[length=4,width=3.5]},thick,shorten <=0.5,shorten >=1] (2.east) to (6);
			\draw[-{Latex[length=4,width=3.5]},thick,shorten <=0.5,shorten >=1] (2.east) to (7);
			\draw[-{Latex[length=4,width=3.5]},thick,shorten <=0.5,shorten >=1] (2.east) to (8);
			\draw[-{Latex[length=4,width=3.5]},thick,shorten <=0.5,shorten >=1] (3.east) to (7);
			\draw[-{Latex[length=4,width=3.5]},thick,shorten <=0.5,shorten >=1] (3.east) to (8);
			\draw[-{Latex[length=4,width=3.5]},thick,shorten <=0.5,shorten >=1] (3.east) to (9);
			\draw[-{Latex[length=4,width=3.5]},thick,shorten <=0.5,shorten >=1] (4.east) to (8);
			\draw[-{Latex[length=4,width=3.5]},thick,shorten <=0.5,shorten >=1] (4.east) to (9);
			\draw[-{Latex[length=4,width=3.5]},thick,shorten <=0.5,shorten >=1] (4.east) to (10);
			\draw[-{Latex[length=4,width=3.5]},thick,shorten <=0.5,shorten >=1] (5.east) to (9);
			\draw[-{Latex[length=4,width=3.5]},thick,shorten <=0.5,shorten >=1] (5.east) to (10);
			\draw[-{Latex[length=4,width=3.5]},thick,shorten <=0.5,shorten >=1] (6.east) to (11);
			\draw[-{Latex[length=4,width=3.5]},thick,shorten <=0.5,shorten >=1] (6.east) to (12);
			\draw[-{Latex[length=4,width=3.5]},thick,shorten <=0.5,shorten >=1] (7.east) to (11);
			\draw[-{Latex[length=4,width=3.5]},thick,shorten <=0.5,shorten >=1] (7.east) to (12);
			\draw[-{Latex[length=4,width=3.5]},thick,shorten <=0.5,shorten >=1] (7.east) to (13);
			\draw[-{Latex[length=4,width=3.5]},thick,shorten <=0.5,shorten >=1] (8.east) to (12);
			\draw[-{Latex[length=4,width=3.5]},thick,shorten <=0.5,shorten >=1] (8.east) to (13);
			\draw[-{Latex[length=4,width=3.5]},thick,shorten <=0.5,shorten >=1] (8.east) to (14);
			\draw[-{Latex[length=4,width=3.5]},thick,shorten <=0.5,shorten >=1] (9.east) to (13);
			\draw[-{Latex[length=4,width=3.5]},thick,shorten <=0.5,shorten >=1] (9.east) to (14);
			\draw[-{Latex[length=4,width=3.5]},thick,shorten <=0.5,shorten >=1] (9.east) to (15);
			\draw[-{Latex[length=4,width=3.5]},thick,shorten <=0.5,shorten >=1] (10.east) to (14);
			\draw[-{Latex[length=4,width=3.5]},thick,shorten <=0.5,shorten >=1] (10.east) to (15);
		\end{tikzpicture}
	\end{minipage}
	\begin{minipage}{0.6\textwidth}
		\caption{Simplified CNN architecture with an input layer $y^{[0]}$ and two subsequent convolutional layers $y^{[1]}$ and $y^{[2]}$, each with a one dimensional kernel of size $m=3$, stride $s=1$ and zero padding $p=1$.  The colored nodes are the receptive field of the neuron $y^{[2]}_3$.}
		\label{fig:receptive}
	\end{minipage}	
\end{figure}
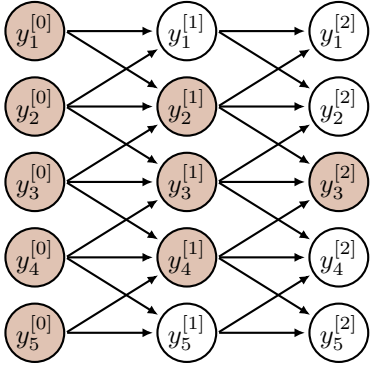

\begin{remark}$\,$
	\begin{enumerate}
		\item[(i)] Convolutional layers have the advantage that they have less variables than fully connected layers applied to the flattened image. For example consider a grayscale image of size $28 \times 28$ as input in LeNet, cf. Figure \ref{fig:LeNet}. The first convolution has 6 kernels with $5\times 5$ entries. Due to padding with $p=2$ the output is $28\times 28 \times 6$, so the image size is retained. Additionally, before applying a detector layer, we add a bias per channel, so 6 bias variables in this case. In total, we have to learn \textbf{156} variables.
		Now, imagine this image is flattened to a $784\times 1$ vector and fed into a FNN with fully connected layer, where we also want to retain the size, i.e. the first hidden layer has $784$ nodes. This results in a much larger number of variables:
		\[ \underbrace{784\cdot784}_{weight}+\underbrace{784}_{bias} = \textbf{615440}.\]
		\item[(ii)] Directly related, a disadvantage of convolutional layers is that every output only sees a subset of all input neurons, cf. e.g. \cite[Section 9.2]{goodfellow2016deep}. We denote this set of seen inputs by \textbf{effective receptive field} of the neuron. In an FNN with fully connected layers the effective receptive field of a neuron is the entire input. However, the receptive field of a neuron increases with depth of the network, as illustrated in Figure \ref{fig:receptive}.
	\end{enumerate}	
\end{remark}

\subsection{Detector Layer}
\label{subsec:detlayers}
In standard CNN architecture after a convolutional layer, a detector layer is applied. This simply means performing an activation function. To this end, we extend the activation function $\sigma:\mathbb{R} \rightarrow \mathbb{R}$ to matrix and tensor valued inputs by applying it component-wise, as we did for vectors before in Section \ref{sec:FNN}, e.g. for a 3-tensor $Y = \{Y_{i,j,k}\}_{i,j,k}$ with $i=1,\ldots,n_1, j=1,\ldots,n_2, k=1,\ldots,n_3$ we get 
\[ \left(\sigma(Y)\right)_{i,j,k} = \sigma(Y_{i,j,k}). \]

\subsection{Pooling Layer}
\label{subsec:poollayers}
After the detector layer, typically a pooling layer (also called downsampling layer) is applied, cf. e.g. \cite[Section 9.3]{goodfellow2016deep}. This layer type is responsible for reducing the first two dimensions (height and width) and usually does not interfere with the third dimension (depth) of the data $Y$, but rather is applied for all channels independently. Consequently, the depth of the output coincides with the depth of the input and we omit the depth in our discussion. 

As in convolutional layers, pooling layers have a filter size $m \times m$, stride $s$ and padding $p$. However, almost always $p=0$ is chosen. The most popular values for for the filter size and stride are $m=s=2$. Again, with an input of size $n_1 \times n_2$ the output dimension is 
\[ \left( \frac{n_1 + 2p -m}{s} +1 \right) \times \left( \frac{n_1 + 2p -m}{s} +1 \right) \quad \stackrel{m=s=2, p=0}{=} \quad  \frac{n_1}{2} \times \frac{n_2}{2}.  \]

One common choice is \textbf{Max Pooling} (or Maximum Pooling), where the largest value is selected, cf. e.g. \cite{cs231n}. Below we see an example of max pooling with a $2 \times 2$ kernel, stride $s=2$ and no padding. 

\begin{equation*}
	\left(
	\begin{array}{c c | c c}
		1 & 3  & 0 & -7 \\
		-2 & \textcolor{myorange}{4} & \textcolor{mygreen}{1} & -1 \\
		\hline 
		0 & 1 & \textcolor{HeiRot}{8} & -3 \\
		\textcolor{myblue}{2} & 0 & 4 & 5 
	\end{array}
	\right)
	\stackrel{\max}{\longrightarrow}
	 	\left(
	 \begin{array}{c | c}
	 	\textcolor{myorange}{4} & \textcolor{mygreen}{1}  \\
	 	\hline 
	 	\textcolor{myblue}{2} & \textcolor{HeiRot}{8} 
	 \end{array}
	 \right)
\end{equation*}

Another common choice is \textbf{Average Pooling}, where we take the mean of all values. Below we see an example of average pooling (abbreviated: "avg") with a $2 \times 2$ kernel,
\[ K = \begin{pmatrix}
	0.25 & 0.25 \\
	0.25 & 0.25
\end{pmatrix},\]
stride $s=2$ and no padding. 
\begin{equation*}
	\hspace{1.2cm}
	\left(
	\begin{array}{c c | c c}
		\textcolor{myorange}{1} & \textcolor{myorange}{3}  & \textcolor{mygreen}{0} & \textcolor{mygreen}{-7} \\
		\textcolor{myorange}{-2} & \textcolor{myorange}{4} & \textcolor{mygreen}{1} & \textcolor{mygreen}{-1} \\
		\hline 
		\textcolor{myblue}{0} & \textcolor{myblue}{1} & \textcolor{HeiRot}{8} & \textcolor{HeiRot}{-3} \\
		\textcolor{myblue}{2} & \textcolor{myblue}{0} & \textcolor{HeiRot}{4} & \textcolor{HeiRot}{5} 
	\end{array}
	\right)
	\stackrel{\text{avg}}{\longrightarrow}
	\left(
	\begin{array}{c | c}
		\textcolor{myorange}{1.50}  & \textcolor{mygreen}{-1.75}  \\
		\hline 
		\textcolor{myblue}{0.75} & \textcolor{HeiRot}{3.50} 
	\end{array}
	\right)
\end{equation*}

\begin{figure}[h!]
	\centering
	\includegraphics[width=0.49\textwidth]{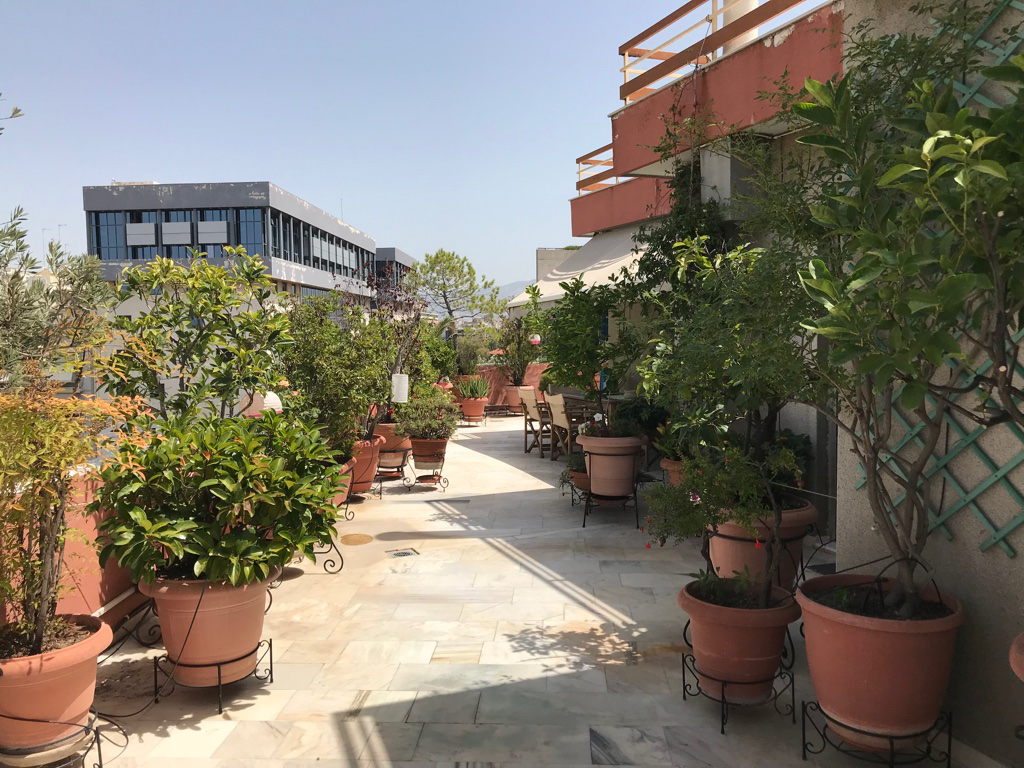}
	\includegraphics[width=0.49\textwidth]{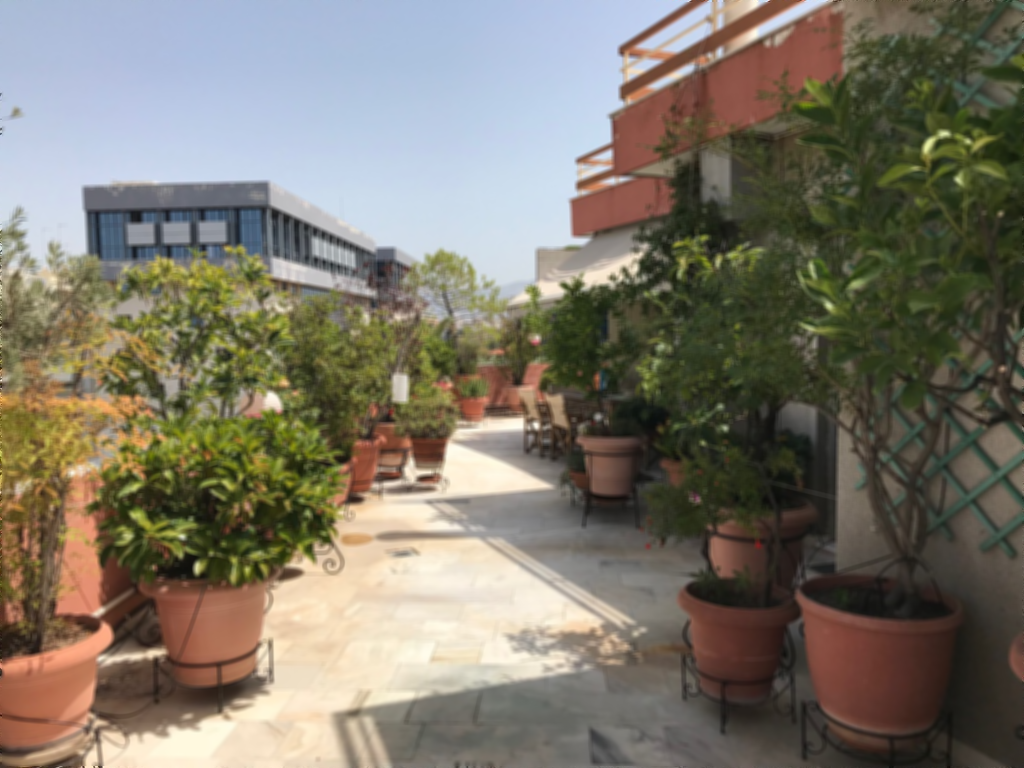}
	\caption{Original image (left) and blurred image produced by average pooling (right) with a $5 \times 5$ kernel, stride $s=1$ and zero padding with $p=2$. Image Source: Laurin Ernst.}
	\label{fig:blurr}
\end{figure}

The effect of average pooling applied to an image is easily visible: It blurs the image. In the new image every pixel is an average of a pixel and its neighboring $m^2-1$ pixels, see Figure \ref{fig:blurr}. Depending on the choice of stride $s$ and padding $p$, the blurred image may also have less pixels. 

\begin{remark}$\,$
	\begin{enumerate}
		\item[(i)] Pooling layers do not contain variables to learn.
		\item[(ii)] We have seen that when using CNNs, we make the following assumptions:
		\begin{enumerate}
			\item[(a)] Pixels far away from each other do not need to interact with each other. 
			\item[(b)] Small translations are not relevant. 
		\end{enumerate}
		If these assumptions do not hold, employing a CNN can result in underfitting.
	\end{enumerate}
\end{remark}

\subsection{Local Response Normalization}
\label{subsec:LRN}

\begin{figure}[h!]
	\centering 
	\includegraphics[width=0.9\textwidth]{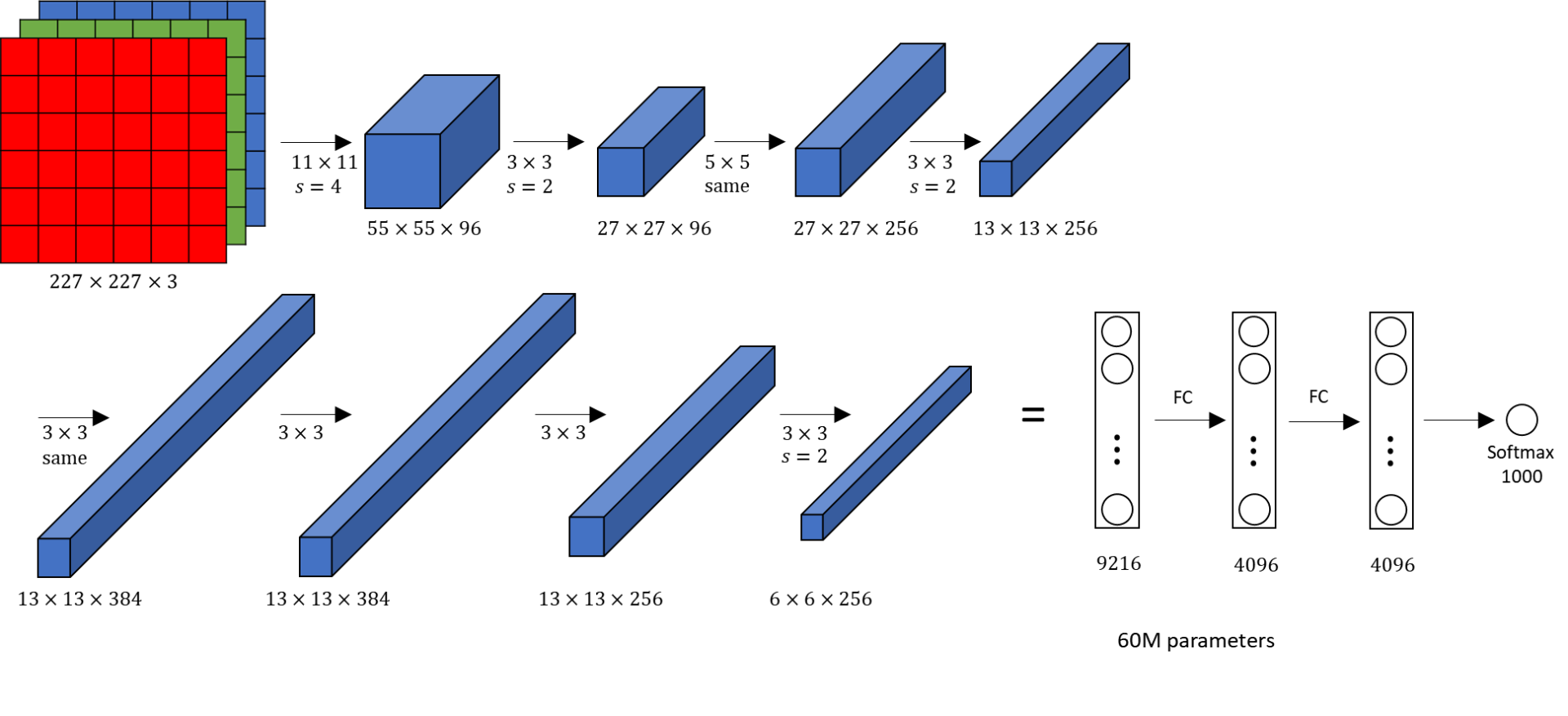}
	\caption{Architecture of AlexNet. ReLU activation is employed in the hidden layers. Image Source: \url{https://datahacker.rs/deep-learning-alexnet-architecture/}.}
	\label{fig:alex}
\end{figure}

Similar to batch normalization, Local Response Normalization (LRN) \cite[Section 3.3]{alexnet} stabilizes training with unbounded activation functions like ReLU. 
This strategy was first introduced within the "AlexNet" architecture \cite{alexnet}, cf. Figure \ref{fig:alex}, because contrary to previous CNNs like "LeNet-5", which used sigmoid activation, "AlexNet" employs ReLU activation.

\begin{figure}[h!]
	\centering
	\includegraphics[width=0.7\textwidth]{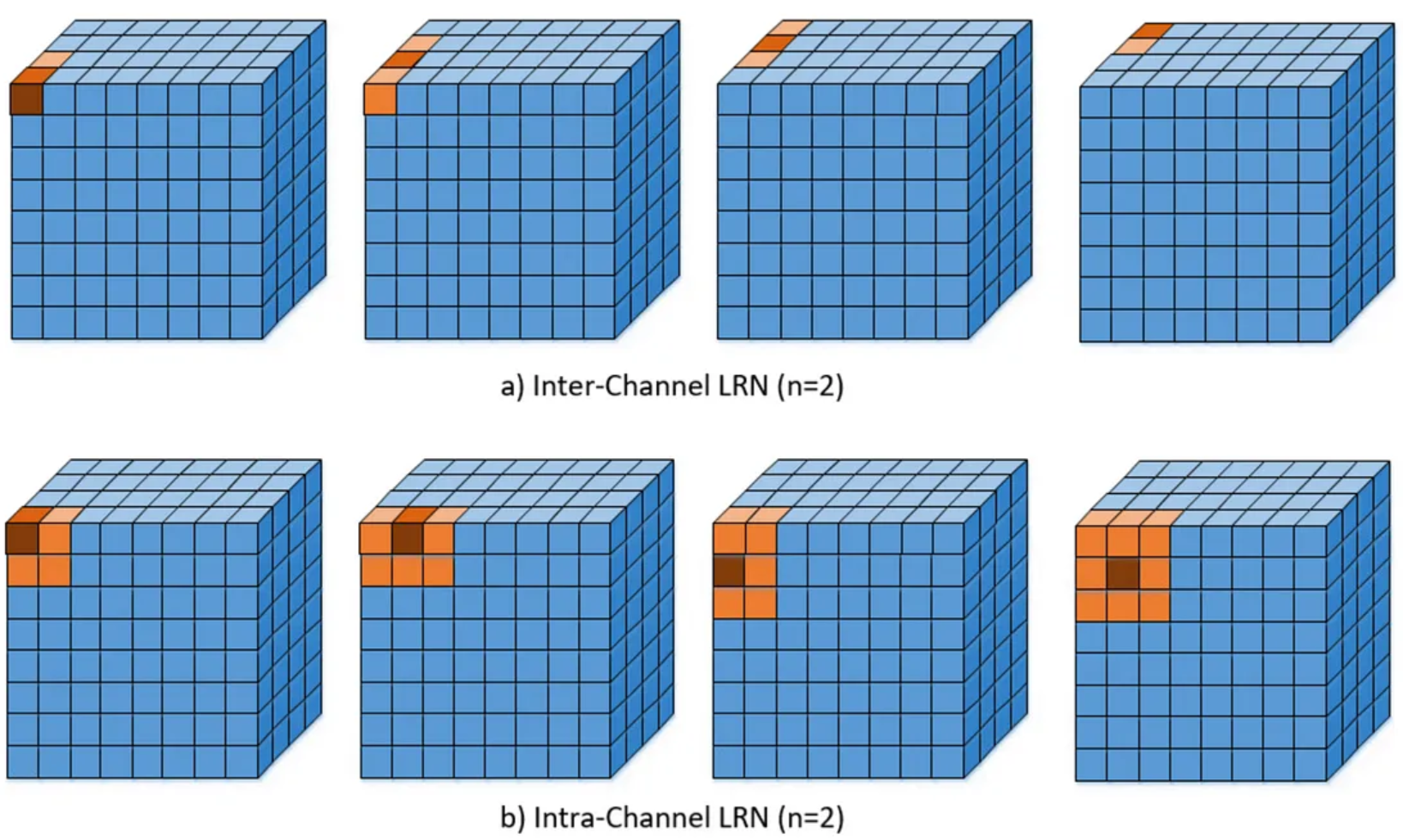}
	\caption{Illustration of local response normalization. Inter-channel version a) as introduced in \cite[Section 3.3]{alexnet} and intra-channel version b). Both for $n=2$. For clarification: The red pixel in the top left cube is $Y_{1,1,1}$, while the red pixel in the top row, second from the left cube is $Y_{1,1,2}$ and the red pixel in the bottom row, second from the left cube is $Y_{1,2,1}$. Image source: \url{https://towardsdatascience.com/difference-between-local-response-normalization-and-batch-normalization-272308c034ac}. }
	\label{fig:LRN}
\end{figure}

The inter-channel LRN, as introduced in \cite[Section 3.3]{alexnet}, see also Figure \ref{fig:LRN} a), is given by  
\begin{equation*}
	\hat{Y}_{i,j,k} = \frac{Y_{i,j,k}}{ \left( \kappa + \gamma \sum\limits_{m = \max(1,k-\frac{n}{2})}^{\min(M,k+\frac{n}{2})} (Y_{i,j,m})^2 \right)^\beta }.
\end{equation*}

Here, $Y_{i,j,k}$ and $\hat{Y}_{i,j,k}$ denote the activity of the neuron before and after normalization, respectively. The indices $i,j,k$ indicate the height, width and depth of $Y$. We have $i=1,\ldots,n_1$, $j=1,\ldots,n_2$ and $k=1,\ldots,M$, where $M$ is the number of filters in the previous convolutional layer. The values $\kappa, \gamma, \beta, n \in \mathbb{R}$ are hyperparameters, where $\kappa$ is used to avoid singularities, and $\gamma$ and $\beta$ are called normalization and contrasting constants, respectively. Furthermore, $n$ dictates how many surrounding neurons are taken into consideration, see also Figure \ref{fig:LRN}. In \cite{alexnet} $\kappa = 2, \gamma = 10^{-4}, \beta = 0.75$ were chosen.

In the case of intra-channel LRN, the neighborhood is extended within the same channel. This leads to the following formula
\begin{equation*}
	\hat{Y}_{i,j,k} = \frac{Y_{i,j,k}}{ \left( \kappa + \gamma \sum\limits_{p = \max(1,i-\frac{n}{2})}^{\min(n_1,i+\frac{n}{2})} \sum\limits_{q = \max(1,j-\frac{n}{2})}^{\min(n_2,j+\frac{n}{2})} (Y_{p,q,k})^2 \right)^\beta }.
\end{equation*}

\begin{remark}
	The LRN layer is non-trainable, since it only contains hyperparameters and no variables. 
\end{remark}


\newpage
\section{ResNet}
\label{sec:ResNet}
We have seen in Example \ref{ex:FNN_grad} that for a FNN with depth $L$ we have the derivative
	\[\frac{\partial \mathscr{L}}{\partial W^{[\ell]}} =  \frac{\partial \mathscr{L}}{\partial y^{[L]}} \cdot 
	\prod_{j=L}^{\ell+2} \frac{\partial y^{[j]}}{\partial y^{[j-1]}} 
	\cdot  \frac{\partial y^{[\ell+1]}}{\partial W^{[\ell]}}.\]
In the case that we consider a very deep network, i.e. large $L$, the product in the derivative can be problematic, \cite{bengio1994learning,glorot2010understanding}, especially if we take derivatives with respect to variables from early layers. Two cases may occur:
\begin{enumerate}
	\item If $  \frac{\partial y^{[j]}}{\partial y^{[j-1]}} < 1$ for all $j$, the product, and hence the whole derivative, tends to zero for growing $L$. This problem is referred to as \textbf{vanishing gradient}.
	\item On the other hand, if $  \frac{\partial y^{[j]}}{\partial y^{[j-1]}} > 1$ for all $j$, the product, and hence the whole derivative, tends to infinity for growing $L$. This problem is referred to as \textbf{exploding gradient}.
\end{enumerate}

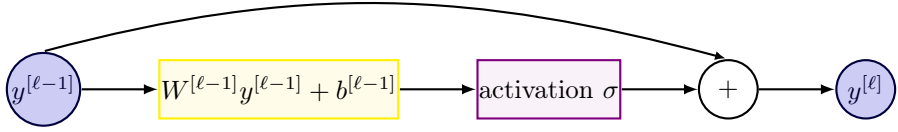
\begin{figure}[h!]
\begin{center}
	\begin{tikzpicture}
		\node[node 2] (1) at (0,0) {$y^{[\ell-1]}$};
		\draw [thick, ->] (1.east) -- ++(1,0)
		node[right,thick,rectangle,draw=yellow,fill=yellow!10,minimum size=22,inner sep=0.5,outer sep=0.6] (2) {$ W^{[\ell-1]} y^{[\ell-1]} + b^{[\ell-1]}$};
		\draw [thick, ->] (2.east) -- ++(1,0)
		node[right,thick,rectangle,draw=violet,fill=violet!5,minimum size=22,inner sep=0.5,outer sep=0.6] (3) {activation $\sigma$};
		\draw [thick, ->] (3.east) -- ++(1,0)
		node[right,thick,circle,draw=black,minimum size=22] (4) {$+$};
		\draw[-{Latex[length=4,width=3.5]},thick,shorten <=0.5,shorten >=1] (1.north) to [out=15,in=165] (4.north);
		\draw [thick, ->] (4.east) -- ++(1,0)
		node[right,node 2] (5) {$y^{[\ell]}$};
	\end{tikzpicture}
\end{center}
\caption{Illustration of a simplified ResNet layer. }
\label{fig:ResNetlayer}
\end{figure}

\textbf{Residual Networks (ResNets)} have been developed in \cite{he2016deep,he2016identity} with the intention to solve the vanishing gradient problem. Employing the same notation as in FNNs, simplified ResNet layers can be represented in the following way
\begin{equation}\label{eq:ResNet}
	y^{[\ell]} = y^{[\ell-1]} + \sigma^{[\ell]} ( W^{[\ell-1]}y^{[\ell-1]} + b^{[\ell-1]} ) \qquad \text{for } \ell=1,\ldots,L,
\end{equation}
with $y^{[0]} = u$ the input data. Essentially, a ResNet is a FNN with an added \textbf{skip connection}, i.e. $+y^{[\ell-1]}$, cf. Figure \ref{fig:ResNetlayer} 

\begin{remark}
	The ResNet layers in the current form \eqref{eq:ResNet} only work, if all feature vectors $y^{[\ell]}$ have the same dimension $n_{\ell}$, so that we can add them up. To allow for different layer sizes, we need to insert projection operators $P_{\ell-1}^{\ell} \in \mathbb{R}^{n_{\ell} \times n_{\ell-1}}$, cf. \cite[Section 4]{antildiazherberg}, i.e.
	\begin{equation*}
		y^{[\ell]} = P_{\ell-1}^{\ell} y^{[\ell-1]} + \sigma^{[\ell]} ( W^{[\ell-1]}y^{[\ell-1]} + b^{[\ell-1]} ) \qquad \text{for } \ell=1,\ldots,L,
	\end{equation*}
\end{remark}

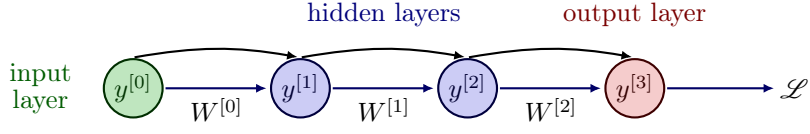
\begin{figure}[h!]
	\begin{center}
		\begin{tikzpicture}[x=2.2cm,y=1.4cm]
			\message{^^JNeural network with arrows}
			\readlist\Nnod{1,1,1,1} 
			\message{^^J  Layer}
			\foreachitem \N \in \Nnod{ 
				\edef\lay{\Ncnt} 
				\message{\lay,}
				\pgfmathsetmacro\prev{int(\Ncnt-1)} 
				\foreach \i [evaluate={\y=\N/2-\i; \x=\lay; \n=\nstyle;}] in {1,...,\N}{ 
					\node[node \n] (N\lay-\i) at (\x,\y) {$y^{[\prev]}$};
					\ifnum\lay>1
					\foreach \j in {1,...,\Nnod[\prev]}{ 
						\draw[connect arrow] (N\prev-\j) -- (N\lay-\i); 
					}
					\fi 
				}
			}
			\draw[connect arrow] (N\Nnodlen-1) -- (4.85,-0.5) ;
			\draw[-{Latex[length=4,width=3.5]},thick,shorten <=0.5,shorten >=1] (N1-1.north) to [out=10,in=170] (N2-1.north);
			\draw[-{Latex[length=4,width=3.5]},thick,shorten <=0.5,shorten >=1] (N2-1.north) to [out=10,in=170] (N3-1.north);
			\draw[-{Latex[length=4,width=3.5]},thick,shorten <=0.5,shorten >=1] (N3-1.north) to [out=10,in=170] (N4-1.north);
			\node[left of =5,align=center,mygreen!60!black] at (0.9,-0.5) {input\\[-0.2em]layer};
			\node[above=4,align=center,myblue!60!black] at (2.5,-0.1) {hidden layers};
			\node[right of=8,align=center] at (4.5,-0.5) {$\mathscr{L}$};
			\node[above=8,align=center,myred!60!black] at (4,-0.2) {output layer};	
			\node at (1.5,-0.7) {$W^{[0]}$};
			\node at (2.5,-0.7) {$W^{[1]}$};
			\node at (3.5,-0.7) {$W^{[2]}$};
		\end{tikzpicture}
	\end{center}
	\caption{A ResNet with 2 hidden layers, one node per layer and depth $L=3$.}
	\label{fig:ResNet_simple}
\end{figure}

We now revisit the simple FNN with two hidden layers from Example \ref{ex:FNN_grad}, and add skip connections to make it a ResNet, see Figure \ref{fig:ResNet_simple}. 

\begin{example}
	Consider a simple ResNet with one node per layer and assume that we only consider weights $W^{[\ell]} \in \mathbb{R}$ and no biases. 
	For the network in Figure \ref{fig:ResNet_simple} we have  $\theta = (W^{[0]},W^{[1]},W^{[2]})^{\top}$, and 
	\[ \mathscr{L} (\theta) = \mathscr{L}(y^{[3]}(\theta)).\]
	We define for $\ell = 1,\ldots,L$
	\[ a^{[\ell]} := \sigma^{[\ell]}(W^{[\ell-1]}y^{[\ell-1]}),\]
	so that in the ResNet setup 
	\[y^{[\ell]} = y^{[\ell-1]} + a^{[\ell]}.\]
	Computing the components of the gradient, we employ the chain rule to obtain e.g. 
	\begin{align*}
		\frac{\partial \mathscr{L}}{\partial W^{[0]}} &= \frac{\partial \mathscr{L}}{\partial y^{[3]}} \cdot \frac{\partial y^{[3]}}{\partial W^{[0]}} \\
		&= \frac{\partial \mathscr{L}}{\partial y^{[3]}} \cdot \frac{\partial}{\partial W^{[0]}} (y^{[2]} + a^{[3]})\\
		&=\frac{\partial \mathscr{L}}{\partial y^{[3]}} \cdot \left(  \frac{\partial y^{[2]}}{\partial W^{[0]}} + \frac{\partial a^{[3]}}{\partial y^{[2]}} \cdot \frac{\partial y^{[2]}}{\partial W^{[0]}}\right) \\
		&= \frac{\partial \mathscr{L}}{\partial y^{[3]}} \cdot \left( \mathbb{I}  + \frac{\partial a^{[3]}}{\partial y^{[2]}} \right) \cdot \frac{\partial y^{[2]}}{\partial W^{[0]}}\\
		&=\frac{\partial \mathscr{L}}{\partial y^{[3]}} \cdot \left( \mathbb{I}  + \frac{\partial a^{[3]}}{\partial y^{[2]}} \right) \cdot \left( \mathbb{I}  + \frac{\partial a^{[2]}}{\partial y^{[1]}} \right) \cdot\frac{\partial y^{[1]}}{\partial W^{[0]}} ,
	\end{align*}
	where $\mathbb{I}$ denotes the identity. In general for depth $L$, we get 
	\begin{equation} \label{eq:ResNetder}
		\frac{\partial \mathscr{L}}{\partial W^{[\ell]}} =  \frac{\partial \mathscr{L}}{\partial y^{[L]}} \cdot 
	\prod_{j=L}^{\ell+2}  \left( \mathbb{I}  + \frac{\partial a^{[j]}}{\partial y^{[j-1]}} \right) 
	\cdot  \frac{\partial y^{[\ell+1]}}{\partial W^{[\ell]}}.
	\end{equation}
\end{example}

If we generalize the derivative \eqref{eq:ResNetder} to ResNet architectures, where we do not only consider weights $W^{[\ell]}$, see e.g. \cite[Theorem 6.1]{antildiazherberg}, the structure of the product in the derivative remains the same, i.e. it also contains an identity term. 

Remember that for FNNs it holds $y^{[j]} = a^{[j]}$, i.e. the fraction in the product coincides in both cases. However, due to the added identity, even if
\[  \frac{\partial a^{[j]}}{\partial y^{[j-1]}} < 1 \] 
holds for all $j$, we will not encounter vanishing gradients in the ResNet architecture. The exploding gradients problem can still occur.

We will see in Section \ref{subsec:ResNetversions} that there exist several versions of ResNets. However, from a mathematical point of view the simplified version \eqref{eq:ResNet} is especially interesting, because it can be related to ordinary differential equations (ODEs), as first done in \cite{haber2017stable}.
Inserting a parameter $\tau^{[\ell]} \in \mathbb{R}$ in front of the activation function $\sigma$ and rearranging the terms of the forward propagation delivers
\begin{align*}
	y^{[\ell]} &= y^{[\ell-1]} + \tau^{[\ell]} \sigma (W^{[\ell-1]} y^{[\ell-1]} + b^{[\ell-1]}) \\
	\Rightarrow \quad \frac{y^{[\ell]} - y^{[\ell-1]}}{\tau^{[\ell]}} &= \sigma (W^{[\ell-1]} y^{[\ell-1]} + b^{[\ell-1]}).
\end{align*}
Here, we consider the same activation function $\sigma$ for all layers. Now, the left hand side of the equation can be interpreted as a finite difference representation of a time derivative, where $\tau^{[\ell]}$ is the time step size and $y^{[\ell]}, y^{[\ell-1]}$ are the values attained at two neighboring points in time. This relation between ResNets and ODEs is also studied under the name of Neural ODEs, \cite{chen2018neural}. It is also possible to learn the time step size $\tau^{[\ell]}$ as an additional variable, \cite{antildiazherberg}.

Let us now introduce the different ResNet versions from the original papers, \cite{he2016deep,he2016identity}.

\subsection{Different ResNet Versions}
\label{subsec:ResNetversions}
\begin{figure}[h!]
	\begin{center}
		\begin{tikzpicture}
			\node[node 2] (1) at (0,0) {$y^{[\ell-1]}$};
			\draw [thick, ->] (1.east) -- ++(0.7,0)
			node[right,thick,rectangle,draw=yellow,fill=yellow!10,minimum size=22,inner sep=0.5,outer sep=0.6] (2) {Weights};
			\draw [thick, ->] (2.east) -- ++(0.7,0)
			node[right,thick,rectangle,draw=brown,fill=brown!10,minimum size=22, inner sep=0.5,outer sep=0.6] (3) {BN};
			\draw [thick, ->] (3.east) -- ++(0.7,0)
			node[right,thick,rectangle,draw=violet,fill=violet!5,minimum size=22, inner sep=0.5,outer sep=0.6] (4) {ReLU};
			\draw [thick, ->] (4.east) -- ++(0.7,0)
			node[right,thick,rectangle,draw=yellow,fill=yellow!10,minimum size=22,inner sep=0.5,outer sep=0.6] (5) {Weights};
			\draw [thick, ->] (5.east) -- ++(0.7,0)
			node[right,thick,rectangle,draw=brown,fill=brown!10,minimum size=22, inner sep=0.5,outer sep=0.6] (6) {BN};
			\draw [thick, ->] (6.east) -- ++(0.7,0)
			node[right,thick,circle,draw=black,minimum size=22] (7) {$+$};
			\draw [thick, ->] (7.east) -- ++(0.7,0)
			node[right,thick,rectangle,draw=violet,fill=violet!5,minimum size=22, inner sep=0.5,outer sep=0.6] (8) {ReLU};
			\draw [thick, ->] (8.east) -- ++(0.7,0)
			node[right,node 2] (9)  {$y^{[\ell]}$};
			\draw[-{Latex[length=4,width=3.5]},thick,shorten <=0.5,shorten >=1] (1.north) to [out=10,in=170] (7.north);
		\end{tikzpicture}
	\end{center}
	\caption{Illustration of a residual block as introduced in \cite{he2016deep}.}
	\label{fig:Residualblock}
\end{figure}
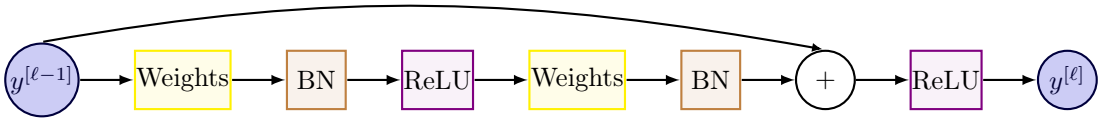

In contrast to the simplified ResNet layer version \eqref{eq:ResNet} that we introduced, original ResNet architectures \cite{he2016deep} consist of \textbf{residual blocks}, cf. Figure \ref{fig:Residualblock}. Here, different layers are grouped together into one residual block and then residual blocks are stacked to form a ResNet. 

In the residual block, cf. Figure \ref{fig:Residualblock}, we have the following layer types:
\begin{itemize}
	\item Weights: fully connected or convolutional layer,
	\item BN: Batch Normalization layer, cf. Section \ref{subsec:Batch},
	\item ReLU: activation function $\sigma = \text{ReLU}$.
\end{itemize} 
Clearly, the residual block and the simplified ResNet layer both contain a skip connection, which is the integral part of ResNets success, since it helps avoid the vanishing gradient problem. However, the residual block is less easy to interpret from a mathematical point of view and can not directly be related to ODEs.

In frameworks like Tensorflow and Pytorch, if you encounter a network architecture called "ResNet", it will usually be built by stacking residual blocks of this original form, Figure \ref{fig:Residualblock}. 

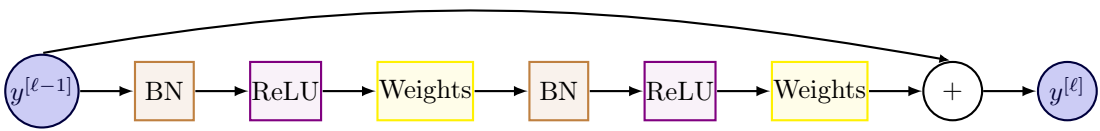
\begin{figure}[h!]
	\begin{center}
		\begin{tikzpicture}
			\node[node 2] (1) at (0,0) {$y^{[\ell-1]}$};
			\draw [thick, ->] (1.east) -- ++(0.7,0)
			node[right,thick,rectangle,draw=brown,fill=brown!10,minimum size=22,inner sep=0.5,outer sep=0.6] (2) {BN};
			\draw [thick, ->] (2.east) -- ++(0.7,0)
			node[right,thick,rectangle,draw=violet,fill=violet!5,minimum size=22, inner sep=0.5,outer sep=0.6] (3) {ReLU};
			\draw [thick, ->] (3.east) -- ++(0.7,0)
			node[right,thick,rectangle,draw=yellow,fill=yellow!10,minimum size=22, inner sep=0.5,outer sep=0.6] (4) {Weights};
			\draw [thick, ->] (4.east) -- ++(0.7,0)
			node[right,thick,rectangle,draw=brown,fill=brown!10,minimum size=22,inner sep=0.5,outer sep=0.6] (5) {BN};
			\draw [thick, ->] (5.east) -- ++(0.7,0)
			node[right,thick,rectangle,draw=violet,fill=violet!5,minimum size=22, inner sep=0.5,outer sep=0.6] (6) {ReLU};
			\draw [thick, ->] (6.east) -- ++(0.7,0)
			node[right,thick,rectangle,draw=yellow,fill=yellow!10,minimum size=22, inner sep=0.5,outer sep=0.6] (7) {Weights};
			\draw [thick, ->] (7.east) -- ++(0.7,0)
			node[right,thick,circle,draw=black,minimum size=22] (8) {$+$};
			\draw [thick, ->] (8.east) -- ++(0.7,0)
			node[right,node 2] (9)  {$y^{[\ell]}$};
			\draw[-{Latex[length=4,width=3.5]},thick,shorten <=0.5,shorten >=1] (1.north) to [out=10,in=170] (8.north);
		\end{tikzpicture}
	\end{center}
	\caption{Illustration of a full pre-activation residual block as proposed in \cite[Fig.1(b)]{he2016identity}.}
	\label{fig:Residualblock_new}
\end{figure}

\begin{figure}[h!]
	\centering
	\includegraphics[width=0.7\textwidth]{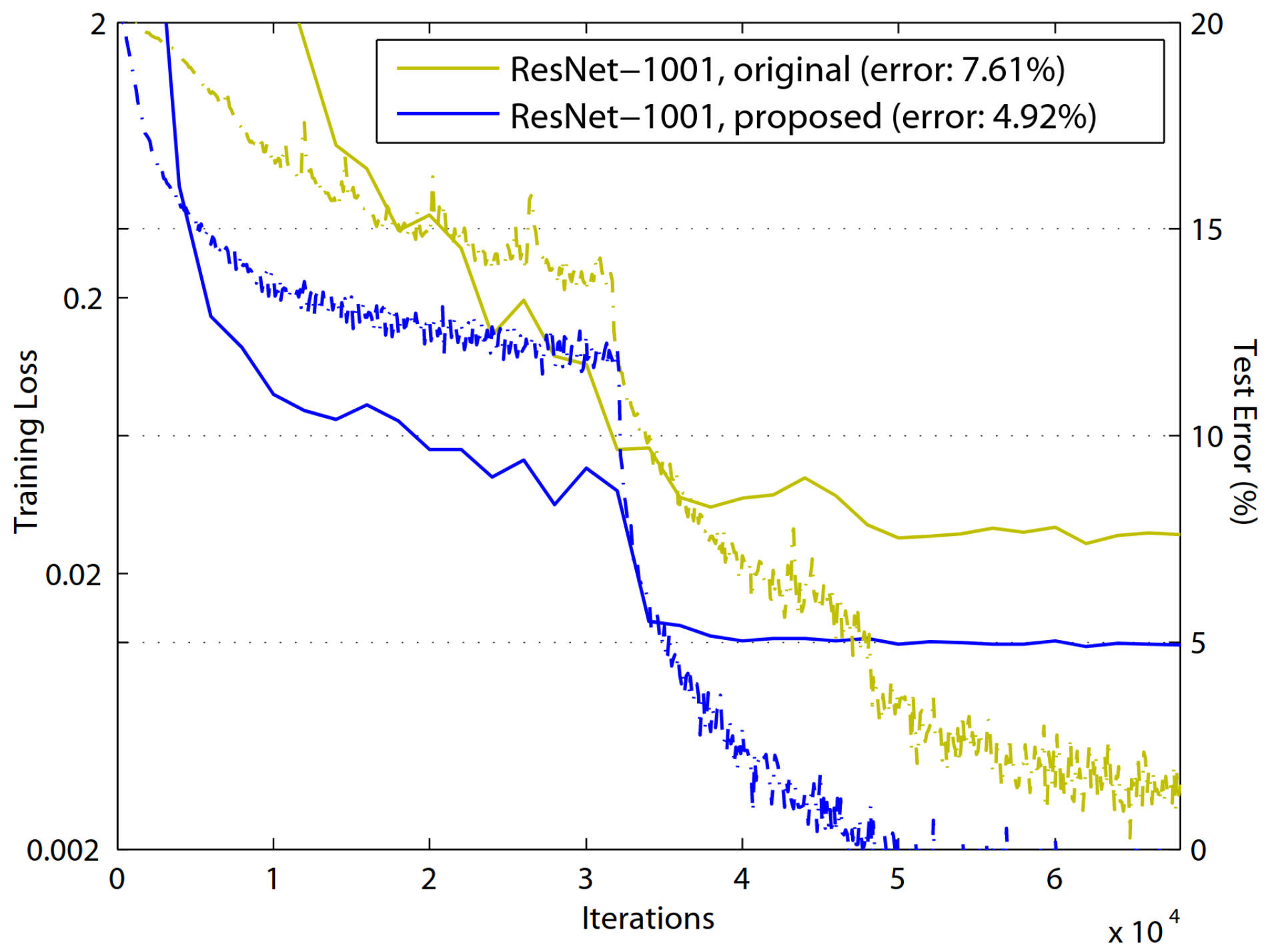}
	\caption{Training loss (dashed line, left y-axis) and test error (solid line, right y-axis) plotted against the iteration counter for a 1001-layer ResNet on the CIFAR-10 dataset. Here, the original residual block (blue) is compared to the proposed full pre-activation residual block (green). Image Source: \cite[Fig.1]{he2016identity}}
	\label{fig:ResNet1001}
\end{figure}

Subsequently, in a follow up paper, \cite{he2016identity}, several other options to sort the occurring layers in a residual block have been introduced. The option which performed best in numerical tests (see also Figure \ref{fig:ResNet1001}) is illustrated in Figure \ref{fig:Residualblock_new}.

\begin{remark}\label{rem:pre}
	The authors of \cite{he2016identity} call the residual block in Figure \ref{fig:Residualblock_new} the \textbf{full pre-activation} residual block, since both activation functions are exercised before (pre) the skip connection. Meanwhile, in the original residual block there is also a post-activation, i.e. an activation function after (post) the skip connection. In this sense, the simplified ResNet layer \eqref{eq:ResNet} can be termed a pre-activation ResNet layer. 
\end{remark}

A ResNet built with full pre-activation residual blocks can be found e.g. in Tensorflow under the name "ResNetV2". In the literature, there also exist other variants of the simplified ResNet layer, e.g. with a weight matrix applied outside the activation function.

In Figure \ref{fig:ResNet1001} we see a comparison of a 1001-layer ResNet built with original residual blocks and a 1001-layer ResNet built with full pre-activation (proposed) residual blocks. This result clearly demonstrates the advantage of full pre-activation for very deep networks, since both training loss and test error can be improved with the proposed residual block. 

\subsection{ResNet18}
\label{subsec:ResNet18}
\begin{figure}[h!]
	\centering
	\includegraphics[width=\textwidth]{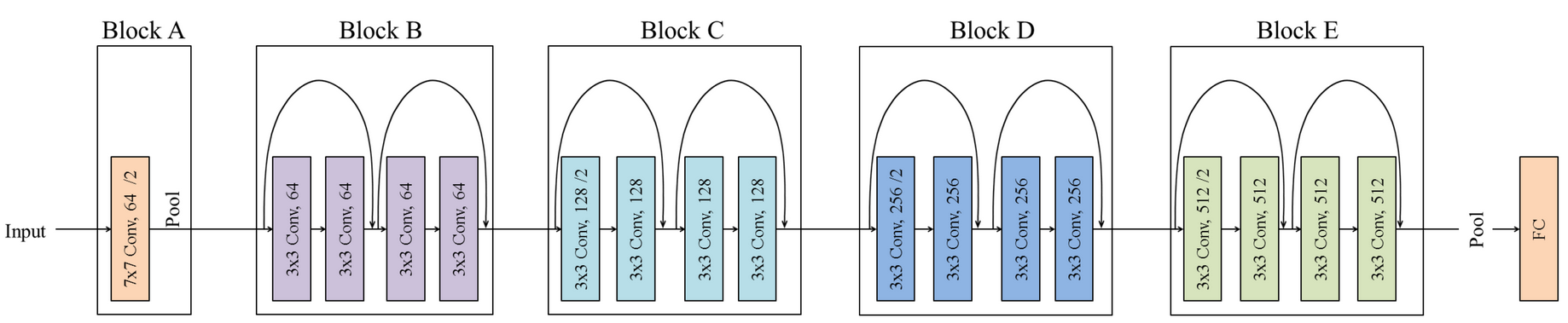}
	\caption{Illustration of "ResNet18" architecture, built of residual blocks. Image Source: \cite{ResNet18}. }
	\label{fig:ResNet18}
\end{figure}

As an example for a ResNet architecture, we look at "ResNet18", cf. Figure \ref{fig:ResNet18}. Here, 18 indicates the number of layers with learnable weights, i.e. convolutional and fully connected layers. Even though the batch normalization layers also contain learnable weights, they are typically not counted here. This is a ResNet architecture intended for use on image data sets, hence the weights layers are convolutional layers, like in a CNN. 

In Block A the input data is pre-processed, while Block B to Block E are built of two residual blocks each. To be certain which type of residual block the network is built of, it is recommended to look into the details of the implementation. However, in most cases the original residual block, cf. Figure \ref{fig:Residualblock}, is employed. Finally, as usual for a classification task, the data is flattened and passed through a fully connected (FC) layer before the output is generated. Altogether, "ResNet18" has over 10 million trainable parameters, i.e. variables.

\subsection{Transfer Learning}
\label{subsec:transfer}
An advantage of having so many different ResNet architectures available and also pre-trained is, that they can be employed for \textbf{Transfer Learning}, see e.g. \cite{torrey2010transfer}. The main idea of transfer learning is to take a model that has been trained on a (potentially large) data set for the same type of task (e.g. image classification), and then adjust the first/last layer to fit your data set. Depending on your task you may need to change one or both layers, for example:
\begin{enumerate}
	\item[(i)] Your input data has a different structure: adapt the first layer.
	\item[(ii)] Your data set has a different set of labels (supervisions): adapt the last layer.
\end{enumerate}
If your input data and labels both coincide with the original task then you don't need to employ transfer learning. You can just use the pre-trained model for your task. When adapting a layer, this layers needs to be initialized. All remaining layers can be initialized with the pre-trained weights, which will most likely give a good starting point. Then you train the adapted network on your data, which will typically take a lot loss time than training with a random initialization. Hence, transfer learning can save a significant amount of computing time. Clearly, transfer learning is also possible with other network architectures, as long as the network has been pre-trained.

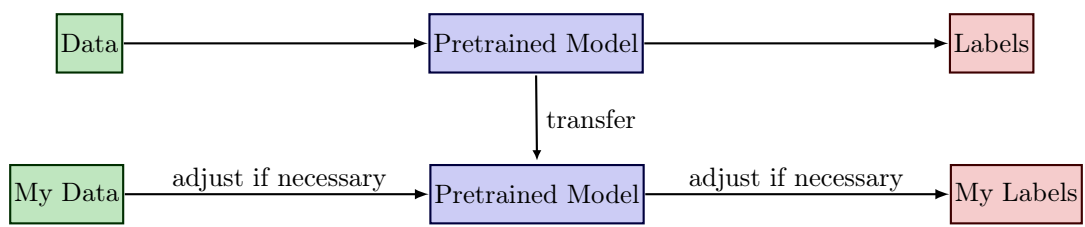
\begin{figure}[h!]
	\begin{center}
		\begin{tikzpicture}
			\node[thick,rectangle,draw=mygreen!30!black,fill=mygreen!25,minimum size=22,inner sep=1.5,outer sep=0.6] (1) at (0,0) {Data};
			\draw [thick, ->] (1.east) -- ++(4,0)
			node[right,thick,rectangle,draw=myblue!30!black,fill=myblue!20,minimum size=22,inner sep=1.5,outer sep=0.6] (2) {Pretrained Model};
			\draw [thick, ->] (2.east) -- ++(4,0)
			node[right,thick,rectangle,draw=myred!30!black,fill=myred!20,minimum size=22, inner sep=1.5,outer sep=0.6] (3) {Labels};
			\node[thick,rectangle,draw=mygreen!30!black,fill=mygreen!25,minimum size=22,inner sep=1.5,outer sep=0.6] (4) at (-0.3,-2) {My Data};
			\draw [thick, ->] (4.east) -- ++(4,0)
			node[right,thick,rectangle,draw=myblue!30!black,fill=myblue!20,minimum size=22,inner sep=1.5,outer sep=0.6] (5) {Pretrained Model};
			\draw [thick, ->] (5.east) -- ++(4,0)
			node[right,thick,rectangle,draw=myred!30!black,fill=myred!20,minimum size=22, inner sep=1.5,outer sep=0.6] (6) {My Labels};
			\draw[-{Latex[length=4,width=3.5]},thick,shorten <=0.5,shorten >=1] (2.south) to (5.north);
			\node[] (7) at (6.6,-1) {transfer};
			\node[left of =5] (8) at (3.5,-1.8) {adjust if necessary};
			\node[left of =5] (9) at (10.3,-1.8) {adjust if necessary};
		\end{tikzpicture}
	\end{center}
	\caption{Illustration of transfer learning.}
	\label{fig:transfer}
\end{figure}

\newpage
\section{Recurrent Neural Network}
\label{sec:RNN}
The Neural Networks we introduced so far rely on the \textbf{assumption of independence} among the training and test examples. They process one data point at a time, which is no problem for data sets, in which every data point is generated independently. However, for sequential data that occurs in machine translation, speech recognition, sentiment classification, etc., the dependence is highly relevant to the task.  

\textbf{Recurrent Neural Networks} (RNNs), cf. e.g. \cite[Section 10]{goodfellow2016deep} and \cite[Section 8.1]{geiger2021DL}, are connectionist models that capture the dynamics of sequences via cycles in the network of nodes. Unlike standard FNNs, recurrent neural networks retain a state that can represent information from an arbitrarily long context window.

\begin{example}[machine translation]\label{ex:RNN} $\,$\\ 
Translate a given english input sentence $u$, consisting of $T_{\operatorname{in}}$ words $u^{<t>}, t=1,\ldots,T_{\operatorname{in}}$, e.g.
	\begin{align*}
		\hspace{3cm}	&\text{The } &&\text{sun } &&\text{is } &&\text{shining } &&\text{today } \hspace{3cm}\\
		&u^{<1>} &&u^{<2>} &&u^{<3>} &&u^{<4>} &&u^{<5>} 
	\end{align*}
	to a german output sentence $y$, consisting of $T_{\operatorname{out}}$ words $y^{<t>}, t=1,\ldots,T_{\operatorname{out}}$. Hopefully, the output will be something like
	\begin{align*}
		\hspace{3.5cm}	&\text{Heute } &&\text{scheint } &&\text{die } &&\text{Sonne } \hspace{3.5cm}\\
		&y^{<1>} &&y^{<2>} &&y^{<3>} &&y^{<4>} 
	\end{align*}
\end{example}

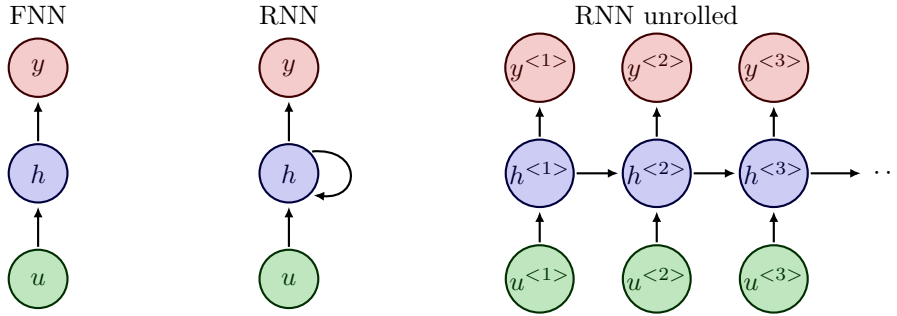
\begin{figure}[h!]
	\begin{center}
		\begin{tikzpicture}[x=2.2cm,y=1.4cm]
			\node[node 1] (1) at (0,0) {$u$};
			\node[node 2] (2) at (0,1) {$h$};
			\node[node 3] (3) at (0,2) {$y$};
			\node[] at (0,2.5) {FNN};
			\node[node 1] (4) at (1.5,0) {$u$};
			\node[node 2] (5) at (1.5,1) {$h$};
			\node[node 3] (6) at (1.5,2) {$y$};
			\node[] at (1.5,2.5) {RNN};
			\draw[-{Latex[length=4,width=3.5]},thick,shorten <=0.5,shorten >=1] (1.north) to (2);
			\draw[-{Latex[length=4,width=3.5]},thick,shorten <=0.5,shorten >=1] (2.north) to (3);
			\draw[-{Latex[length=4,width=3.5]},thick,shorten <=0.5,shorten >=1] (4.north) to (5);
			\draw[-{Latex[length=4,width=3.5]},thick,shorten <=0.5,shorten >=1] (5.north) to (6);
			\draw[-{Latex[length=4,width=3.5]},thick,shorten <=0.5,shorten >=1] (5.north east) to [out=5, in=355,looseness=3] (5.south east);
			\node[node 1] (7) at (3,0) {$u^{<1>}$};
			\node[node 2] (8) at (3,1) {$h^{<1>}$};
			\node[node 3] (9) at (3,2) {$y^{<1>}$};
			\node[node 1] (10) at (3.7,0) {$u^{<2>}$};
			\node[node 2] (11) at (3.7,1) {$h^{<2>}$};
			\node[node 3] (12) at (3.7,2) {$y^{<2>}$};
			\node[node 1] (13) at (4.4,0) {$u^{<3>}$};
			\node[node 2] (14) at (4.4,1) {$h^{<3>}$};
			\node[node 3] (15) at (4.4,2) {$y^{<3>}$};
			\node[] at (3.7,2.5) {RNN unrolled};
			\node[] (16) at (5.1,1) {$\cdots$};
			\draw[-{Latex[length=4,width=3.5]},thick,shorten <=0.5,shorten >=1] (7.north) to (8);
			\draw[-{Latex[length=4,width=3.5]},thick,shorten <=0.5,shorten >=1] (8.north) to (9);
			\draw[-{Latex[length=4,width=3.5]},thick,shorten <=0.5,shorten >=1] (10.north) to (11);
			\draw[-{Latex[length=4,width=3.5]},thick,shorten <=0.5,shorten >=1] (11.north) to (12);
			\draw[-{Latex[length=4,width=3.5]},thick,shorten <=0.5,shorten >=1] (13.north) to (14);
			\draw[-{Latex[length=4,width=3.5]},thick,shorten <=0.5,shorten >=1] (14.north) to (15);
			\draw[-{Latex[length=4,width=3.5]},thick,shorten <=0.5,shorten >=1] (8.east) to (11);
			\draw[-{Latex[length=4,width=3.5]},thick,shorten <=0.5,shorten >=1] (11.east) to (14);
			\draw[-{Latex[length=4,width=3.5]},thick,shorten <=0.5,shorten >=1] (14.east) to (16);
		\end{tikzpicture}
	\end{center}
	\caption{Feedforward Neural Network compared to Recurrent Neural Network with input $u$, output $y$ and hidden computation nodes $h$. The index is understood as time instance.}
	\label{fig:RNN}
\end{figure}

A comparison of FNN and RNN architecture can be seen in Figure \ref{fig:RNN}. For simplicity of notation we condense all hidden layers of the FNN into a representative computation node $h$. 

\begin{figure}[h!]
	\begin{minipage}{0.69\textwidth}
	\begin{center}
		\begin{tikzpicture}[x=2.2cm,y=1.4cm]
			\draw[rounded corners,myblue!30!black,fill=myblue!20,thick] (0.5,1.5) rectangle (3,3);
			\node[node 1] (1) at (1,1) {$u^{<t>}$};
			\node[node 2] (2) at (0,2) {$h^{<t-1>}$};
			\node[rounded corners,rectangle,draw,thick,fill=white] (3) at (1,2) {Weights};
			\node[rounded corners,rectangle,draw,thick,fill=white] (4) at (1.8,2) {tanh};
			\node[rounded corners,rectangle,draw,thick,fill=white] (5) at (2.5,2.5) {Weights};
			\node[node 2] (6) at (3.5,2) {$h^{<t>}$};
			\node[node 3] (7) at (2.5,3.6) {$y^{<t>}$};
			\node[] at (1.3,2.7) {\textcolor{myblue!40}{\Large{RNN Cell}}};
			\draw[-{Latex[length=4,width=3.5]},thick,shorten <=0.5,shorten >=1] (1.north) to (3);
			\draw[-{Latex[length=4,width=3.5]},thick,shorten <=0.5,shorten >=1] (2.east) to (3);
			\draw[-{Latex[length=4,width=3.5]},thick,shorten <=0.5,shorten >=1] (3.east) to (4);
			\draw[-{Latex[length=4,width=3.5]},thick,shorten <=0.5,shorten >=1] (4.east) to (5);
			\draw[-{Latex[length=4,width=3.5]},thick,shorten <=0.5,shorten >=1] (4.east) to (6);
			\draw[-{Latex[length=4,width=3.5]},thick,shorten <=0.5,shorten >=1] (5.north) to (7);
		\end{tikzpicture}
	\end{center}
\end{minipage}
\begin{minipage}{0.3\textwidth}
	\caption{Architecture of a RNN cell.}
	\label{fig:RNNcell}
\end{minipage}
\end{figure}
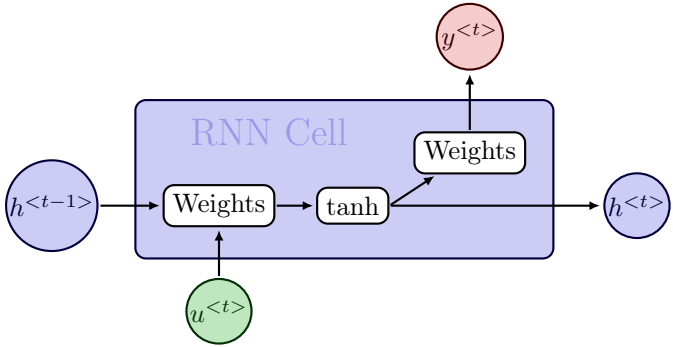

In RNNs the computation nodes $h$ are often called \textbf{RNN cells}, cf. \cite[Section 10.2]{goodfellow2016deep}. A RNN cell for a time instance $t$ takes as an input $u^{<t>}$ and $h^{<t-1>}$, and computes the outputs $h^{<t>}$ and $y^{<t>}$, cf. Figure \ref{fig:RNNcell}. More specifically for all $t=1,\ldots,T_{\operatorname{out}}$
\begin{align} \label{eq:RNN1}
	h^{<t>} &= \sigma \left( W_{\operatorname{in}} \cdot [h^{<t-1>};u^{<t>}] + b \right), \\
	y^{<t>} &= W_{\operatorname{out}}\cdot h^{<t>}.\label{eq:RNN2}
\end{align}

The equations \eqref{eq:RNN1} and \eqref{eq:RNN2} describe the forward propagation in RNNs. Here, $[h^{<t-1>} ; u^{<t>}]$ denotes the concatenation of the vectors, and $h^{<0>}$ is set to a vector of zeros, so that we do not need to formulate a special case for $t=1$. Depending on the application, a softmax function may be applied to $W_{\operatorname{out}} h^{<t>}$ to get the output $y^{<t>}$. 

\begin{figure}[h!]
	\begin{center}
		\begin{tikzpicture}[x=2.2cm,y=1.4cm]
			\node[node 1,minimum size=15] (1) at (0,0) {};
			\node[node 2,minimum size=15] (2) at (0,0.8) {};
			\node[node 3,minimum size=15] (3) at (0,1.6) {};
			\node[node 2,minimum size=15] (4) at (0.4,0.8) {};
			\node[node 3,minimum size=15] (5) at (0.4,1.6) {};
			\node[node 2,minimum size=15] (6) at (0.8,0.8) {};
			\node[node 3,minimum size=15] (7) at (0.8,1.6) {};
			\node[] at (0.4,2.1) {one to many};
			\draw[connect arrow] (1) -- (2);
			\draw[connect arrow] (2) -- (3);
			\draw[connect arrow] (2) -- (4);
			\draw[connect arrow] (4) -- (5);
			\draw[connect arrow] (4) -- (6);
			\draw[connect arrow] (6) -- (7);
			\node[node 1,minimum size=15] (8) at (1.6,0) {};
			\node[node 2,minimum size=15] (9) at (1.6,0.8) {};
			\node[node 1,minimum size=15] (10) at (2,0) {};
			\node[node 2,minimum size=15] (11) at (2,0.8) {};
			\node[node 1,minimum size=15] (12) at (2.4,0) {};
			\node[node 2,minimum size=15] (13) at (2.4,0.8) {};
			\node[node 3,minimum size=15] (14) at (2.4,1.6) {};
			\node[] at (2,2.1) {many to one};
			\draw[connect arrow] (8) -- (9);
			\draw[connect arrow] (9) -- (11);
			\draw[connect arrow] (10) -- (11);
			\draw[connect arrow] (11) -- (13);
			\draw[connect arrow] (12) -- (13);
			\draw[connect arrow] (13) -- (14);
			\node[node 1,minimum size=15] (15) at (3.2,0) {};
			\node[node 2,minimum size=15] (16) at (3.2,0.8) {};
			\node[node 1,minimum size=15] (17) at (3.6,0) {};
			\node[node 2,minimum size=15] (18) at (3.6,0.8) {};
			\node[node 1,minimum size=15] (19) at (4,0) {};
			\node[node 2,minimum size=15] (20) at (4,0.8) {};
			\node[node 2,minimum size=15] (21) at (4.4,0.8) {};
			\node[node 3,minimum size=15] (22) at (4.4,1.6) {};
			\node[node 2,minimum size=15] (23) at (4.8,0.8) {};
			\node[node 3,minimum size=15] (24) at (4.8,1.6) {};
			\node[] at (4,2.1) {many to many};
			\draw[connect arrow] (15) -- (16);
			\draw[connect arrow] (16) -- (18);
			\draw[connect arrow] (17) -- (18);
			\draw[connect arrow] (18) -- (20);
			\draw[connect arrow] (19) -- (20);
			\draw[connect arrow] (20) -- (21);
			\draw[connect arrow] (21) -- (22);
			\draw[connect arrow] (21) -- (23);
			\draw[connect arrow] (23) -- (24);
			\node[node 1,minimum size=15] (25) at (5.6,0) {};
			\node[node 2,minimum size=15] (26) at (5.6,0.8) {};
			\node[node 3,minimum size=15] (27) at (5.6,1.6) {};
			\node[node 1,minimum size=15] (28) at (6,0) {};
			\node[node 2,minimum size=15] (29) at (6,0.8) {};
			\node[node 3,minimum size=15] (30) at (6,1.6) {};
			\node[node 1,minimum size=15] (31) at (6.4,0) {};
			\node[node 2,minimum size=15] (32) at (6.4,0.8) {};
			\node[node 3,minimum size=15] (33) at (6.4,1.6) {};
			\node[] at (6,2.1) {many to many};
			\draw[connect arrow] (25) -- (26);
			\draw[connect arrow] (26) -- (27);
			\draw[connect arrow] (26) -- (29);
			\draw[connect arrow] (28) -- (29);
			\draw[connect arrow] (29) -- (30);
			\draw[connect arrow] (29) -- (32);
			\draw[connect arrow] (31) -- (32);
			\draw[connect arrow] (32) -- (33);
		\end{tikzpicture}
	\end{center}
	\caption{Illustration of different types of RNN architectures.}
	\label{fig:RNN_architectures}
\end{figure}
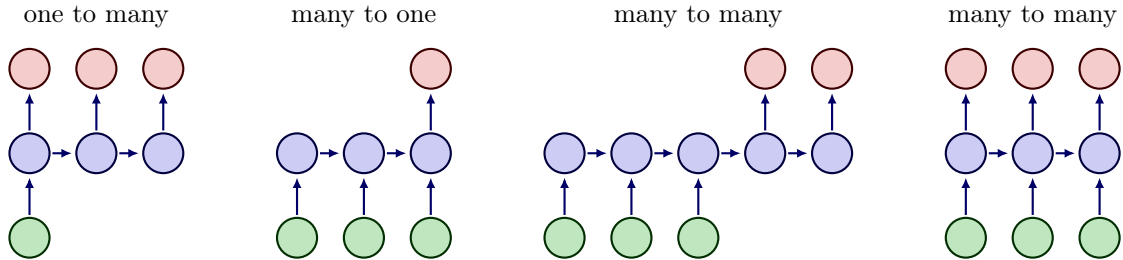

It may happen that input and output have different lengths $T_{\operatorname{in}} \neq T_{\operatorname{out}}$, see e.g. Example \ref{ex:RNN}. Depending on the task and the structure of the data, there exist various types of RNN architectures, cf. \cite[Section 8.1]{geiger2021DL} and Figure \ref{fig:RNN_architectures}:

\begin{itemize}
	\item one to many, e.g. image description (image to sentence),
	\item many to one, e.g. sentiment analysis (video to word),
	\item many to many, e.g. machine translation (sentence to sentence), like Example \ref{ex:RNN},
	\item many to many, e.g. object tracking (video to object location per frame).
\end{itemize}

We note that the weights $W_{\operatorname{in}}, W_{\operatorname{out}}$ and bias $b$ in \eqref{eq:RNN1} and \eqref{eq:RNN2} do not change over time, but coincide for all temporal layers of the RNN. Sharing the variables allows the RNN to model variable length sequences, whereas if we had specific parameters for each value of the order parameter, we could not generalize to sequence lengths not seen during training. Typically, $\sigma = \tanh$ is chosen in RNNs, and this does also not vary between the layers. To obtain a complete optimization problem \eqref{eq:LP}, we still need a loss function $\mathscr{L}$, since the RNN represents only the network $\mathcal{F}$. To this end, each output $y^{<t>}$ is evaluated with a loss function $\mathscr{L}^{<t>}$ and the final loss is computed by taking the sum over all time instances 
\[ \mathscr{L}(\theta) = \sum_{t=1}^{T_{\operatorname{out}}} \mathscr{L}^{<t>}(y^{<t>}(\theta)). \]
Here, as usual, $\theta$ contains the weights $W_{\operatorname{in}},W_{\operatorname{out}}$, and bias $b$.


\subsection{Variants of RNNs}
\label{subsec:RNNVariants}
We briefly introduce two popular variants of RNNs.

\begin{figure}[h!]
	\centering
	\includegraphics[width=\textwidth]{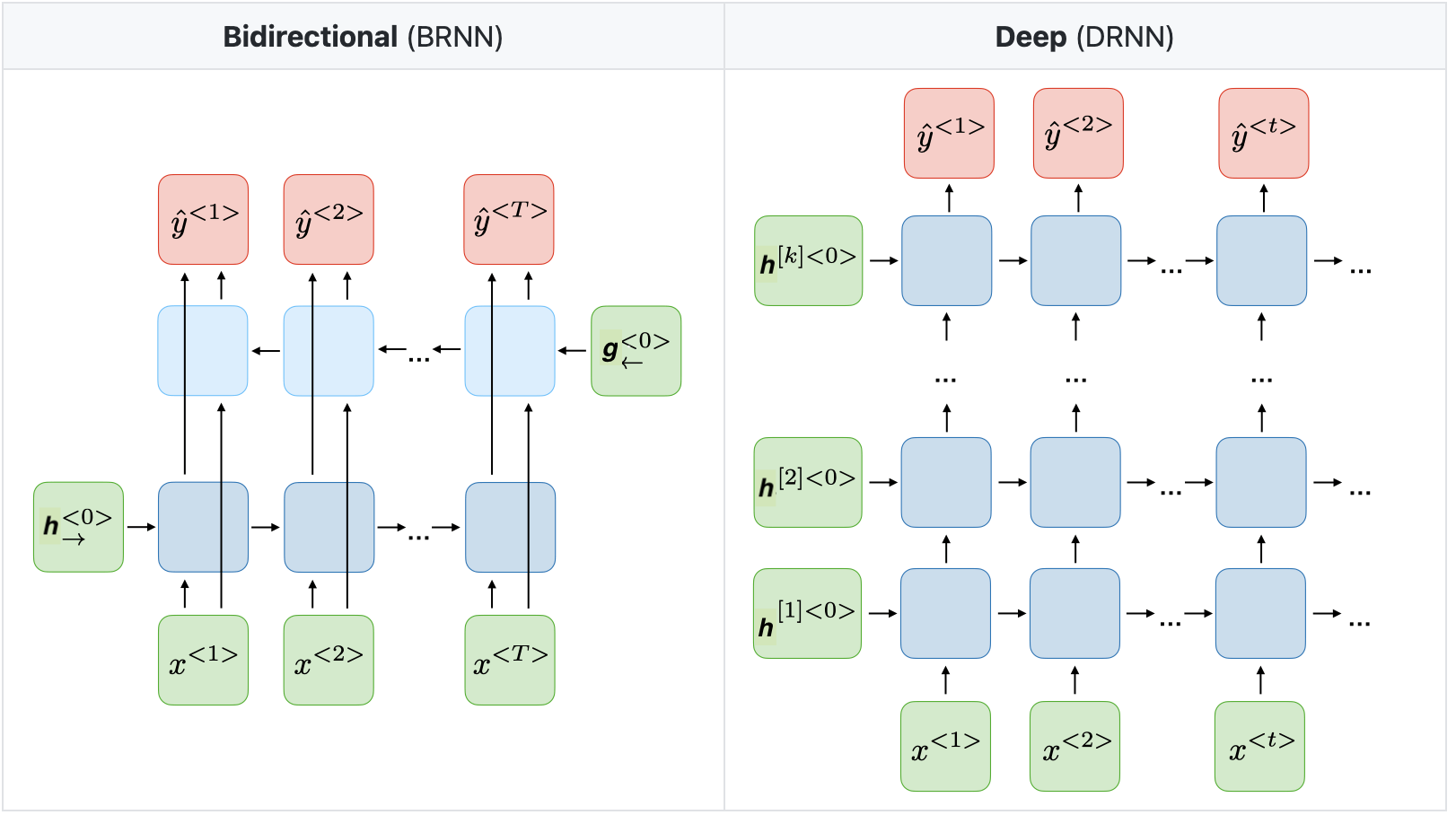}
	\caption{Examples of Bidirectional RNNs and Deep RNNs. Here, the inputs are denoted by $x$ instead of $u$. Image source: \url{https://stanford.edu/~shervine/teaching/cs-230/}. }
	\label{fig:bi}
\end{figure}

In many applications the output at time $t$ should be a prediction depending on the whole input sequence, not only the "earlier" inputs $u^{<i>}$ with $i \leq t$. E.g., in speech recognition, the correct interpretation of the current sound as a phoneme may depend on the next few phonemes because of co-articulation and potentially may even depend on the next few words because of the linguistic dependencies between nearby words.
As a remedy, we can combine a forward-going RNN and a backward-going RNN, which is then called a \textbf{Bidirectional RNN}, \cite[Section 10.3]{goodfellow2016deep}. This architecture allows to compute an output $y^{<t>}$ that depends on both the past and the future inputs, but is most sensitive to the input values around time $t$. Figure \ref{fig:bi} (left) illustrates the typical bidirectional RNN, with $h^{<t>}$ and $g^{<t>}$ representing the states of the sub-RNNs that move forward and backward through time, respectively.

Another variant of RNNs is the \textbf{Deep RNN}, \cite[Section 10.5]{goodfellow2016deep}. As seen in FNNs, Section \ref{sec:FNN}, multiple hidden layers allow the network to have a higher expressiveness. Similarly, a RNN can be made deep by stacking RNN cells, see Figure \ref{fig:bi} (right). 

\subsection{Long term dependencies}
\label{subsec:RNNltd}
In this section we investigate one of the main challenges, that a RNN can encounter, cf. \cite[Section 10.7]{goodfellow2016deep}.
Consider the following illustrative example.

\begin{example}$\,$
	
Predict the next word in the sequence:
\begin{enumerate}
	\item The cat, which ..., \textcolor{myred}{was} ...
	\item The cats, which ..., \textcolor{myred}{were} ...
\end{enumerate}
Here, depending on whether we are talking about one cat or multiple cats the verb has to be adjusted. The "..." part in the sentence can be very extensive, so that the dependence becomes long.
\end{example}

The gradient from the output $y^{<t>}$ with large $t$ has to propagate back through many layers to affect weights in early layers. Here, the vanishing gradient and exploding gradient problems (cf. Section \ref{sec:ResNet}) may occur and hinder training. The exploding gradient problem can be solved relatively robust by \textbf{gradient clipping}, see e.g. \cite[Section 10.11.1]{goodfellow2016deep}. The idea is quite simple. If a gradient $\partial_{\theta_i}{\mathscr{L}}$, with respect to some variable $\theta_i$ gets too large, we rescale it. I.e. if $\| \partial_{\theta_i}{\mathscr{L}}\| \geq C \in \mathbb{R}$ for a hyperparameter $C$, we set 
\begin{equation*}
	\partial_{\theta_i}{\mathscr{L}} \leftarrow C \cdot \frac{ \partial_{\theta_i}{\mathscr{L}}}{\|\partial_{\theta_i}{\mathscr{L}} \|}.
\end{equation*}
Let us remark that this is a heuristic approach. In contrast, the vanishing gradient problem is more difficult to solve. 

\begin{figure}[h!]
	\centering
	\begin{tikzpicture}[x=2.2cm,y=1.4cm]
		\node[thick,circle,draw=black,fill=myorange,inner sep=0.5,outer sep=0.6,minimum size=15] (1) at (0,0) {$u^{<1>}$};
		\node[thick,circle,draw=black,fill=myorange,inner sep=0.5,outer sep=0.6,minimum size=15] (2) at (0,1) {$h^{<1>}$};
		\node[thick,circle,draw=black,fill=myorange,inner sep=0.5,outer sep=0.6,minimum size=15] (3) at (0,2) {$y^{<1>}$};
		\node[thick,circle,draw=black,inner sep=0.5,outer sep=0.6,minimum size=15] (4) at (0.7,0) {$u^{<2>}$};
		\node[thick,circle,draw=black,fill=myorange!70,inner sep=0.5,outer sep=0.6,minimum size=15] (5) at (0.7,1) {$h^{<2>}$};
		\node[thick,circle,draw=black,fill=myorange!70,inner sep=0.5,outer sep=0.6,minimum size=15] (6) at (0.7,2) {$y^{<2>}$};
		\node[thick,circle,draw=black,inner sep=0.5,outer sep=0.6,minimum size=15] (7) at (1.4,0) {$u^{<3>}$};
		\node[thick,circle,draw=black,fill=myorange!50,inner sep=0.5,outer sep=0.6,minimum size=15] (8) at (1.4,1) {$h^{<3>}$};
		\node[thick,circle,draw=black,fill=myorange!50,inner sep=0.5,outer sep=0.6,minimum size=15] (9) at (1.4,2) {$y^{<3>}$};
		\node[thick,circle,draw=black,inner sep=0.5,outer sep=0.6,minimum size=15] (10) at (2.1,0) {$u^{<4>}$};
		\node[thick,circle,draw=black,fill=myorange!30,inner sep=0.5,outer sep=0.6,minimum size=15] (11) at (2.1,1) {$h^{<4>}$};
		\node[thick,circle,draw=black,fill=myorange!30,inner sep=0.5,outer sep=0.6,minimum size=15] (12) at (2.1,2) {$y^{<4>}$};
		\node[thick,circle,draw=black,inner sep=0.5,outer sep=0.6,minimum size=15] (13) at (2.8,0) {$u^{<5>}$};
		\node[thick,circle,draw=black,fill=myorange!10,inner sep=0.5,outer sep=0.6,minimum size=15] (14) at (2.8,1) {$h^{<5>}$};
		\node[thick,circle,draw=black,fill=myorange!10,inner sep=0.5,outer sep=0.6,minimum size=15] (15) at (2.8,2) {$y^{<5>}$};
		\node[thick,circle,draw=black,inner sep=0.5,outer sep=0.6,minimum size=15] (16) at (3.5,0) {$u^{<6>}$};
		\node[thick,circle,draw=black,inner sep=0.5,outer sep=0.6,minimum size=15] (17) at (3.5,1) {$h^{<6>}$};
		\node[thick,circle,draw=black,inner sep=0.5,outer sep=0.6,minimum size=15] (18) at (3.5,2) {$y^{<6>}$};
		\node[thick,circle,draw=black,inner sep=0.5,outer sep=0.6,minimum size=15] (19) at (4.2,0) {$u^{<7>}$};
		\node[thick,circle,draw=black,inner sep=0.5,outer sep=0.6,minimum size=15] (20) at (4.2,1) {$h^{<7>}$};
		\node[thick,circle,draw=black,inner sep=0.5,outer sep=0.6,minimum size=15] (21) at (4.2,2) {$y^{<7>}$};
		\draw[connect arrow] (1) -- (2);
		\draw[connect arrow] (2) -- (3);
		\draw[connect arrow] (2) -- (5);
		
		\draw[connect arrow] (4) -- (5);
		\draw[connect arrow] (5) -- (6);
		\draw[connect arrow] (5) -- (8);
		
		\draw[connect arrow] (7) -- (8);
		\draw[connect arrow] (8) -- (9);
		\draw[connect arrow] (8) -- (11);
		
		\draw[connect arrow] (10) -- (11);
		\draw[connect arrow] (11) -- (12);
		\draw[connect arrow] (11) -- (14);
		
		\draw[connect arrow] (13) -- (14);
		\draw[connect arrow] (14) -- (15);
		\draw[connect arrow] (14) -- (17);
		
		\draw[connect arrow] (16) -- (17);
		\draw[connect arrow] (17) -- (18);
		\draw[connect arrow] (17) -- (20);
		
		\draw[connect arrow] (19) -- (20);
		\draw[connect arrow] (20) -- (21);
	\end{tikzpicture}
	\caption{Illustration of vanishing gradient problem for RNNs. The shading of the
		nodes indicates the sensitivity over time of the network nodes to the input $u^{<1>}$ (the darker the shade, the greater the sensitivity). The sensitivity
		decays over time.}
	\label{fig:GV}
\end{figure}
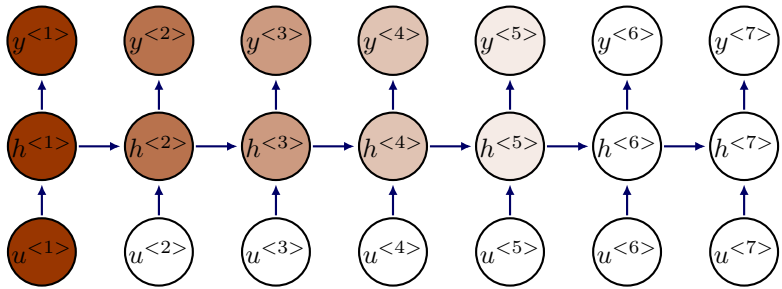

A common remedy is to modify the RNN cell so that it can capture long term dependencies better, and avoids vanishing gradients. Two popular options are \textbf{Gated Recurrent Unit (GRU)} from 2014 \cite{cho2014learning}, and the cell architecture as suggested already in 1997 in \textbf{Long Short Term Memory (LSTM)} networks \cite{hochreiter1997long}. The core idea in both cell architectures is to add gating mechanisms. These gates have a significant influence on whether, and how severely, the input and previous hidden state influence the output and new hidden state. Additionally, the gating mechanism helps to solve the vanishing gradient problem. 

\subsubsection{Gated Recurrent Unit}
The gated recurrent unit has a reset (or relevance) gate $\Gamma_r$ and an update gate $\Gamma_u$. The computations for one unit are as follows
\begin{align*}
	\Gamma_r &= \sigma\left( W_r \cdot[h^{<t-1>};u^{<t>}] + b_r \right), &&\textit{reset gate}\\
	\Gamma_u &= \sigma\left( W_u\cdot [h^{<t-1>};u^{<t>}] + b_u \right), &&\textit{update gate}\\
	\widetilde{h}^{<t>} &= \tanh \left( W_{\operatorname{in}} \cdot[\Gamma_r \odot h^{<t-1>} ; u^{<t>}] + b \right), &&\textit{hidden state candidate} \\
	h^{<t>} &= \Gamma_u \odot \widetilde{h}^{<t>} + (1-\Gamma_u) \odot h^{<t-1>},  &&\textit{hidden state}\\
	y^{<t>} &= W_{\operatorname{out}} \cdot h^{<t>}.  &&\textit{output}
\end{align*}
The computations of the hidden state candidate $\widetilde{h}^{<t>}$ and the output $y^{<t>}$ resemble the computations in the RNN cell \eqref{eq:RNN1} and \eqref{eq:RNN2}, respectively. However, e.g. if $\Gamma_u = 0$, then the new hidden state will coincide with the previous hidden state and the candidate will not be taken into account. Also, the GRU has significantly more variables per cell, in comparison with the standard RNN cell. 

\begin{figure}[h!]
		\begin{center}
			\begin{tikzpicture}[x=2.2cm,y=1.4cm]
				\draw[rounded corners,myblue!30!black,fill=myblue!20,thick] (0.7,0.5) rectangle (5,5.5);
				\node[node 1] (1) at (1,0) {$u^{<t>}$};
				\node[thick,circle,draw=black,fill=white,minimum size=14] (2) at (1,1.8) {};
				\node[thick,circle,draw=black,fill=white,minimum size=14] (3) at (1.7,1) {};
				\node[thick,circle,draw=black,fill=white,minimum size=14] (4) at (1.7,3.8) {$\cdot$};
				\node[rectangle,draw,thick,fill=white,minimum size=20] (5) at (2.5,2.5) {$\sigma$};
				\coordinate (6) at (2.5,1.8) ;
				\node[rectangle,draw,thick,fill=white,minimum size=20] (7) at (3.5,2.5) {$\sigma$};
				\coordinate (8) at (3.5,1.8) ;
				\node[rectangle,draw,thick,fill=white,minimum size=20] (9) at (4.5,2.5) {$\tanh$};
				\coordinate (10) at (4.5,1) ;
				\node[thick,circle,draw=black,fill=white,minimum size=14] (11) at (4.5,3.8) {$\cdot$};
				\node[thick,circle,draw=black,fill=white,minimum size=14] (12) at (3.5,3.8) {$1 -$};
				\node[thick,circle,draw=black,fill=white,minimum size=14] (13) at (3.5,4.7) {$\cdot$};
				\node[thick,circle,draw=black,fill=white,minimum size=14] (14) at (4.5,4.7) {};
				\coordinate (15) at (1,4.7);
				\node[node 2] (16) at (0.3,4.7) {$h^{<t-1>}$};
				\node[node 2] (17) at (5.5,4.7) {$h^{<t>}$};
				\node[node 3] (18) at (4.5,6.2) {$y^{<t>}$};
				\coordinate (19) at (1,1);
				\node[] at (2.3,3.1) {$\Gamma_r$};
				\node[] at (3.35,3.1) {$\Gamma_u$};
				\node[] at (4.75,3.1) {$\widetilde{h}^{<t>}$};
				\draw[connect arrow] (1) -- (2);
				\draw[connect] (1) -- (19);
				\draw[connect arrow] (19) -- (3);
				\draw[connect arrow] (4) -- (3);
				\draw[connect] (2) -- (6);
				\draw[connect arrow] (6) -- (5);
				\draw[connect] (6) -- (8);
				\draw[connect arrow] (8) -- (7);
				\draw[connect arrow] (5) -- (4);
				\draw[connect] (3) -- (10);
				\draw[connect arrow] (10) -- (9);
				\draw[connect arrow] (9) -- (11);
				\draw[connect arrow] (7) -- (12);
				\draw (14.north) -- (14.south)
				(14.west) -- (14.east);
				\draw[connect arrow] (12) -- (13);
				\draw[connect arrow] (11) -- (14);
				\draw[connect arrow] (7) -- (11);
				\draw[connect arrow] (13) -- (14);
				\draw[connect arrow] (15) -- (2);
				\draw[connect arrow] (15) -- (4);
				\draw[connect arrow] (15) -- (13);
				\draw[connect] (16) -- (15);
				\draw[connect arrow] (14) -- (17);
				\draw[connect arrow] (14) -- (18);
				\node[] at (2,5.1) {\textcolor{myblue!40}{ \LARGE{GRU} } };
			\end{tikzpicture}
		\end{center}
		\caption{Architecture of a gated recurrent unit. Weights are omitted in this illustration. A white circle illustrates concatenation, while a circle with a dot represents the Hadamard product and a circle with a plus indicates an addition.}
		\label{fig:GRU}
\end{figure}
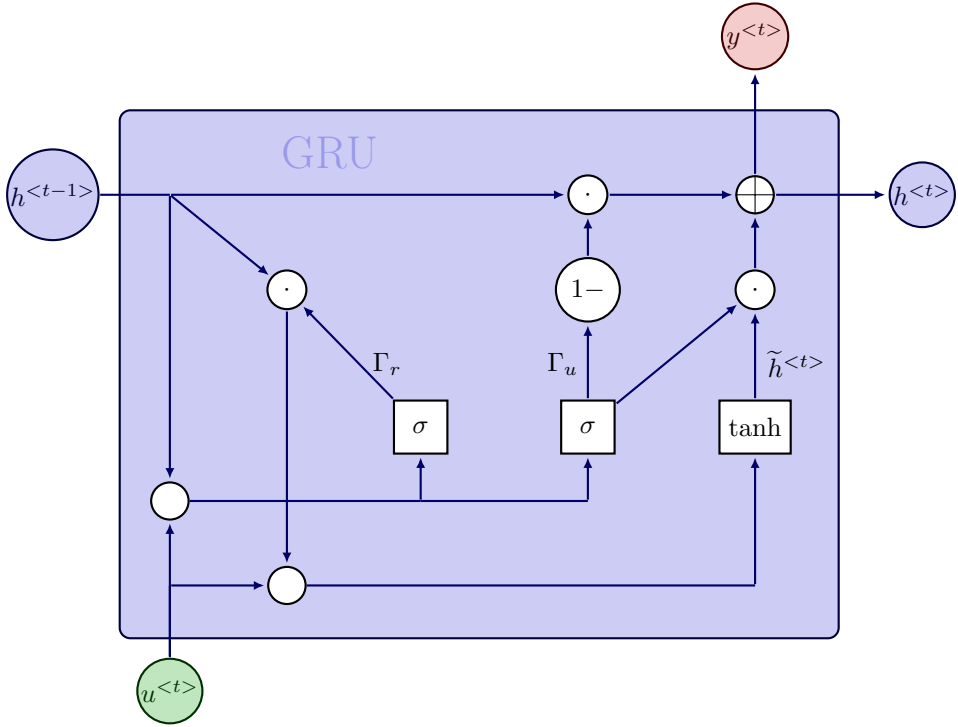

\subsubsection{Long Short Term Memory}
\begin{figure}[h!]
	\begin{center}
		\begin{tikzpicture}[x=2.2cm,y=1.4cm]
			\draw[rounded corners,myblue!30!black,fill=myblue!20,thick] (0.7,0.5) rectangle (5.5,5.5);
			\node[node 1] (1) at (1,0) {$u^{<t>}$};
			\node[node 2] (2) at (0,1.3) {$h^{<t-1>}$};
			\node[node 2] (3) at (0.3,4.7) {$c^{<t-1>}$};
			\node[node 2] (4) at (5.9,1.3) {$h^{<t>}$};
			\node[node 2] (5) at (6.3,4.7) {$c^{<t>}$};
			\node[node 3] (6) at (5.9,6.2) {$y^{<t>}$};
			\node[thick,circle,draw=black,fill=white,minimum size=14] (7) at (1,1.3) {};
			\node[rectangle,draw,thick,fill=white,minimum size=20] (8) at (1.5,2.2) {$\sigma$};
			\node[rectangle,draw,thick,fill=white,minimum size=20] (9) at (2.3,2.2) {$\sigma$};
			\node[rectangle,draw,thick,fill=white,minimum size=20] (10) at (3.1,2.2) {$\tanh$};
			\node[rectangle,draw,thick,fill=white,minimum size=20] (11) at (3.9,2.2) {$\sigma$};
			\node[thick,circle,draw=black,fill=white,minimum size=14] (12) at (1.5,4.7) {$\cdot$};
			\node[thick,circle,draw=black,fill=white,minimum size=14] (13) at (3.1,4.7) {};
			\node[thick,circle,draw=black,fill=white,minimum size=14] (14) at (3.1,3.3) {$\cdot$};
			\node[thick,circle,draw=black,fill=white,minimum size=14] (15) at (4.7,3.3) {$\cdot$};
			\node[rectangle,draw,thick,fill=white,minimum size=20] (16) at (4.7,4) {$\tanh$};
			\coordinate (17) at (1.5,1.3);
			\coordinate (18) at (2.3,1.3);
			\coordinate (19) at (3.1,1.3);
			\coordinate (20) at (3.9,1.3);
			\coordinate (21) at (4.7,1.3);
			\coordinate (22) at (4.7,4.7);
			\coordinate (23) at (5.9,4.5);
			\coordinate (24) at (5.9,4.9);
			
			\node[] at (1.65,2.8) {$\Gamma_f$};
			\node[] at (2.55,2.8) {$\Gamma_i$};
			\node[] at (3.3,2.8) {$\widetilde{c}^{<t>}$};
			\node[] at (4.15,2.8) {$\Gamma_o$};

			\draw[connect arrow] (3) -- (12);
			\draw[connect] (7) -- (17);
			\draw[connect arrow] (12) -- (13);
			\draw[connect arrow] (13) -- (5);
			\draw[connect] (17) -- (18);
			\draw[connect arrow] (17) -- (8);
			\draw[connect] (18) -- (19);
			\draw[connect arrow] (18) -- (9);
			\draw[connect arrow] (19) -- (10);
			\draw[connect] (19) -- (20);
			\draw[connect arrow] (20) -- (11);
			\draw[connect arrow] (8) -- (12);
			\draw[connect arrow] (9) -- (14);
			\draw[connect arrow] (10) -- (14);
			\draw[connect arrow] (14) -- (13);
			\draw[connect arrow] (11) -- (15);
			\draw[connect arrow] (22) -- (16);
			\draw[connect arrow] (16) -- (15);
			\draw[connect] (15) -- (21);
			\draw[connect arrow] (21) -- (4);
			\draw (13.north) -- (13.south)
			(13.west) -- (13.east);
			\draw[connect arrow] (1) -- (7);
			\draw[connect arrow] (2) -- (7);
			\draw[connect] (4) -- (23);
			\draw[connect arrow] (24) -- (6);
			\node[] at (2,5.1) {\textcolor{myblue!40}{ \LARGE{LSTM Cell} } };
		\end{tikzpicture}
	\end{center}
	\caption{Architecture of a LSTM cell. Weights are omitted in this illustration. A white circle illustrates concatenation, while a circle with a dot represents the Hadamard product and a circle with a plus indicates an addition.}
	\label{fig:LSTM}
\end{figure}
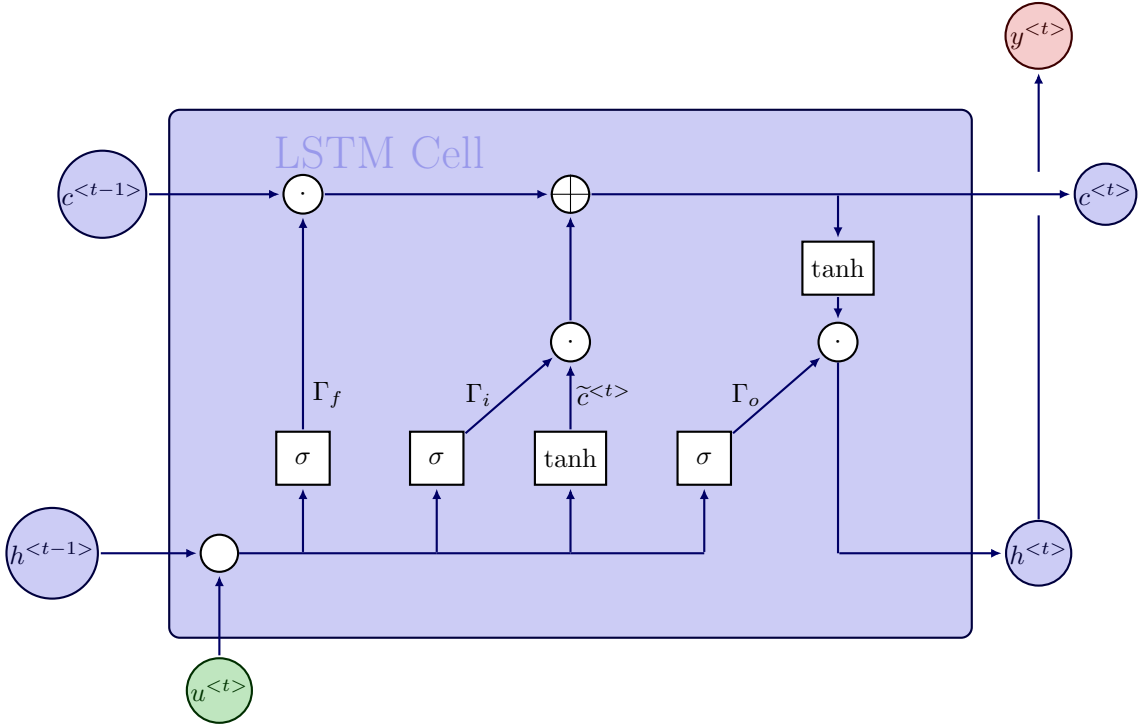

The key to LSTM networks is that in addition to the hidden state, there also exists a cell state $c^{<t>}$, which is propagated through the network. It can be understood like a conveyor belt, which only has minor interactions and runs down the entire chain of LSTM cells, see Figure \ref{fig:LSTM}. This allows information to flow through the network easily. In contrast to GRU, the LSTM cell contains three gates: the forget gate $\Gamma_f$, input gate $\Gamma_i$ and output gate $\Gamma_o$. 
\begin{align*}
	\Gamma_f &= \sigma\left( W_f \cdot[h^{<t-1>};u^{<t>}] + b_f \right), &&\textit{forget gate}\\
	\Gamma_i &= \sigma\left( W_i\cdot [h^{<t-1>};u^{<t>}] + b_i \right), &&\textit{input gate}\\
	\Gamma_o &= \sigma\left( W_o\cdot [h^{<t-1>};u^{<t>}] + b_o \right), &&\textit{output gate}\\
	\widetilde{c}^{<t>} &= \tanh \left( W_c \cdot[h^{<t-1>} ; u^{<t>}] + b_c \right), &&\textit{cell state candidate} \\
	c^{<t>} &= \Gamma_f \odot \widetilde{c}^{<t-1>} + \Gamma_i \odot \widetilde{c}^{<t>},  &&\textit{cell state}\\
	h^{<t>} &= \Gamma_o \odot \tanh(c^{<t>}), &&\textit{hidden state}\\
	y^{<t>} &= W_{\operatorname{out}} \cdot h^{<t>}.  &&\textit{output}
\end{align*}

\begin{remark}
	Without gates, i.e. $\Gamma_f = \Gamma_i = \Gamma_o = 1$, the LSTM network has a certain similarity with the ResNet structure, which was developed later than the LSTM, in 2016 in \cite{he2016deep}. This is not so surprising, since both networks aim at solving the vanishing gradient problem. In fact, propagating the cell state has similar effects on the gradients as introducing skip connections. 
\end{remark}

\subsection{Language processing}
An important application of RNNs is language processing, e.g. machine translation, see Example \ref{ex:RNN}. In such tasks the words need to be represented, so that the RNN can work with them. Furthermore, we need a way to deal with punctuation marks, and an indicator for the end of a sentence.

To represent the words, we form a dictionary. For the english language we will end up with a vector containing more than 10000 words. Intuitively, we sort the words alphabetically and to simplify computations we use a \textbf{one-hot representation}. E.g., if "the" is the 8367th word in the english dictionary vector, we represent the first input
\begin{equation*}
	u^{<1>} = \begin{pmatrix}
		0 & \ldots & 0 & 1 & 0 & \ldots & 0
	\end{pmatrix}^{\top}
	= e_{8367},
\end{equation*}
with the 8367th unit vector. This allows for an easy way to measure correctness in supervised learning and later on we can use the dictionary to recover the words. Additionally, it is common to create a token for unknown words, which are not in the dictionary. Punctuation marks can either be ignored, or we also create tokens for them. However, the dictionary should at least contain an "end of sentence" token to separate sentences from each other.


\printbibliography

\end{document}